\documentclass[accepted]{uai2022} 

\usepackage[american]{babel}

\usepackage{natbib} 
\usepackage{mathtools} 
\usepackage{booktabs} 
\usepackage{tikz} 

\usepackage{microtype}
\usepackage{graphicx}
\usepackage{subfigure}
\usepackage{booktabs} 
\usepackage{url}
\usepackage{amsmath,amsfonts,bm}
\usepackage{amsthm,amssymb}
\usepackage{bbm}   
\usepackage{diagbox}
\usepackage{booktabs}  
\usepackage{color}
\usepackage{xcolor}
\definecolor{darkblue}{RGB}{25, 50, 112}
\usepackage{float}
\usepackage{lipsum}

\usepackage{amsmath,amsfonts,amssymb}
\usepackage{booktabs}
\usepackage{graphicx}
\usepackage{hyperref}
\usepackage{url}
\usepackage{newtxmath}
\usepackage{enumitem}
\usepackage{multirow}
\usepackage{tablefootnote}

\usepackage{bbm}
\usepackage{wrapfig}
\usepackage{nicefrac}

\usepackage{chngcntr}

\newcommand{\h}{\ell}

\usepackage{amsmath}
\usepackage{amssymb}
\usepackage{mathtools}
\usepackage{amsthm}

\usepackage[capitalize,noabbrev]{cleveref}  

\theoremstyle{plain}

\theoremstyle{definition}

\theoremstyle{remark}

\DeclareMathOperator{\KL}{KL}
\DeclareMathOperator*{\argmax}{arg\ max}

\newcommand{\newtext}[1]{\textcolor{black}{#1}}

\title{Resolving Label Uncertainty with Implicit Posterior Models}

\author[1,6]{\href{mailto:esther_rolf@berkeley.edu}{Esther~Rolf$^{*\ }$}}
\author[2,6]{Nikolay~Malkin$^*$}
\author[3,6]{Alexandros~Graikos}
\author[4]{Ana~Jojic}
\author[5]{Caleb~Robinson}
\author[6]{Nebojsa~Jojic}

\affil[1]{
    University of California,
    Berkeley, CA, USA
}
\affil[2]{
    Mila and Universit\'e de Montr\'eal,
    Montreal, QC, Canada
}
\affil[3]{
    Stony Brook University,
    Stony Brook, NY, USA
}
\affil[4]{
    Paul G. Allen School of Computer Science and Engineering, University of Washington,
    Seattle, WA, USA
}
\affil[5]{
    Microsoft AI for Good, 
    Redmond, WA, USA
}
\affil[6]{
    Microsoft Research, 
    Redmond, WA, USA
}

\begin{document}
\maketitle

\begin{abstract}
We propose a method for jointly inferring labels across a collection of data samples, where each sample consists of an observation and a \emph{prior belief} about the label. By implicitly assuming the existence of a generative model for which a differentiable predictor is the posterior, we derive a training objective that allows learning under weak beliefs. This formulation unifies various machine learning settings; the weak beliefs can come in the form of noisy or incomplete labels, likelihoods given by a different prediction mechanism on auxiliary input, or common-sense priors reflecting knowledge about the structure of the problem at hand. We demonstrate the proposed algorithms on diverse problems: classification with negative training examples, learning from rankings, weakly and self-supervised aerial imagery segmentation, co-segmentation of video frames, and coarsely supervised text classification.
\end{abstract}

\section{Introduction}

In prediction problems, coarse and imprecise sources of input can provide rich information about labels.
Negative labels (what an instance is \emph{not}), rankings (which of two instances is larger), or coarse labels (aggregated by taxonomy or geography) give clues on what the ground truth label of an instance \emph{might} be, but not what it \emph{is} directly.
We consider a collection of data samples, indexed by $i$, consisting of observations (features) $x_i$ and corresponding sample-specific \emph{prior beliefs} about their latent label variables, $p_i(\h)$. This paper proposes algorithms to \textbf{resolve the uncertainty in these prior beliefs} by jointly inferring an assignment of target labels $\h_i$ and a model that predicts $\h_i$ given $x_i$.

Partial or aggregate annotations and auxiliary data sources are often more widely available and convenient to collect than ``ground-truth" or high-resolution labels, but they are not readily used by discriminative learners. Supervision from probabilistic targets can result in uncertain predictions (\S\ref{sec:background_and_approach}). Most approaches to resolve these uncertainties involve iterative generation of hard pseudolabels~\citep{zhang2021refining} or loss functions promoting low entropy of  predictions~\citep{nguyen2008classification,yu2016maximum,zou2020pseudoseg,yao2020ambiguous}. Typically, these approaches are application-specific \citep{han2014object,zheng2021Weakly,bao2021MRTA,li2021Change}. In many settings, fusing weak input data into a probability distribution over classes is a more natural  alternative to transforming the weak input into hard labels~\citep{mac2019presence}. Further connections and comparisons to prior work are made throughout this paper and synthesized in \S\ref{sec:related_work_extras} and \S\ref{sec:implicit_details}.

Our key modeling insight (\S\ref{sec:implicit_generative_model}) is to identify the output distribution of a discriminative model, a feed-forward neural network $q$, with an approximate posterior over latent variables in an \emph{generative} model of features, of which the given prior belief is a part. Bayesian reasoning about the generative model and its posterior makes it possible to learn the inference network \emph{without instantiating the full generative model}, while reaping the benefits of generative modeling: high certainty in the posterior under soft priors and rich opportunities to model structure in the prior beliefs. 

Prior beliefs about labels can arise from many sources (\S\ref{sec:priors}). We validate the effectiveness of our approach with experiments (\S\ref{sec:experiments}, \S\ref{sec:additional_exp}) on multiple domains and data modalities that highlight: prior beliefs as a natural way to fuse weak inputs, graceful degradation of performance with increasingly noisy or incomplete inputs, and comparison with explicitly generative modeling approaches.

\begin{figure*}[t]
    \centering
    \hspace{-4mm}
    \setlength{\fboxrule}{2mm}
    \definecolor{nicecolor}{rgb}{0.94,0.94,1.0}
    \definecolor{xi_color1}{rgb}{0,0,0}
    \definecolor{xi_color2}{rgb}{1,1,1}
    \fcolorbox{white}{white}{
    \hspace{-1em}
    \scalebox{1.35}{
    \includegraphics[width=0.105\textwidth,trim=0 160 0 0,clip]{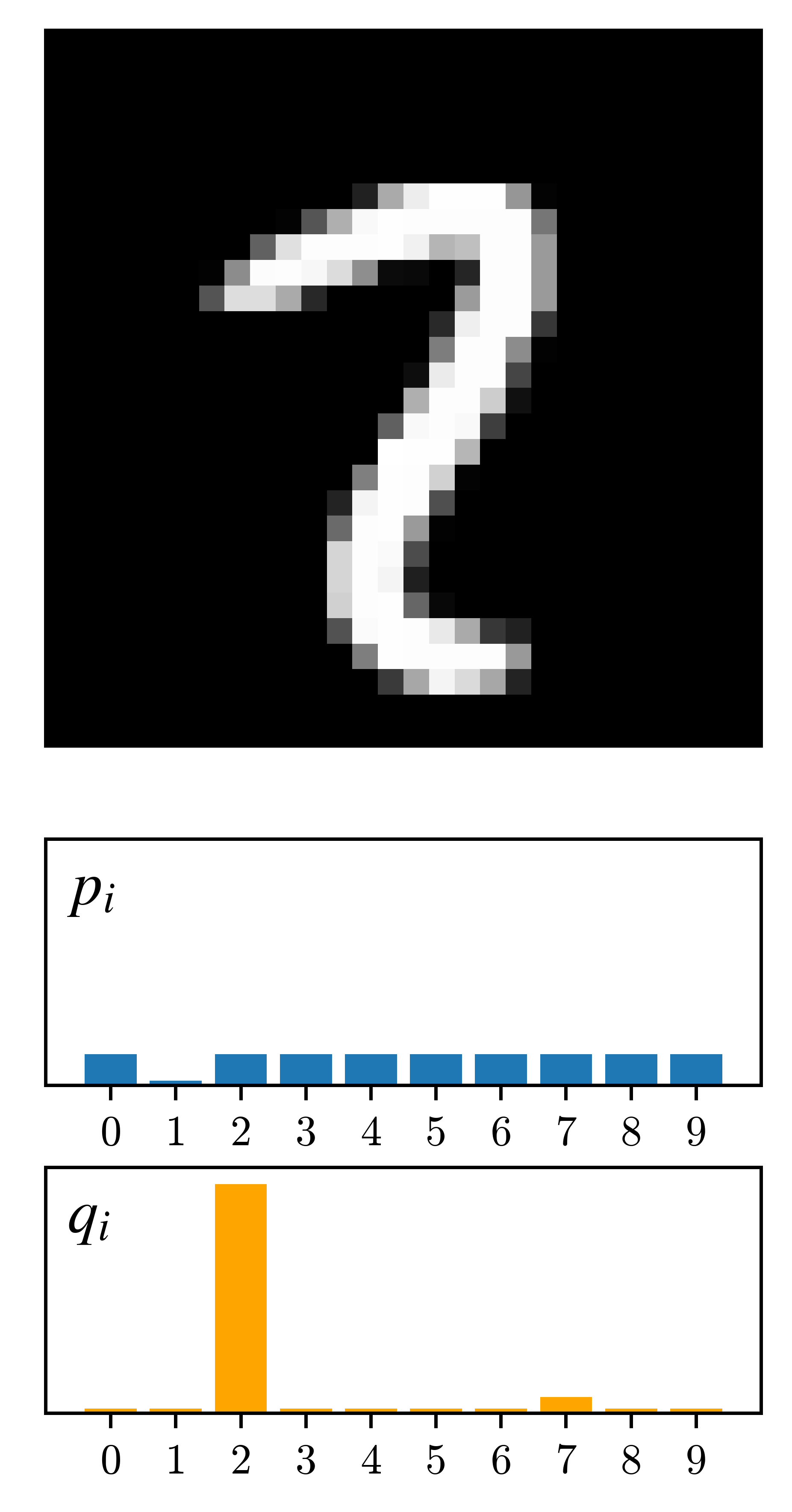}
    \hspace{-16.5mm}\raisebox{15mm}{\textcolor{xi_color2}{$x_i$}\hspace{13mm}}
    \includegraphics[width=0.125\textwidth,trim=0 0 0 180,clip]{figures/mnist4.png}
    
        \includegraphics[width=0.105\textwidth,trim=0 160 0 0,clip]{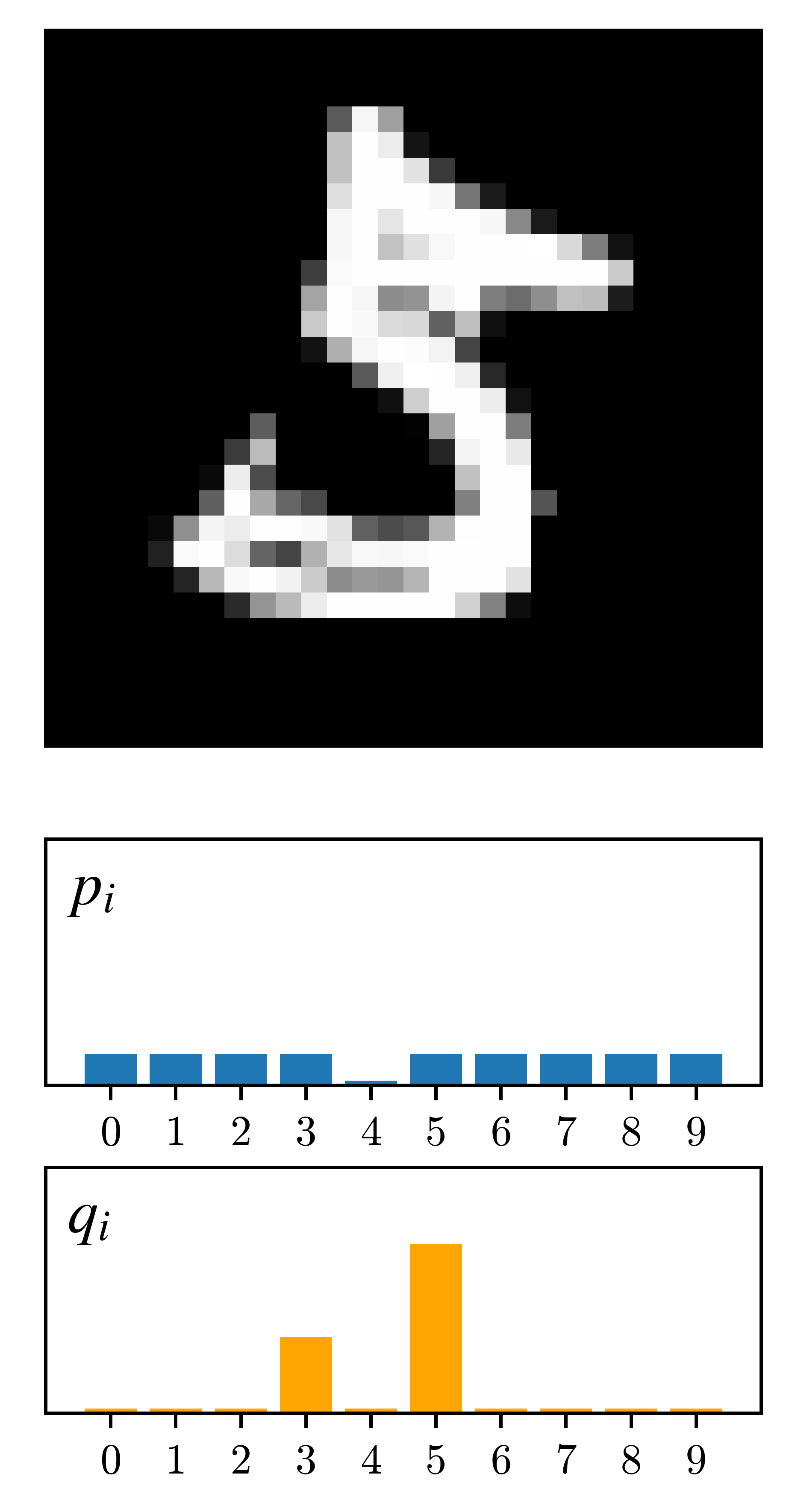}
    \hspace{-16.5mm}\raisebox{15mm}{\textcolor{xi_color2}{$x_i$}\hspace{13mm}}
    \includegraphics[width=0.125\textwidth,trim=0 0 0 180,clip]{figures/mnist2.png}
        \includegraphics[width=0.105\textwidth,trim=0 160 0 0,clip]{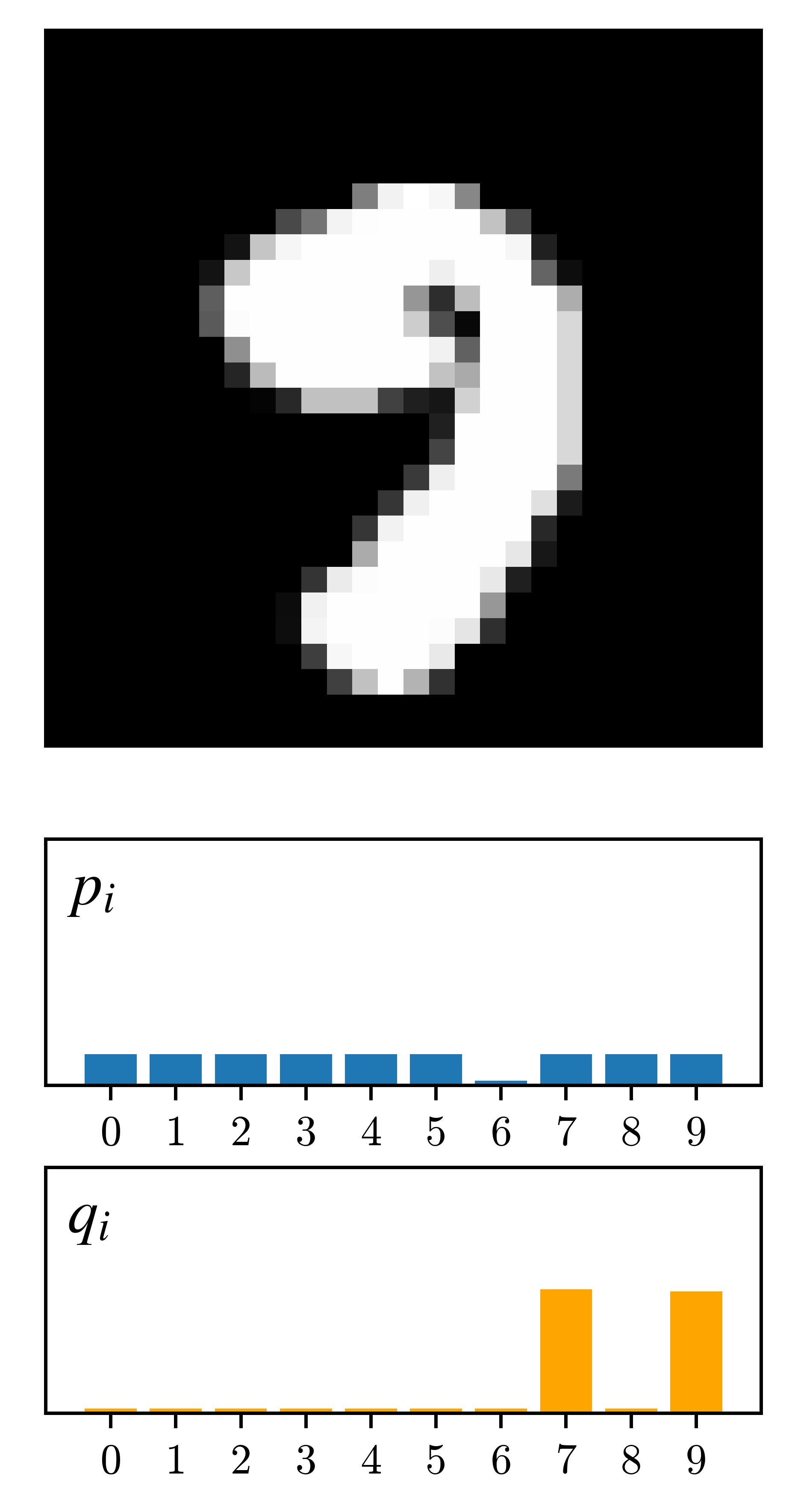}
    \hspace{-16.5mm}\raisebox{15mm}{\textcolor{xi_color2}{$x_i$}\hspace{13mm}}
    \includegraphics[width=0.125\textwidth,trim=0 0 0 180,clip]{figures/mnist1.png}
    }
    \hspace{-1em}
    }

    \begin{tabular}{@{}c@{\hspace{8pt}}c@{\hspace{8pt}}c@{}}
    \hline \\
    \includegraphics[width=0.31\textwidth]{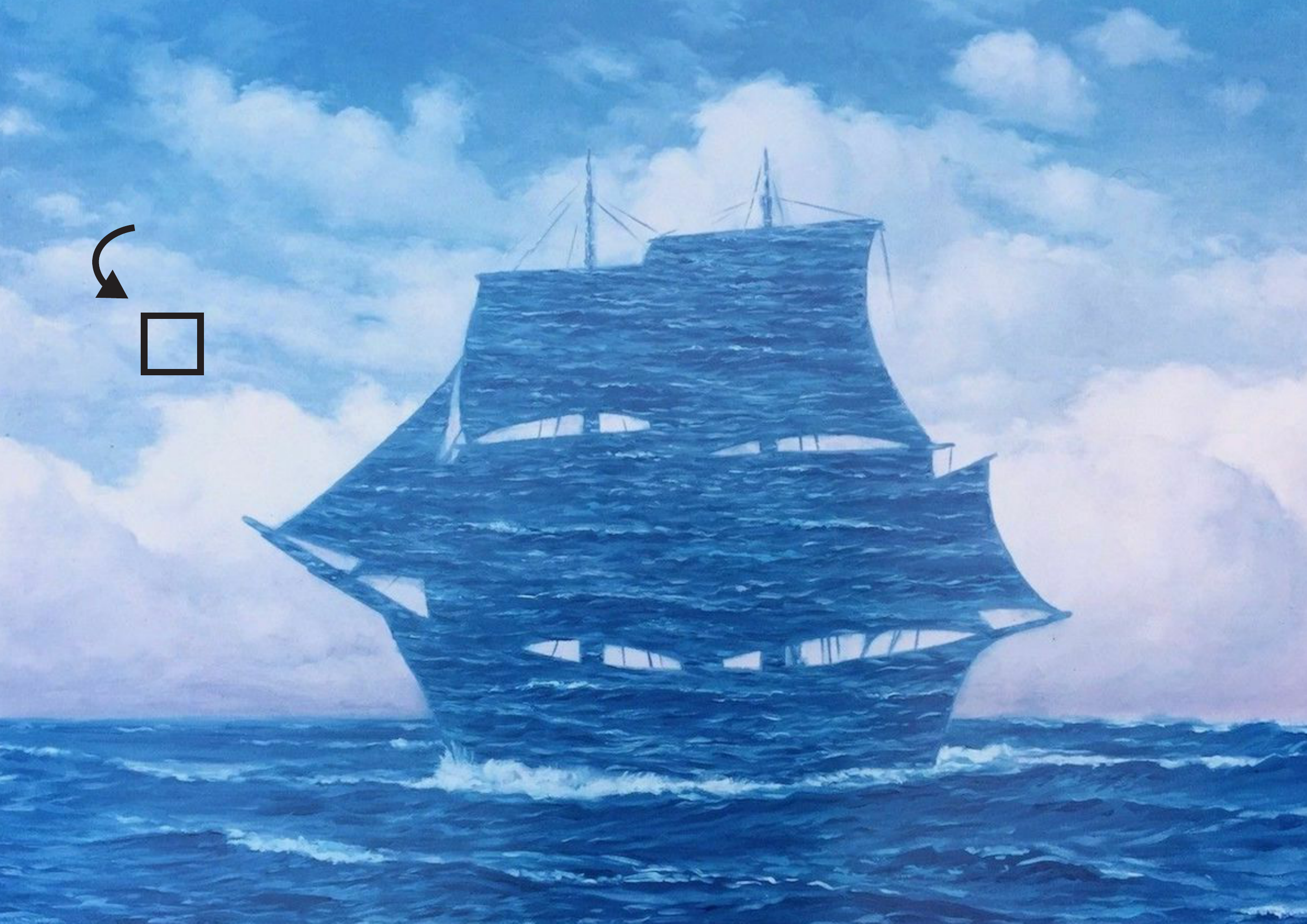} 
 
    \hspace{-48mm}\raisebox{27mm}{\textcolor{xi_color1}{\huge $x_i$}\hspace{45mm}}
    & \includegraphics[width=0.31\textwidth]{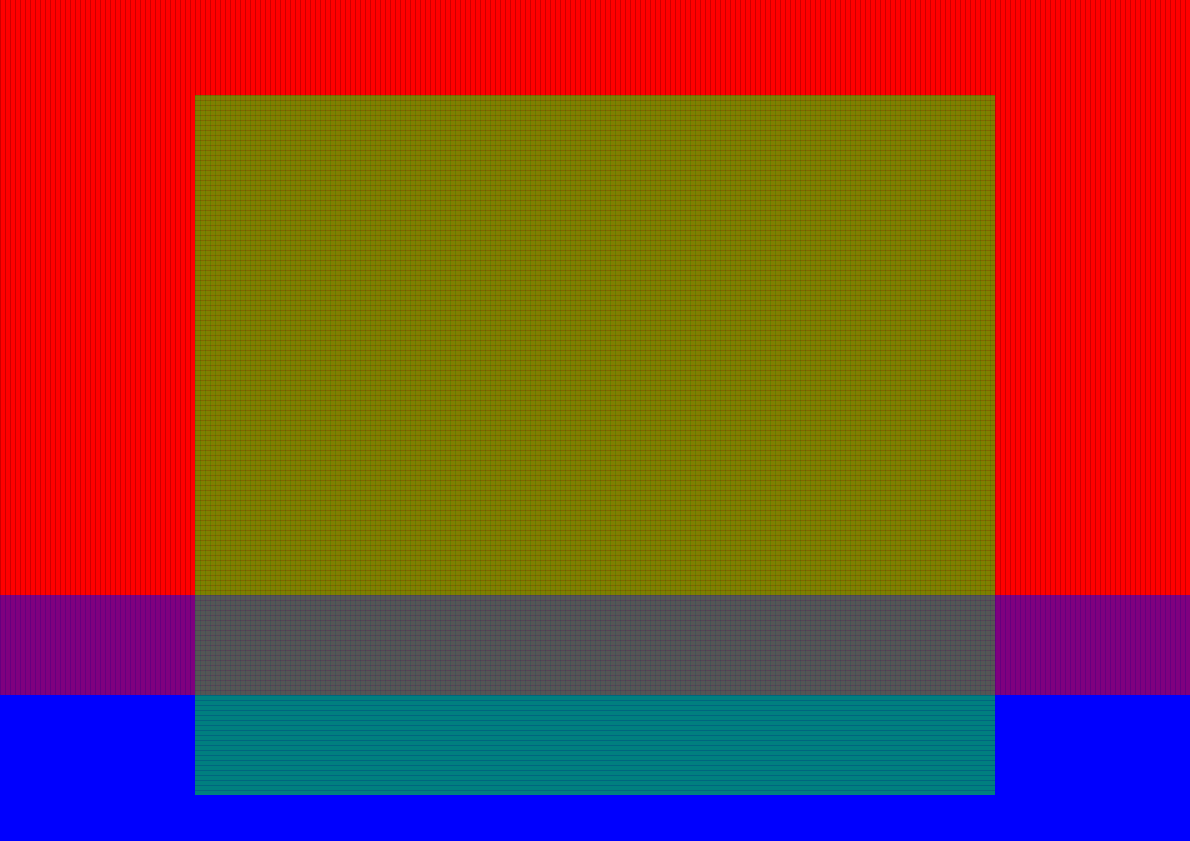}\hspace{3mm}
    & \includegraphics[width=0.31\textwidth]{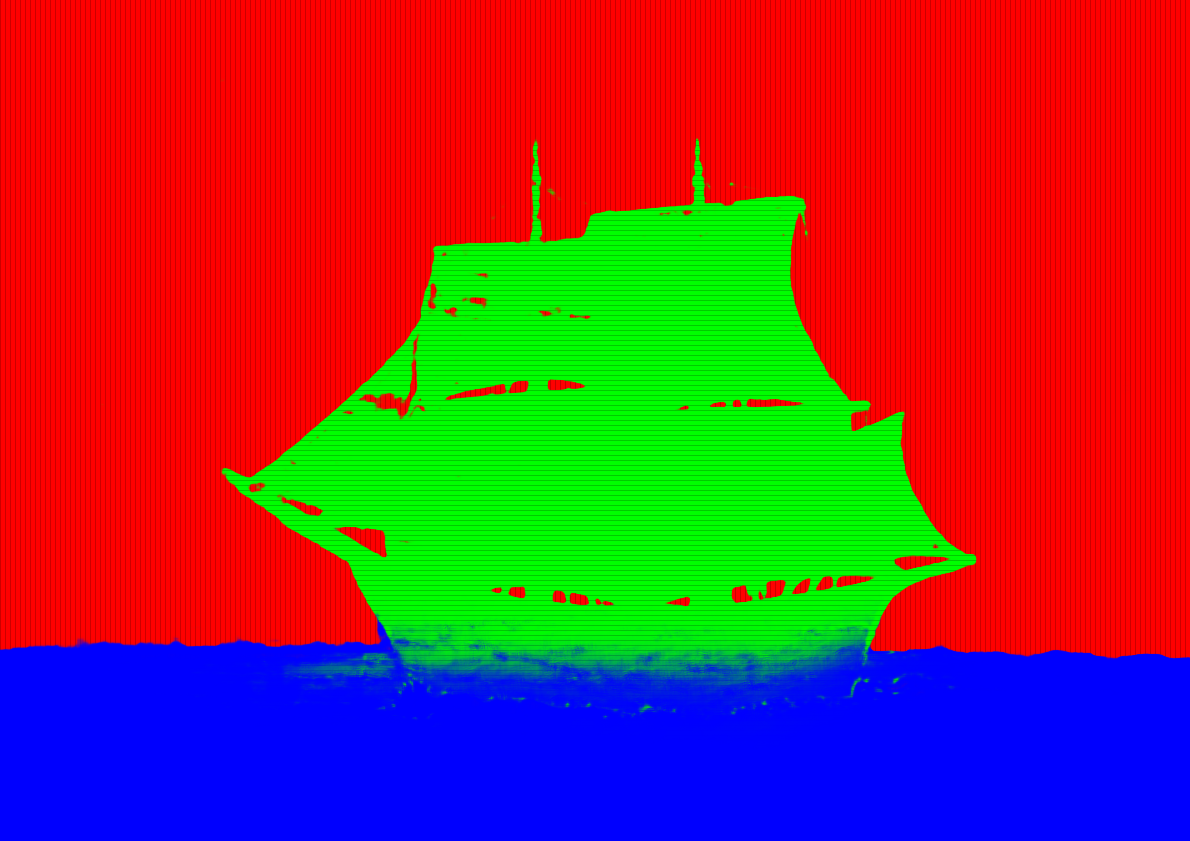}\\
    (a) $\{x_i\}$: \textit{Le s\'educteur}, Ren\'e Magritte
    & (b) $p_i(\ell)$: \textit{Boat Prior}, anonymous artist
    & (c) $q_i(\ell)$: Inferred segmentation
    \end{tabular}
    \caption{\textbf{Above:} Inference of latent MNIST digit classes with negative label supervision using a small CNN trained on the \textbf{RQ} criterion (\S\ref{sec:implicit_generative_model}).  \textbf{Below:} (a) Joint inference of latent pixel classes in an image. (b) Prior beliefs $p_i(\ell)$ over three classes -- sky (red), boat (green), water (blue) -- are manually set. (c) A small CNN trained on $(x_i, p_i(\ell))_i$ infers the posterior classes.}
    \label{fig:examples}
\end{figure*}

\section{Background and approach}
\label{sec:background_and_approach}

\paragraph{Two motivating examples.}

Two illustrative examples are shown in Fig.~\ref{fig:examples}. In the first example, the $x_i$ are 784-dimensional vectors representing 28$\times$28 MNIST digits. We aim to infer the digit classes $\h_i \in \{0,1,...,9\}$ for all images in the given collection based on data in which we are given just one \emph{negative} label per sample, i.e., the prior beliefs $p_i(\h)$ (top row) are uniform over all classes except for one incorrect class. The procedure described in this paper produces inferred distributions over labels (bottom row) that are usually peaky and place the maximum at the correct digit 97\% of the time (see Fig.~\ref{fig:mnist_cifar} and \S\ref{sec:negative_labels}).

In the second example, the observations $\{x_i\}_{i \in \textrm{pixels}}$ are image patches centered around each pixel coordinate $i$ in a Surrealist painting, with patch size ($11\times11$) equal to the receptive field of a 5-layer convolutional neural network used in our inference procedure. The prior beliefs $p_i(\h)$ are distributions over 3 classes (sky, boat, water) depending on the coordinate $i$. 
The joint inference of all labels in this image yields a feasible segmentation despite the high similarity in colors and textures (see \S\ref{sec:seducer} for more details).

These examples illustrate the problem of training on weak beliefs, which is often encountered in some form in machine learning. Weak supervision, semi-supervised learning, domain transfer, and integration of modalities are all settings where coarse, partial, or inexact sources of data can provide rich information about the state of a prediction instance, though not always a ``ground truth'' label for each instance. 
An inference technique that uses
weak beliefs as the sole source of supervision
needs to estimate statistical links between observations $x_i$ and corresponding latents $\h_i$. 
These links should simultaneously be highly confident (i.e., lead to low entropy in the posterior distributions) and explain the varying prior beliefs, which typically have low confidence (high entropy in the prior distributions).

\paragraph{Supervised learning on prior beliefs.}

Supervised learning models, including many neural nets, are typically trained to minimize the cross-entropy $-\sum_i\sum_\h p_i^d(\h)\log q_i(\h)$  between a ``hard" distribution over labels with $p_i^d(\h) \in \{0,1\} $ and the distribution $q_i(\h)=q(\h| x_i;\theta)$ output by a predictor $q$ using data features $x_i$.
This is equivalent to minimizing the KL divergence $\sum_i \KL(p^d_i \| q_i)$,
minimized when the two distributions $p^d_i(\h)$ and $q_i(\h)$ are equal. 
Thus, when  $p^d_i(\h)$ is a ``softer" prior over latent labels, $p_i(\h)$, 
the trained model $q$ will reflect this, and also be highly uncertain. 

Transforming soft labels into hard training targets, (e.g. training on $\mathbbm{1}[\h=\argmax_\h p^d_i(\h)]$
), can introduce the opposite bias. In these cases, the cost would be minimized by predictions with zero entropy, but learning such a prediction function faces difficulty with overconfident labels which are often wrong, and the possibility that certain labels often receive substantial weight in the prior, but never the maximum. These issues are illustrated in Fig.~\ref{fig:sammamish_loss_comparison}.

\paragraph{Generative modeling resolves the prior's uncertainty.}

The approach to classification problems through \emph{generative} modeling, instead of targeting the conditional probability of latents given the data features, assumes that there is a forward (generative) distribution $p(x_i|\h)$ and optimizes the log-likelihood of the observed features, $\sum_i \log(x_i)=\sum_i \log \sum_\h p(x_i|\h)p_i(\h)$, with respect to the parameters of that distribution. The posterior under the model $q(\h|x_i)\propto p(x_i|\h)p_i(\h)$ is then used to infer latent labels for individual data points \citep{seeger}. The generative modeling approach does not suffer from uncertainty in the posterior distribution over latents given the input features, even when the priors $p_i(\h)$ are soft. (Recall that the posterior distributions in a mixture of high-dimensional Gaussians are often peaky even when the priors are flat.) 

However, expressive generative models are typically harder and more expensive to train compared to supervised neural networks, as they often require sampling (e.g., sampling of the posterior in variational auto-encoders \cite[VAEs;][]{kingma2014auto} and sampling of the generator in GANs \citep{gan}). Furthermore, the modeling often requires doubling of parameters to express both the forward (generative) model \emph{and} the reverse (posterior) model. And, in case of GANs, the learning algorithms may not even cover all modes in the data, which would prevent joint inference for \emph{all} data points. 
(See \S\ref{sec:implicit_details} for further discussion.)

\subsection{Optimizing implicit posterior models}
\label{sec:implicit_generative_model}

Suppose 
that there exists a generative model $p(x|\h)$ of observed features conditioned on latent labels. Optimization of the log-likelihood of observed features, $\sum_i\log p(x_i)=\sum_i\log(\sum_\h p(x_i|\h)p_i(\h))$, can be achieved by introducing a variational posterior distribution $q(\h|x_i)$ over the latent variable for each instance $x_i$ and minimizing the free energy (a negated evidence lower bound (ELBO)), 
defined as
\begin{equation}
    - \sum_i\sum_\h q(\h|x_i)\log\frac{p(x_i|\h)p_i(\h)}{q(\h|x_i)}
    \geq-\sum_i \log p(x_i).
    \label{eq:free_eng}
\end{equation}
Minimizing the free energy involves estimating both the forward distributions $p(x_i|\h)$ and the posteriors $q(\h|x_i)$. 

One could parametrize both $p(x|\h)$ and $q(\h|x)$ as functions $p(x|\h,\theta_p)$ and $q(\h|x,\theta_q)$ using neural networks, as done by VAEs (although VAEs use continuous latent variables $\ell$ and do not involve sample-specific priors).
However, in our algorithms, we only parametrize $q(\ell|x; \theta)$  as a neural network taking input $x$ and producing a distribution over $\ell$. The generative conditional $p(x_i|\ell)$ is defined only on data points $x_i$ and is calculated by minimizing 
\eqref{eq:free_eng} for fixed $q(\h|x)$,
subject to the constraint that
$\sum_{i} p(x_i|\h)=1$ for all $\h$.\footnote{\newtext{This constraint allows nonzero likelihood under the generative model only for the observed data points $x_i$. The derivation still holds if the assumption is relaxed to $\sum_{i} p(x_i|\h) \leq 1$. Subject to this weaker condition, the minimum of free energy is achieved on the boundary of the constraint domain, when $\sum_{i} p(x_i|\h) = 1$.}} The optimum is achieved by:
\begin{equation}
    p(x_i|\h)=a_{i,\ell}=\frac{q(\h|x_i)}{\sum_j q(\h|x_j)}~.
    \label{eq:aux_p}
\end{equation}
Here the generative conditional $p(x|\ell)$ is not fully specified for all values $x$. Rather, it is represented as a matrix of numbers $a_{i,\ell}$
describing the conditional probabilities of different values of $x_i$  given different latent labels $\ell$. The probabilities $p(x_i|\ell)$  are greater for the data points $i$ for which $q(\h|x_i)$ is more certain, 
relative to how popular assignment to class $\h$ is across data points (denominator in \eqref{eq:aux_p}).

In our formulation, $q$ plays the role of a variational posterior, but \emph{implicitly}, in a generative model consisting of varying instance-specific priors $p_i(\ell)$ and a complex conditional $p(x|\ell)$ that is never fully estimated, but is instead maximized for the data points studied. The full link between $x$ and $\h$ 
is left entirely to the neural network $q$ to capture explicitly. 

In variational methods, the free energy (\ref{eq:free_eng}) is usually rewritten as $\sum_i {\rm KL}(q(\ell|x_i)\|r_i(\ell)))-\log p(x_i)$, where $r$ is the posterior of the forward model,
i.e., for the  points $i$, $r_i(\ell) \propto p_i(\ell)p(x_i|\ell)$. The minimization of free energy then reduces to minimizing the KL divergence between $r$ and $q$.

\begin{figure}[t]
    \centering
    \includegraphics[width=0.48\textwidth]{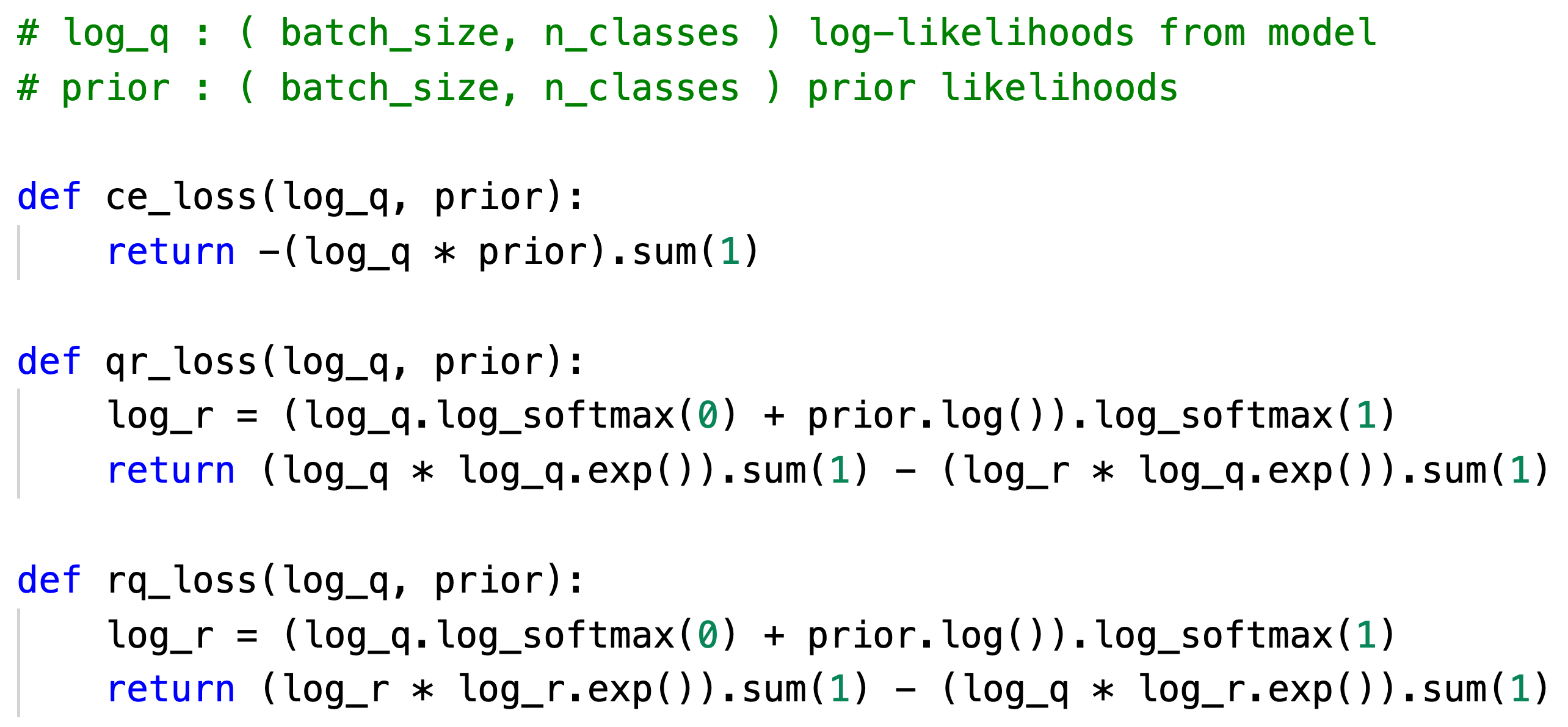}
    \caption{Cross-entropy and  implicit \textbf{QR} / \textbf{RQ} losses in PyTorch. Here the normalization in (\ref{eq:aux_p}) is done within batches.
    }
    \label{fig:qr_losses_torch}
\end{figure}

We define $q_i(\h)=q(\h|x_i;\theta)$. After our reduction of $p(x_i|\ell)$ to the auxiliary matrix in \eqref{eq:aux_p}, the posterior $r$ has the form  
\begin{equation}
    r_i(\h) = c_i \cdot p_i(\h)  p(x_i|\h) = c_i \frac{p_i(\h)q_i(\h)}{\sum_j q_j(\h)}~,
    \label{eq:true_post}
\end{equation}
where $c_i$ are scalars making $\sum_\ell r_i(\h) = 1$. 
For each instance $i$ we have two outputs: the direct model outputs of the variational posterior $q_i$  and their \emph{implied posterior} $r_i$, which is computed by multiplying the renormalized model outputs with the provided prior at each instance as in \eqref{eq:true_post}. 
Using these two outputs, we can optimize a single set of model parameters $\theta$ to minimize \eqref{eq:free_eng}:
\begin{align}
\label{eq:KL_written_out}
   & \min_\theta \sum_i \text{KL}( q_i \| r_i) = \\[-4em]
&  \min_\theta \sum_i \text{KL}\bigg(  
\underbracket{\Big(q(\ell|x_i;\theta)\Big)_\ell }_{
\substack{\text{model output } \\ \text{with input $x_i$}}}
\Big\| 
\Big(
c_i \cdot\hspace{-2mm}
\overbracket{p_i(\ell)
\vphantom{\dfrac{q(|)}{\sum_jq(|)}} 
}^{
\substack{\text{per-} \\ \text{\phantom{p}instance\phantom{p}} \\ \text{priors}}}
\overbracket{\frac{q(\ell | x_i; \theta)}{\sum_j q(\ell | x_j; \theta)}}^{
\substack{
\text{model output} \\ \text{\phantom{p}normalized\phantom{p}} \\ \text{per-class} \\ \text{as in Eq.~\eqref{eq:aux_p}}
}}
\Big)_\ell 
 \bigg) \nonumber ~.
\end{align}
While \eqref{eq:KL_written_out} optimizes the free energy \eqref{eq:free_eng}  by minimizing $\textrm{KL}(q_i \| r_i)$, minimizing $\textrm{KL}(r_i \| q_i)$ would also find solutions for which the direct model and its implied posterior are close.
We propose to optimize either of these two objectives with respect to the model parameters $\theta$ by gradient steps. We iterate over data instances $x_i$ with priors $p_i(\h)$:
\begin{enumerate}[label*=(\arabic*)]
    \item Calculate the distributions $r_i$ in terms of $q_i$ as in (\ref{eq:true_post}).
    \item {Update the parameters of $q$ with a gradient step:  \\
    \mbox{$\bullet$ Option \textbf{QR}: $\theta \leftarrow \theta -\eta \nabla_\theta \sum_i \KL(q_i \| r_i)$.}\\
    \mbox{$\bullet$ Option \textbf{RQ}: $\theta \leftarrow \theta -\eta \nabla_\theta \sum_i \KL(r_i \| q_i)$.}
    }
\end{enumerate}
Gradients of the objectives are propagated to the expression of $r_i$ through $q_i$ (see (\ref{eq:KL_written_out}) and Fig.~\ref{fig:qr_losses_torch}). 
Both losses have a stable point when $q_i=r_i$, and \textbf{RQ} reduces to the cross-entropy loss in the case of priors which put all mass on one label (e.g. $p_i(\h) = \mathbbm{1}[\h = \ell_i]$). A discussion of the relative benefits and limitations of the \textbf{QR} and \textbf{RQ} losses is given in \S\ref{sec:practical_considerations}, along with practical considerations for implementation.

By defining the conditional model $p(x|\ell)$ as an auxiliary matrix of probabilities $a_{i,\ell}$ that 
is fit to the
reverse model $q$ during learning, we avoid parametrizing both directions of the link $\ell-x$ with highly nonlinear models.\footnote
{Note that the use of an auxiliary matrix $a_{i,\ell}$ is also found in expectation-maximization \citep[EM;][]{dempster1977em}, which also minimizes the free energy. However, in EM, it is the variational posterior $q(\ell|x_i)$ which is optimized as a matrix of numbers $a_{i,\ell}$ only on data points, while the \emph{generative} model $p$ is fully parametrized (see Table~\ref{tab:em_vae_comparison}).} We thus manage to keep the problem in the realm of training a single feed-forward network $q$ as a predictor of variables $\ell$, but in a way that treats the instance-specific priors $p_i(\ell)$ as they would be in generative modeling.

Next, we discuss the consequences of implicitly modeling the generative model $p$ with an auxiliary distribution.  
Option \textbf{QR} uses the KL distance in the direction it appears in (\ref{eq:free_eng}) and thus guarantees continual improvements in free energy and convergence to a local minimum (with the exception for the effects of stochasticity in minibatch sampling).
Substituting $r_i$ from (\ref{eq:true_post}), 
the free energy (\ref{eq:free_eng}) becomes:
\begin{equation}
    F=\sum_{i,\h} q_i(\h) \log \left(\sum_{j} q_j(\h)\right)  - 
    \sum_{i,\h} q_i(\h)\log \left(p_i(\h) \right)
    \label{eq:QR}
\end{equation}
This criterion does not encourage entropy of individual $q_i$ distributions, but of their \emph{average}. The second term alone would be minimized if $q$ could put all the mass on $\argmax_\h p_i(\h)$ for each data point, but the first term promotes diversity in assignment of latents (labels) $\h$ across the entire dataset. Thus a network $q$ can optimize 
(\ref{eq:QR})
if it makes different confident predictions for different data points. 

To illustrate this, 
consider the case when all data points have the \emph{same} prior, $p_i(\h)=p(\h)$. Then
(\ref{eq:QR}) and the \textbf{RQ} objective
are minimized when $\frac{1}{N}\sum_i q_i(\h) = p(\h)$. This can be achieved when $q$ learns a constant distribution $q(\h|x_i; \theta)=p(\h)$. But both objectives are also minimized if $q$ predicts only a single label for each data point with high certainty, but it varies in predictions so that the counts of label predictions match the prior. 

As demonstrated in Fig.~\ref{fig:examples} and in our experiments, avoiding degenerate solutions is not hard. We attribute this to two factors. First, the situations of interest typically involve uncertain, but varying priors $p_i(\h)$ which break symmetries that could lead to predictors ignoring the data features $x_i$. Second, the neural networks used to model $q$, and their training algorithms, come with their own constraints and inductive biases. 
In fact, as discussed in \S\ref{sec:priors} and \S\ref{sec:aaai_regret}, even unsupervised clustering is possible with suitably chosen priors that break symmetry, allowing this approach to be used for self-supervised training. See also \S\ref{sec:related_work_extras}, \S\ref{sec:implicit_details} for more on relationships with other approaches.

\newtext{
In practice, the normalization in (\ref{eq:aux_p}) is done within batches,  rather than across the entire dataset (see Fig.~\ref{fig:qr_losses_torch}). This may be sufficient if batches are large and representative of the diversity in the data. 
Experiments in \S\ref{sec:practical_considerations} examine the effect of batch size  on performance. While our algorithm is relatively tolerant to moderate batch sizes, performance degrades for small batches, in particular when batches are likely to be missing samples of some classes.
Addressing this problem in more general settings is an interesting subject for future work. When intra-batch diversity is an issue, the denominator in  (\ref{eq:true_post}) may need to be updated in an online fashion or even replaced by a learned parametric estimate. 
}

\section{Sources of label priors}
\label{sec:priors}

Having detailed our approach for learning from prior beliefs as weak supervision in \S\ref{sec:background_and_approach}, 
we now describe a range of machine learning settings where priors $p_i(\h)$ emerge. 
All of these settings are illustrated by experiments in \S\ref{sec:experiments} and \S\ref{sec:additional_exp}.

\paragraph{Negative or partial labels (\S\ref{sec:negative_labels}).}

When we are given a set of equally possible labels $L_i$ for each point data point $i$, instead of a single label $\h_i$, then we set the prior $p_i(\h)=\frac{1}{|L_i|}\mathbbm{1}[\h \in L_i]$. An extreme example is when one negative label is given and hence can be ``ruled out" (Fig.~\ref{fig:examples}).

\paragraph{Joint labels and learning from rankings (\S\ref{sec:ranking}).}

Priors may also come in the form of joint distributions over labels of multiple instances. For example, \textit{ranking supervision} -- the knowledge of which example in a pair is greater with respect to an ordering of the labels -- gives prior beliefs about \emph{pairs} of labels. Suppose our data is organized into pairs of images of digits $T_j=\{x_{j,1},x_{j,2}\}$, and for each pair we are told which image represents the digit (0--9) which is greater (or equal). This sets a prior $p(\h_1,\h_2)$ over pairs of labels in each pair, represented by either an upper or a lower triangular matrix, depending on which digit in the pair is known to be greater, with all nonzero entries equal to $\nicefrac{1}{55}$.

We assume the underlying generative model has the form $p(x_1,x_2|\h_1,\h_2)=p(x_1|\h_1)p(x_2|\h_2)$. We aim to fit its posterior model $q(\h|x;\theta)$. For each pair $T_j$, we have two outputs of the predictor network, $q(\h_1|x_{j,1})$ and $q(\h_2|x_{j,2})$, for the two images in the pair. The joint posterior under the generative model is
\begin{align}
    r_j(\h_1,\h_2) \propto p(\h_1,\h_2)p(x_{j,1}|\h_1)p(x_{j,2}|\h_2) \propto\nonumber\\\propto \frac{ p(\h_1,\h_2)q(\h_1|x_{j,1})q(\h_2|x_{j,2})}{ \sum_{j} q(\h_1|x_{j,1}) \sum_{j} q(\h_2|x_{j,2})},
    \label{eq:pair_posterior}
\end{align}
and we can now use \textbf{QR} or \textbf{RQ} loss to fit $q(\h_1|x_{j,1})$ to the marginal $r_j(\h_1)$ and $q(\h_2|x_{j,2})$ to $ r_{j}(\h_2)$.

\paragraph{Coarse data in weakly supervised segmentation (\S\ref{sec:chesapeake_experiments}, \S\ref{sec:lymphocytes}, \S\ref{sec:seducer}).}

We often have side information $z$ associated to each instance $i$ that allows setting the priors $p_i(\h)=p(\h|z_i)$ for each point directly by hand. These include situations when we have beliefs about labels for different points, as in the \emph{Seducer} example (Fig.~\ref{fig:examples}). 
Interesting weak supervision settings also arise in remote sensing (\S\ref{sec:chesapeake_experiments}) and medical pathology (\S\ref{sec:lymphocytes}) applications. For example, in a task of segmenting aerial imagery into land cover classes, we often have coarse labels $c$ associated to large \emph{blocks} of pixels, but not the target labels $\ell$ for individual pixels. If the conditional $p(\ell|c)$ is known, it sets a belief about the high-resolution labels $\ell$ for pixels in a block of class $c$.

\paragraph{Fusing models and data sources (\S\ref{sec:enviroatlas_experiments}, S\ref{sec:nlp}).}

Auxiliary information $z$ may not always come with a known correspondence $p(\h|z)$. In the land cover mapping problem, auxiliary information includes different modalities and resolutions (road maps, sparse point labels, etc.). While these sources can be fused into a prior by hand-coded rules, the prior may be more accurately set as the output of a model $p(\h|z_i)$ \emph{trained} on a separate dataset of points $(\h_i,z_i)$. This is especially useful when the data $x_i$ (imagery) is informative about the latents $\h_i$ but is prone to domain shift problems, while the auxiliary data $z_i$ does not suffer from domain shift issues but is not sufficient on its own to predict the labels. In a text classification problem, $z_i$ might be the encoding of text $x_i$ by a pretrained language model, and $p(\h|z_i)$ a noisy distribution over labels given by their likelihoods under the language model as continuations of a prompt.

\paragraph{Priors for self-supervision (\S\ref{sec:aaai_regret}).}
In \S\ref{sec:implicit_generative_model} we discussed the pitfalls of using a constant prior $p_i(\h)=p(\h)$ for all data points in training models under the \textbf{QR} loss as a potential method for unsupervised clustering. However, in \S\ref{sec:aaai_regret} we give an example of \emph{joint} learning of the posterior model $q$ and an energy model (Markov random field) on the latent labels $\h_i$ that expresses local structure of labels in an image. This results in unsupervised clusterings that are useful in downstream segmentation tasks. Such an approach is an example of a benefit of generative modeling -- the possibility of learning of a parametrized distribution over latents -- being inherited by implicit posterior models.

\paragraph{Priors with latent structure (\S\ref{sec:tracking}).}
  
Implicit posterior modeling allows building hierarchical latent structure into the prior (another benefit of classical generative models), as we demonstrate in \S\ref{sec:tracking} on a video segmentation task. The prior is an admixture of possible segmentations with a structure similar to \citet{jojic2009stel}, but using a set of mask proposals $p(\h_i|m)$ from a Mask R-CNN model \citep{he2017maskrcnn}, indexed by a latent $m$. The prior is $p_i(\h)=\sum_m p(\h_i|m)p(m)$, where $p(m)$, a probabilistic selection of the masks for the admixture in the given frame, is estimated by minimizing the free energy.

\section{Experiments}
\label{sec:experiments}

\newtext{
The experiments in this section and in \S\ref{sec:additional_exp} cover a variety of domains, illustrating the sources of label priors listed in \S\ref{sec:priors}. 
The experimental baselines are chosen to reflect the different goals of each experiment. 
Experiments on classification with negative training examples (\S\ref{sec:negative_labels}) 
and learning from rankings (\S\ref{sec:ranking})
serve to illustrate how our algorithm works in different conditions.  
For experiments on label super-resolution in image segmentation (\S\ref{sec:chesapeake_experiments}, \S\ref{sec:enviroatlas_experiments}, \S\ref{sec:aaai_regret}) 
and text classification (\S\ref{sec:nlp}), self-supervision for image clustering (\S\ref{sec:lymphocytes}), and video segmentation (\S\ref{sec:tracking}), baseline methods provide a comparison by which to benchmark performance, showing that we are reaching or close to state-of-the-art accuracy across these domains with a unified approach.
 }

\subsection{Partial labels in MNIST and CIFAR-10}
\label{sec:negative_labels}

\begin{figure}[t!]
    \centering
    \includegraphics[width=0.45\textwidth]{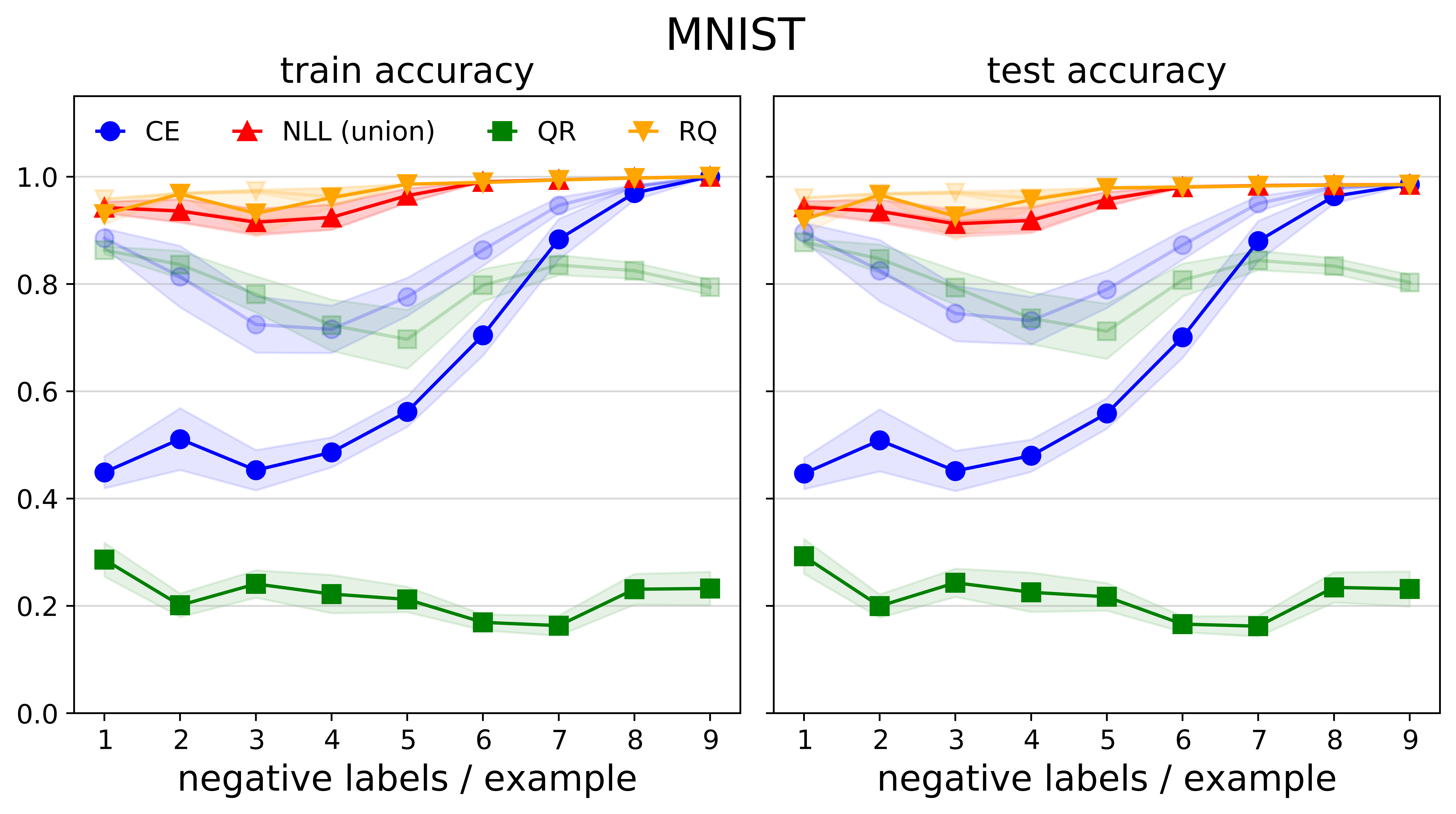}
    \\
    \includegraphics[width=0.45\textwidth]{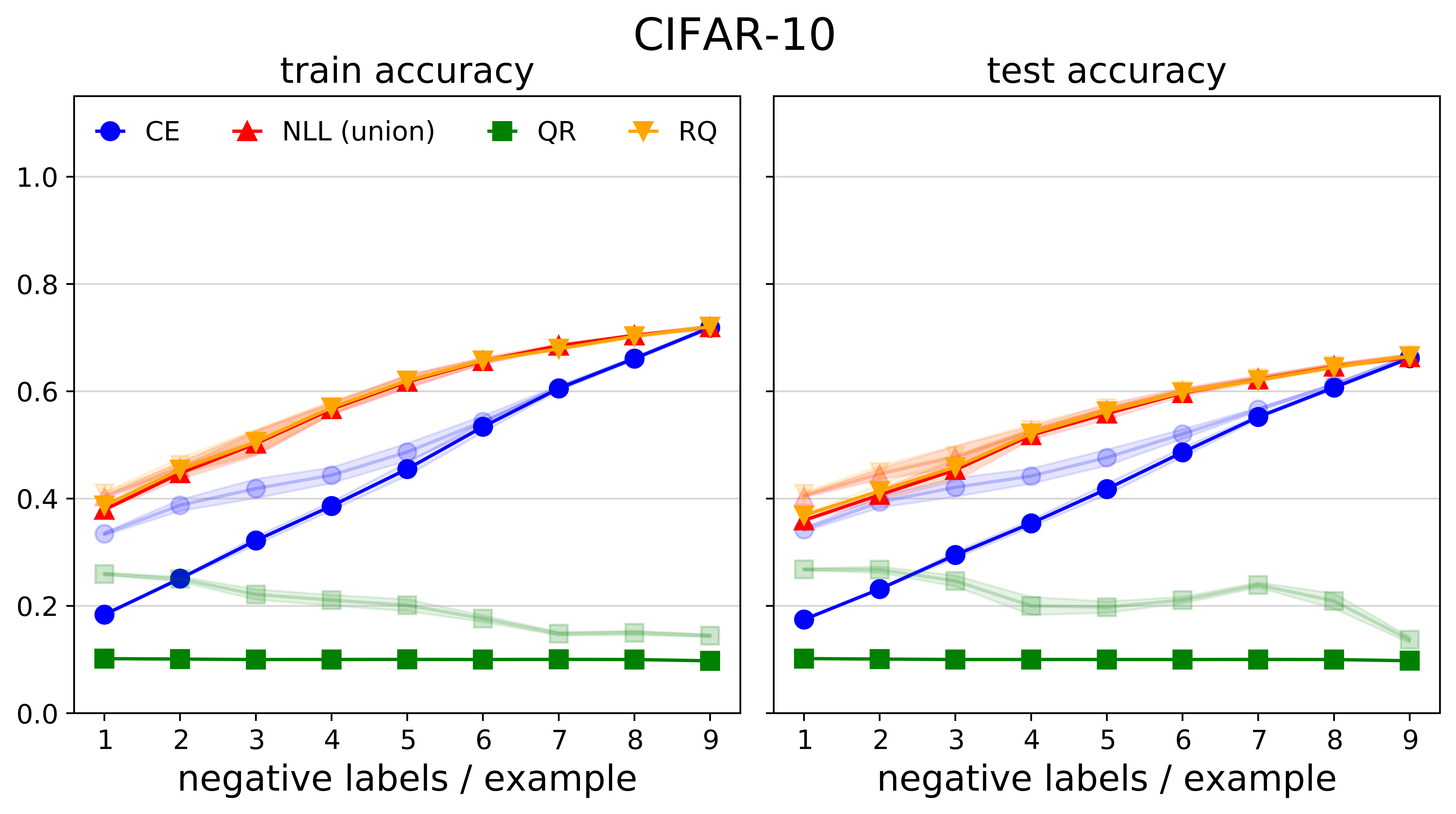}
    \\
    \caption{Accuracies of MNIST and CIFAR-10 classifiers trained with varying numbers of negative labels per example; the lighter variant of each color and marker shows the peak accuracy over 300 training epochs. (Average of 10 runs with standard error region.)}
    \label{fig:mnist_cifar}
\end{figure}

In this experiment, we compare algorithms for learning with partial labels on two 10-class image classification datasets, MNIST and CIFAR-10. 
To each training example $x_i$, we randomly assign a set $N_i$ of $k$ negative labels, chosen from the 9 labels distinct from the ground truth. The prior $p_i(\h)$ is set to be uniform over $\h\notin N_i$ and 0 for $\h\in N_i$. We vary $k$ from 1 (one negative label per example) to 9 (one-hot prior, full supervision). The data of $k$ negative labels carries $-\log_2(1-k/10)$ bits of label information; if $k=1$, $22\times$ less label information than in the fully supervised setting.

For both datasets, the base model $q$ is taken to be a small convolutional network, with four layers of ReLU-activated $3\times3$ convolutions with stride 2 and a linear map to the 10 output logits ($\sim$33k learnable parameters for MNIST, $\sim$34k for CIFAR-10). We experiment with four training losses:\\
$\bullet\;$ \textbf{CE:} cross-entropy between predictions $q(\h|x_i;\theta)$ and the prior $p_i(\h)$.\\
$\bullet\;$ \textbf{NLL (union):} negative logarithm of the sum of likelihoods assigned by $q$ to labels in $\h\notin N_i$, or, equivalently, $\log\sum_\h p_i(\h)q(\h|x_i;\theta)$, as done, e.g., by \citet{jin2002learning,nlnl}.\\
$\bullet\;$ The \textbf{QR} and \textbf{RQ} losses defined in \S\ref{sec:implicit_generative_model}.

The \textbf{CE}, \textbf{NLL (union)}, and \textbf{RQ} loss objectives are equivalent when $k=9$. 
The \textbf{RQ} and \textbf{NLL (union)} losses are equivalent when $\sum_iq_i(\h)$ is uniform over $\h$ \newtext{(see derivation in \S\ref{sec:related_work_extras})}, which approximately holds after a sufficient number of training epochs.

All models are trained for 300 epochs on batches of 256 images with the Adam optimizer \citep{kingma2014adam} and a learning rate of $10^{-4}$. After each epoch, we compute the accuracy of the predictor $q$ on the ground truth labels in the train and test sets. Fig.~\ref{fig:mnist_cifar} shows the final train and test set accuracies, as well as the maximum accuracies achieved at any epoch. Reported results are averaged over 10 choices of partial label sets and random initializations. 

Models trained on \textbf{RQ} loss perform best, with the greatest benefit over \textbf{CE} seen for very few negative labels. 
\newtext{This reinforces the claim in \S\ref{sec:background_and_approach} that optimizing the \textbf{CE} loss results in uncertain predictions when the priors are highly ambiguous}.
As expected, the performance of \textbf{RQ} and \textbf{NLL (union)} is very similar across $k$. We hypothesize that the small advantage of \textbf{RQ} over \textbf{NLL (union)} loss can be attributed to regularization in early training. Meanwhile, \textbf{QR} performs as well as \textbf{CE} for very uncertain priors at the peak epoch (light curves), but its predictions degenerate -- usually toward uniform predictions -- with longer training.

\subsection{Multiple-instance supervision: Learning from ranks}
\label{sec:ranking}

\begin{figure}[t!]
    \centering
    \includegraphics[width=0.45\textwidth,trim=5 5 360 2,clip]{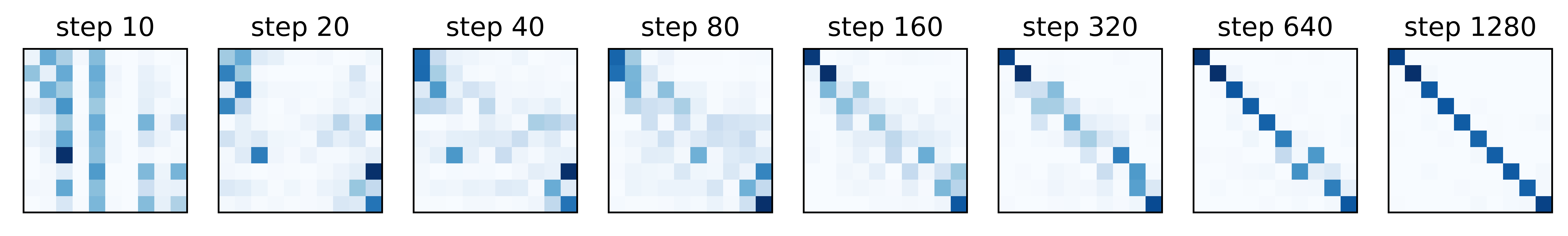}\\
    \includegraphics[width=0.45\textwidth,trim=360 10 5 0,clip]{figures/mnist-rank.png}\\
    \caption{ Confusion matrices of MNIST classifiers in the course of training on batches of 128 ranked pairs of digits. The trajectory of convergence to the diagonal shows that uncertainty is first resolved for the digits 0/9, then 1/8, etc.}
    \label{fig:mnist_pairs}
\end{figure}

We train a CNN of the same architecture as in \S\ref{sec:negative_labels} on MNIST, but with the only supervision coming in the form of pairs of images in which it is known which image represents the greater digit. The training set of 60k images is divided into pairs that are fixed throughout the training procedure; each digit appears in exactly one pair. We optimize to match the predictor $q$ with the implicit posterior model (\ref{eq:pair_posterior}) using the \textbf{RQ} loss. Fig.~\ref{fig:mnist_pairs} shows the confusion matrices at initial iterations of training. The learned classifier has 97\% accuracy on both training and testing sets, which means that from pairwise comparisons alone, we can group the digit images and place them in order.

\subsection{Label super-resolution}
\label{sec:chesapeake_experiments}

\begin{table}[t!]
    \centering
    \caption{Pixel accuracy and class mean intersection over union on the Chesapeake Land Cover dataset. All models use only coarse NLCD labels as supervision. For our proposed methods, we evaluate both the trained predictor ($q_i$) and the posterior under the generative model ($r_i$). The score of the best overall model is \textbf{bolded}. }
    \resizebox{\linewidth}{!}{
\begin{tabular}{@{}lllllll}
    \toprule
   
    & \multicolumn{2}{c}{PA} & \multicolumn{2}{c}{NY} & \multicolumn{2}{c}{Chesapeake} \\
    \cmidrule(lr){2-3} \cmidrule(lr){4-5} \cmidrule(lr){6-7} 
    Model & acc \% & IoU \% & acc \% & IoU \% & acc \% & IoU \% \\
    \midrule

    Self-epitomic$^a$  
             & \textbf{86.2} & 67.6 & 86.4  & 70.5 & 86.3 & 69.7 \\
    Hard na\"ive$^b$  & 85.3 & 63.0 & 83.6 & 59.8 & 83.6 & 59.7 \\ 
    \midrule
    \textbf{QR} ($q$) & 85.9 & 69.3 & 87.3 & 73.0 & 86.4 & 71.1 
 \\
    \textbf{QR} ($r$) & \textbf{86.2}  & \textbf{69.9} & \textbf{87.9}  & \textbf{74.4} & \textbf{86.8}  & \textbf{72.1} 
  \\
    \textbf{RQ} ($q$) & 81.5 & 63.1 & 77.4 & 60.2 & 79.8 & 62.2 
\\
    \textbf{RQ} ($r$) & 81.5  & 63.2 & 77.5  & 60.3 & 79.8  & 62.4 
\\

    \bottomrule
\end{tabular}
}

\footnotesize
$^a$\citep{malkin2020mining}
$^b$\citep{malkin2019label}
    \label{tab:landcover_results_chesapeake}
\end{table}

\begin{figure*}[t!]
    \centering
    \includegraphics[width=0.9\textwidth]{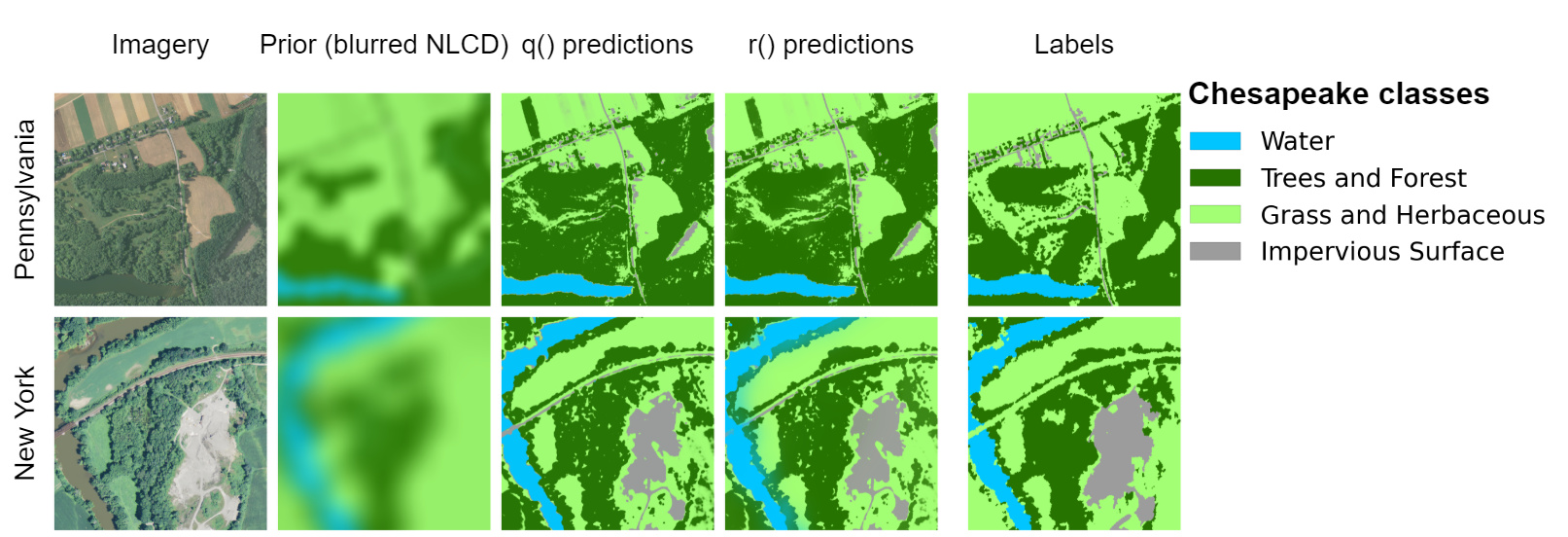}
    \caption{Predictions of models trained with \textbf{QR} loss on the  NLCD-only prior in the Chesapeake region, shown on regions of 1000$\times$1000 pixels in Pennsylvania and 500$\times$500 pixels in New York.}
    \label{fig:chesapeake_predictions}
\end{figure*}

We benchmark our method's performance on the Chesapeake Land Cover dataset
\footnote{\href{https://lila.science/datasets/chesapeakelandcover}{https://lila.science/datasets/chesapeakelandcover}}, 
a large 1m-resolution land cover dataset used previously for label super-resolution \citep{robinson2019large,malkin2019label}. It consists of several aligned data layers, including: NAIP (4-channel high-resolution aerial imagery at about 1m/px), NLCD (16-class, 30m-resolution coarse land cover labels), and \text{high-resolution land cover labels (LC)} in four classes. The task is to train high-resolution segmentation models, in the four target classes, using only NLCD labels as supervision. The NLCD layer is at 30$\times$ lower resolution than the imagery and target labels and follows a different class scheme. Cooccurrence statistics of NLCD classes $c$ and LC labels $\h$ are assumed to be known (Fig.~\ref{fig:cooccurrence_matrices}).

To form a prior over land cover classes $\h$ at each pixel position, we map the 
NLCD classes to probabilities over the target LC classes using these known cooccurrence counts and apply a spatial blur to reduce low-resolution block artifacts (Fig.~\ref{fig:chesapeake_predictions}, ``Prior"). We then train small convolutional networks (receptive field $11\times11$) to predict high-resolution land cover from input imagery.
We evaluate both the \textbf{QR} and \textbf{RQ} variants of our approach on the two states that comprise the ``Chesapeake North" test set: Pennsylvania (PA) and New York (NY), and the two states combined, after picking hyperparameters based on an independent validation set in Delaware (details in \S\ref{app:experiment_details_landcover}). A depiction of the data and prediction results is given in Fig.~\ref{fig:chesapeake_predictions}.

\Cref{tab:landcover_results_chesapeake} compares our algorithms against the algorithmic technique with the best published performance on the Chesapake dataset, self-epitomic LSR \citep{malkin2020mining} and the hard na\"ive baseline from \citet{malkin2019label}. Self-epitomic LSR, a generative modeling approach that explicitly produces likelihoods $p(x|\h)$, analyzes small patches of data by making a large number of comparisons between sampled $7\times7$ image patches and \emph{all other} image patches. It does not produce a trained feedforward inference model, and the inference procedure is at least an order of magnitude slower than evaluation of our convolutional model. The hard na\"ive baseline maps the  NLCD classes to LC classes based on a given concurrence matrix, then trains a standard semantic segmentation model on these pseudo-labels. 

Training on the \textbf{QR} loss outperforms (in once case, matches) performance of self-epitomic LSR (\Cref{tab:landcover_results_chesapeake}), and the generative model for $p(x|c)$ from (\ref{eq:aux_p}) is largely consistent with the epitomic generative model  (Fig.~\ref{fig:p_x_given_c_figure}). Moreover, our methods handle \emph{batched input},
where self-epitomic LSR trains on one data tile at a time. Similar per-tile approaches have been shown to degrade in performance and exhaust computation capacity when training on multiple tiles \citep{malkin2020mining}).
Optimization under an implied generative model has the computational advantage of scaling naturally to large training data while maintaining the benefits of leading generative modeling approaches. (See also \S\ref{sec:lymphocytes}.)

\subsection{Data fusion and learned priors}
\label{sec:enviroatlas_experiments}

\begin{figure*}[t!]
    \centering
    
    \includegraphics[width=0.9\textwidth]{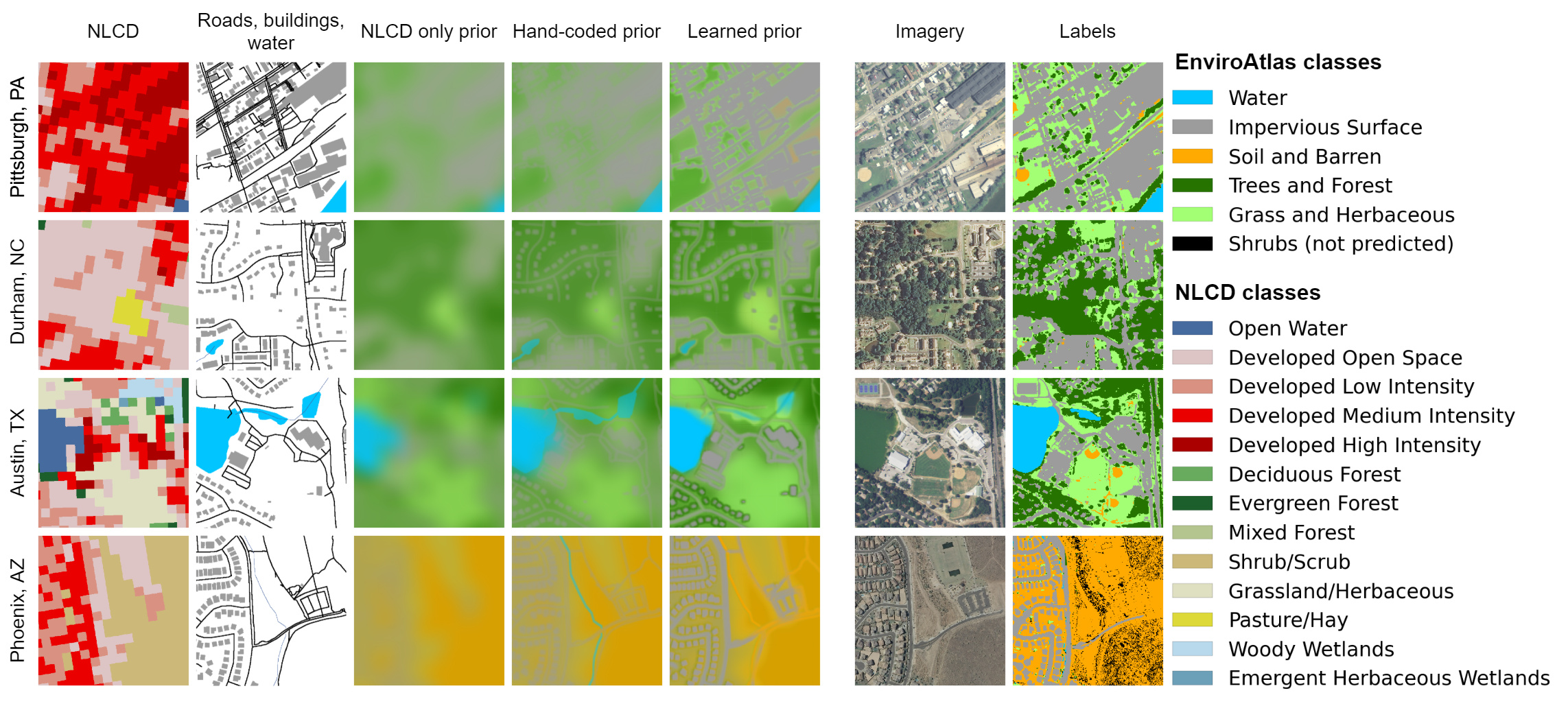}
    \caption{Prior generation for land cover mapping: ``NLCD only prior" (\S\ref{sec:chesapeake_experiments}) and ``$\{$Hand-coded, Learned$\}$ prior" (\S\ref{sec:enviroatlas_experiments}). }
    \label{fig:prior_generation_landcover_mapping}
\end{figure*}

In this set of experiments, we augment NLCD with information about the presence of buildings, road networks, and waterbodies/waterways from public sources (see Fig.~\ref{fig:prior_generation_landcover_mapping} and \S\ref{app:landcover_data_sources}). To evaluate the ability of models to generalize to across regions, we use 1m 5-class land cover labels from the geographically diverse EnviroAtlas dataset \citep{pickard2015enviroatlas} in four cities in the US: Pittsburgh, PA, Durham, NC, Austin, TX, and Phoenix, AZ. The NLCD-based prior model from \S\ref{sec:chesapeake_experiments} is augmented with the auxiliary information to obtain a hand-coded prior for each image (see \S\ref{app:forming_priors_landcover}). These types of priors can be made everywhere in the United States, while hard 1m-resolution labels are rarely available.

An alternative to performing local inference under such priors is to simply apply supervised models trained on hard labels elsewhere, hoping that the domain shift is tolerable. \Cref{tab:landcover_results_enviroatlas} compares the performance of a model (of the same architecture as in \S\ref{sec:chesapeake_experiments}) trained on Pittsburgh high-resolution data (HR) in each of the three other cities with that of models tuned on the hand-coded prior in each other city. 
The \textbf{QR} method trained on the local handmade prior outperforms the HR model in each evaluation city. This may be attributed to the extra data in each city given to our method in the form of prior beliefs. To isolate this effect, we also compare to a high-resolution model that consumes the prior belief to \emph{input} data, concatenated with the NAIP imagery (HR + aux). While the HR + aux model does increase performance substantially from the HR model with NAIP imagery alone as input, the \textbf{QR} model remains the highest-fidelity approach in two of the three cities. These results illustrate that information that generalizes across domains may find its best use within a separate model -- to build a prior in our setting -- and then used to supervise local inference. 

In practice, prior beliefs could be crafted by a domain expert to reflect the uniquities in geographic and structural features for each city.
We emulate incorporating such context-specific knowledge  by training (on a disjoint set of instances) a neural network that consumes the inputs to the handmade prior function (NLCD and auxiliary map data), and predicts high-resolution labels
(Fig.~\ref{fig:prior_generation_landcover_mapping}, ``Learned prior"). 
Alongside structural interactions between the inputs that generalize across cities (e.g., tree canopy supersedes rivers, roads supersede water), the learned prior
captures region-specific knowledge (e.g., buildings in Durham tend to have grass surrounding them and trees farther out, while in Austin,  this is reversed, and in Phoenix, riverbeds surrounded by barren land are likely to be dry).  Using these tailored prior beliefs during \textbf{QR} training tends to increase scores (Table~\ref{tab:landcover_results_enviroatlas}).

The final row in \Cref{tab:landcover_results_enviroatlas} benchmarks the performance of a high-resolution land cover model trained on imagery and labels over the entire contiguous US~\citep{robinson2019large}. This large model takes NAIP, Landsat 8 satellite imagery, and building footprints as inputs. Small, local models with priors created from only weak supervision outperform the US-wide model in all cities. (See \S\ref{app:additional_results} for details.)

\begin{table}[t]
    \centering
    \caption{Land cover classification experiments for generalizing across cities. In each column, the score of the best model not depending on auxiliary data as input is \textit{italicized} and the score of the best overall model is \textbf{bolded}. (A larger set of experimental results is given in \Cref{tab:landcover_results_enviroatlas_full}.)}
    \resizebox{\linewidth}{!}{
\begin{tabular}{@{}llllllll}
\toprule
& 
& 
\multicolumn{2}{c}{Durham, NC} & \multicolumn{2}{c}{Austin, TX} & \multicolumn{2}{c}{Phoenix, AZ} \\
\cmidrule(lr){3-4}
\cmidrule(lr){5-6}\cmidrule(lr){7-8}
Train region & Model 
& acc  & IoU  & acc  & IoU  & acc & IoU  \\\midrule
Pittsburgh & HR & 74.2 & 35.9 & 71.9 & 36.8 & 6.7 & 13.4 
 \\
(supervised) & HR + aux 
& 78.9 & 47.9 & 77.2 & 50.5 & 62.8 & 24.2 
 \\\midrule
Local  & \textbf{QR} ($q$) 
& 78.9 & 47.7 & 76.6 & 49.1 & \textit{75.8} & \textit{45.4} 
 \\
(hand-coded prior) & \textbf{QR} ($r$) 
& 79.0  & 48.4 & 76.6  & 49.5 & \textbf{76.2}  & \textbf{46.0} \\
\midrule
Local   & \textbf{QR} ($q$) 
& \textit{79.0} & \textit{48.7} & \textit{\textbf{79.4}} & \textit{51.3} & 73.4 & 42.8  \\
(learned prior) & \textbf{QR} ($r$) 
& \textbf{79.2}  & 49.5 & 79.1  & \textbf{51.9} & 73.6  & 43.1  \\ \midrule
Full US$^a$ & U-Net Large  
& 77.0 & \textbf{49.6} & 76.5 & 51.8 & 24.7 & 23.6 \\
\bottomrule
\end{tabular}
}

\footnotesize
$^a$\citep{robinson2019large}
    \label{tab:landcover_results_enviroatlas}
\end{table}

\subsection{Text classification}
\label{sec:nlp}

This experiment follows the recent work of \citet{mekala2021coarse} and illustrates the effectiveness of learning on prior beliefs beyond computer vision. We work with a dataset of $\sim$12k New York Times news articles. Each article belongs to one of 20 fine categories (e.g., `energy companies', `tennis',`golf'), which are grouped into 5 coarse categories (e.g., `business', `sports'). The goal is to train text classifiers that predict fine labels, but only the coarse label for each article is available in training. 

Some external knowledge about the fine categories is necessary to resolve the coarse labels into fine labels. Past work on this problem \citep{meng2018weakly,mekala2020contextualized,meng2020text,wang2021xclass} has trained supervised models on pseudolabels created by mechanisms such as propagation of seed words and querying large pretrained models. On the other hand, \citet{mekala2021coarse} create training data by sampling additional \emph{features} (articles) from a finetuned version of the large generative language model GPT-2 \citep{radford2019language} conditioned on fine categories, then tune a classifier based on the almost equally large model BERT \citep{devlin2019bert} in a supervised manner.

We obtain comparable results using an elementary predictor, far less computation, and no finetuning of massive language models (Table~\ref{tab:coarse_text_results}). We form a prior $p_i(\ell)$ on the fine class $\ell$ of each article $x_i$ by querying GPT-2 for the likelihood of each fine category name $\ell$ compatible with the known coarse label  following the prompt ``[article text] Topic: '' and normalizing over $\ell$. We then divide $p_i(\ell)$ by the mean likelihood of $\ell$ over all articles $x_i$ and renormalize.
We represent each article as a vector of alphabetic trigram counts ($26^3$ features, of which only 8k are ever nonzero) and train a logistic regression with the \textbf{RQ} objective against this `GPT-2 prior'.
After ten epochs of training ($\sim$10s on a Tesla K80 GPU), the trained classifier nears or exceeds the performance of models requiring at least $100\times$ longer to train, even excluding the time to generate any pseudo-training data.

\begin{table}[t]
\centering 
\caption{F1-scores of various models on the coarsely supervised text classification task. The first five rows are taken from \citet{mekala2021coarse}. The last two rows use the GPT-2 prior defined in \S\ref{sec:nlp} as weak supervision  with cross-entropy and \textbf{RQ} loss, respectively (mean of 10 random trials).}
\resizebox{1\linewidth}{!}{
\begin{tabular}{@{}llcc}
\toprule
 & Algorithm & Micro-F1 \% & Macro-F1 \% \\\midrule
\multirow{4}{*}{pseudolabeling}
 & {WeSTClass}$^a$ & 76.23 & 69.82 \\
 & {ConWea}$^b$    & 73.96 & 65.03 \\
 & {LOTClass}$^c$  & 15.00 & 20.21 \\
 & {X-Class}$^d$   & 91.16 & 81.09 \\\midrule
 pseudodata & C2F$^e$         & 92.62 & \textbf{87.01} \\\midrule
 \multirow{3}{*}{\begin{minipage}{25mm}GPT-2 prior\\(trigram features)\end{minipage}}
 & prior argmax    & 86.33 & 77.61 \\
 & CE              & 87.18 & 77.90 \\
 & \textbf{RQ}     & \textbf{93.18} & 84.26 \\
 \bottomrule
\end{tabular}
}

\footnotesize
$^a$\cite{meng2018weakly}
$^b$\cite{mekala2020contextualized}
$^c$\cite{meng2020text}
$^d$\cite{wang2021xclass}
$^e$\cite{mekala2021coarse}
\label{tab:coarse_text_results}
\end{table}

\section{Discussion and conclusion}

In summary, we found that the generative distribution in a free energy criterion can be left implicit to the minimization process in posterior (discriminative) model training. This allowed us to unite the training of neural networks $q(\h|x_i; \theta)$ for prediction of labels $\h$ from features $x$ with the modeling of the prior $p_i(\h)$, possibly with its own latent structure. Implicit modeling of the conditional generative distributions removes the burden of training accurate (and therefore large or deep) generative models, but still allows natural generative approaches to modeling priors.

Learning a discriminative network $q$ and its implicit posterior model $r$ via the \textbf{QR} and \textbf{RQ} methods can unify common supervised learning paradigms with realistic label supervision settings, enabling high-fidelity predictions from weak supervision sources carrying far less information. 
The additional experimental results in \S\ref{sec:additional_exp} detail further results for weakly supervised image segmentation, self-supervised learning, and co-segmentation in video data.  

Code is available in an accompanying GitHub repository (see \S\ref{sec:code}): \href{https://github.com/estherrolf/implicit-posterior}{\url{ https://github.com/estherrolf/implicit-posterior}}.

\begin{contributions}
\newtext{       
E.R., N.M., A.G., N.J. jointly conceived the main ideas and their analysis and presentation in this work. 
E.R. conducted the land cover experiments.
N.M. conducted the experiments on negative labels and ranks, text, and lymphocytes and ran the land cover baselines.
A.G. conducted the experiments on video tracking and the \textit{Le s\'educteur} experiments.
A.J. and N.J. conducted the experiments on self-supervised image clustering.
C.R. helped with compute and storage resources and with implementation of land cover experiments in TorchGeo. 
All authors collaboratively wrote the paper.
}
\end{contributions}

\begin{acknowledgements}

\newtext{
We thank Anthony Ortiz for helpful feedback during the ideation and writing stages of this work. We also thank the anonymous reviewers for their comments and suggestions.}

\newtext{The main contributions of this work were conceptualized and conducted while E.R. and A.G. were interns at Microsoft Research, Redmond.
Computation resources were provided by Microsoft AI for Earth. E.R. additionally acknowledges the support of a Google PhD Fellowship.}
\end{acknowledgements}

\bibliography{references}

\appendix
\newpage
\onecolumn

\counterwithin{figure}{section}
\counterwithin{table}{section}
\counterwithin{equation}{section}

\section{Code}
\label{sec:code}

This paper is accompanied by a code repository at \href{https://github.com/estherrolf/implicit-posterior}{\tt github.com/estherrolf/implicit-posterior}. The  repository contains three directories. Two of them illustrate our algorithms for partial-label learning and weakly supervised segmentation and are sufficient to reproduce predictions resembling those in  Fig.~\ref{fig:examples}. The third directory contains code for the land cover mapping experiments (\S\ref{sec:chesapeake_experiments}, \S\ref{sec:enviroatlas_experiments}).

\section{Practical Considerations}
\label{sec:practical_considerations}
\paragraph{Mini-batches:} 
Figure~\ref{fig:qr_losses_torch} shows a PyTorch implementation of the QR and RQ loss functions, where loss is computed over \emph{batches} of training data.
Our experiments validate that so long as these batches are large enough to include enough diversity of $(x_i,p_i(l))$ pairs, our method works when \cref{eq:aux_p} and \cref{eq:true_post} are applied directly to batches. As discussed in \S\ref{sec:enviroatlas_experiments}, handling batched input 
is important for leveraging the scale of large training datasets.
As discussed in \S\ref{sec:implicit_generative_model},
should mini-batch training become an issue in future implementations, it may be beneficial to estimate the denominator of \cref{eq:aux_p} across multiple batches.

\newtext{
To illustrate the dependence of the algorithm on batch size, we ran the MNIST experiment with one negative label (\S\ref{sec:negative_labels}) with differing batch sizes (Table~\ref{tab:batch_size}). The performance degrades at batch sizes 32 and smaller, when batches are likely to be missing samples of some classes. 
}

\begin{table}
\centering
\caption{
\newtext{
Peak test accuracies (following the same experiment settings as in \S\ref{sec:negative_labels}) and standard deviations over 10 random seeds with different training batch sizes. The last two columns show properties of the distribution over the number of distinct classes in a randomly sampled batch: the likelihood that all ten MNIST classes occur at least once and the expected number of distinct classes that occur.}}
\begin{tabular}{rccll}
\toprule

&\multicolumn{2}{c}{peak test acc \%}\\
\cmidrule(lr){2-3}
batch size & RQ & NLL& $\mathbb{P}[\text{all 10 classes appear in batch}]$& $\mathbb{E}[\# \ \text{distinct classes in batch}]$ \\\midrule
256&95.96$\pm$0.24&94.57$\pm$3.12&100.00\%&10.00\\
128&96.32$\pm$0.39&94.83$\pm$3.21&100.00\%&10.00\\
64&96.66$\pm$0.21&96.15$\pm$0.25&98.82\%&9.99\\
32&94.18$\pm$1.05&96.64$\pm$0.20&69.10\%&9.66\\
16&93.35$\pm$3.21&96.85$\pm$0.22&7.03\%&8.14\\
8&92.41$\pm$4.65&96.78$\pm$0.19&0&5.70\\
4&91.10$\pm$6.42&96.99$\pm$0.23&0&3.44\\
2&89.04$\pm$10.29&96.93$\pm$0.18&0&1.90\\
\bottomrule
\end{tabular}
\label{tab:batch_size}
\end{table}

\paragraph{Relative benefits/limitations of the QR and RQ loss formulations:} 
The algorithm presented in \S\ref{sec:implicit_generative_model} details two loss options: a \textbf{QR} option and an \textbf{RQ} option, both with unique strengths.
The QR algorithm is guaranteed to converge as each step reduces loss (except for randomness in the learning algorithm). The RQ algorithm, on the other hand, has the appealing property that it reduces to standard minimization of cross entropy loss in the case of hard labels. In \S\ref{sec:implicit_details}, we discuss connections between QR option and variational auto-encoders (VAEs), and between the RQ option and the wake-sleep algorithm. Ultimately, though, we find that which option works better may depend on the application, with RQ working across all applications we tried but sometimes being slightly beaten by QR. 

Comparing performance across these varied learning settings can shed light on the performance of the proposed \textbf{QR} and \textbf{RQ} methods under different conditions. 
Future research %
could systematize and formalize settings where one variant would be superior to the other; results in this work show that both can be effective ways to resolve uncertainty in non-``ground-truth" labels. 

\paragraph{Simple ways to avoid degenerate solutions:}
As discussed in \S\ref{sec:implicit_generative_model}, minimizing \cref{eq:free_eng} can lead to degenerate solutions. However, avoiding these solutions can be quite simple, and in most of our experiments we did not make any interventions to explicitly avoid such local minima. In a targeted experiment in~\cref{tab:landcover_results_enviroatlas_full} we show that pre-training on hard labels (even out-of-domain) or using sharper learned priors can help break symmetries during early training phases.
When hard labels are not available, one could similarly start the training process with a cross-entropy loss on the prior belief, and then switch to RQ or QR loss. 
The intuition is that first training to minimize cross-entropy breaks the symmetry at the start, while implicit posterior modeling sharpens the predictions in later iterations.

\section{Additional Related Work}
\label{sec:related_work_extras}

There are several approaches to learning with uncertain, weak, or coarse labels under different assumptions and settings.
Work on partial-label learning often employs loss functions that aim to decrease prediction entropy \citep{nguyen2008classification,yao2020ambiguous,yu2016maximum}. These approaches do not use a generative formulation in these loss functions, making them less suitable for problems with more varied forms of uncertainty encoded in  priors.
\newtext{
Another approach to learning with imprecise or fuzzy data is to learn a model which finds the best (deterministic) disambiguation of uncertain observations, often by generalizing traditional loss minimization techniques \citep{hullermeier2014learning,couso2018general,cabannnes2020structured}.
}

\newtext{
In \S\ref{sec:priors}, we discuss several opportunities to form prior beliefs from weak (e.g. coarse, imprecise, or uncertain) observations, including fusing multiple data sources.
While these illustrative examples set the stage for experiments in \S\ref{sec:experiments} and \S\ref{sec:additional_exp}, 
several alternative and additional techniques have been developed to 
model and utilize data from weak sources  \citep{hernandez2016weak,zhou2018brief}. 
For example, data programming  \citep{ratner2016data,ratner2017snorkel} provides an opportunity to collect and learn from multiple weak user-provided labeling functions.}
\newtext{Another} line of work studies the generation and use of pseudolabels in learning settings.
Specifically, \cite{zou2020pseudoseg} relies on a domain-specific augmentation procedure for semantic segmentation with image-level labels, and, \cite{zhang2021refining} studies unsupervised clustering applied to object re-identification. 
Application-specific solutions also include object detection in remote sensing images~\citep{han2014object} and change detection with multitemporal satellite imagery~\citep{zheng2021Weakly,bao2021MRTA,li2021Change}.

In our experimental setups, we chose a mix of baselines to both compare algorithm design and benchmark performance on certain tasks. 
To compare our approach on an \emph{algorithmic basis}, we compare to the negative logarithm of the sum of likelihoods (NLL), which is used in prior works to handle multiple ambiguous labels~\citep{jin2002learning} and negative labels~\citep{nlnl}. 
We compare to self-epitomic LSR~\citep{malkin2020mining} as an algorithmic comparison by which to contrast our method with an ``explicit" generative modeling approach. 
Our similar performance to self-epitomic LSR in regimes where self-epitomic LSR has been shown to perform well (super-resolution in land cover mapping (\S\ref{sec:chesapeake_experiments})
and the tumor-infiltrating lymphocytes task (\S\ref{sec:lymphocytes})) is an important validation of our motivation in \S\ref{sec:background_and_approach}.

To benchmark \emph{performance} of our approach across tasks, we compare to state-of-the-art pseudo-labeling methods in supervised text classification~ (see \S\ref{sec:nlp}), an established 1m resolution map of land cover predictions across the United States~\citep{robinson2019large} and best-performing published results for the land cover mapping tasks we study~\citep{malkin2020mining}~\citep{robinson2020human}, the best known published results for the tumor-infiltrating lymphocyte segmentation task \citep{malkin2019label,malkin2020mining}, and a host of comparisons for the video instance segmentation task (see \Cref{tab:tracking_results} for a full list). 

\newtext{
As stated in \S\ref{sec:negative_labels},  the NLL (union) objective and \textbf{RQ} are equivalent when $\sum_i q_i(\ell)$ is uniform over $\ell$ and the prior is uniform over all classes in the negative label sets, evidenced by the comparable performance between the two in \Cref{fig:mnist_cifar}. In this case, the denominator in (\ref{eq:true_post}) is independent of $\ell$, and thus \[r_i(\ell)=\begin{cases}\frac{1}{C-|N_i|}q_i(\ell)&\ell\notin N_i\\0&\ell\in N_i\end{cases},\]
where $C$ is the number of classes and $N_i$ is the negative label set for sample $i$. The \textbf{RQ} loss then simplifies as
\[{\rm KL}(r_i\|q_i)=\mathbb{E}_{\ell\sim r_i}\left[\log\left(\frac{r_i(\ell)}{q_i(\ell)}\right)\right]=\sum_{\ell\notin N_i}\frac{1}{C-|N_i|}q_i(\ell)\log\frac{1}{C-|N_i|},\]
which is a constant multiple of the NLL (union) loss $\sum_{\ell\notin N_i}q_i(\ell)$.
}

Lastly, it is worth noting that the similar term ``implicit generative model" has been used in prior literature to refer to amortized sampling procedures for nonparametric (or not specified) energy functions, such as generative adversarial models (e.g., \cite{igm}). Although we do not make an explicit connection with such models, our formulation also does not assume a parametrization of the data distribution, and one can understand the term ``implicit posterior'' as referring to a function that is a posterior for an implicit (i.e., uninstantiated, unparametrized) generative model. However, we assume tractability of sampling from a posterior over certain distinguished latents (classes) conditioned on observed data (features, e.g., images), rather than directly sampling latents.

\section{Relationships with EM, VAE, and wake-sleep algorithm}
\label{sec:implicit_details}

\begin{table}
\caption{
Comparison of modeling forms for variational auto-encoders (VAE), wake-sleep algorithms (WS), expectation-maximization (EM), and our proposed implicit posterior (IP). Variational auto-encoders parametrize both a generative model $p$ and a posterior model $q$. Here we distinguish between $\theta_p$ and $\theta_q$ as these models can differ in both architecture and parameters.
The EM formulation parametrizes the generative model $p(x_i|\h;\theta_p)$ and the posterior is instantiated as auxiliary matrix with entries $a_{i,\h}$ calculated to maximize the objective given the estimated $p(x_i|\h;\theta_p)$ on the observed instances $i$. 
In implicit posterior modeling, the posterior $q(\h|x_i; \theta_q)$ is modeled and parametrized directly, with the generative link $p$ instantiated as an auxiliary matrix with entries of the form $a_{i,\h}$. Combining this auxiliary matrix with the prior beliefs $p_i(\h)$ at each instance as in Eq.~\eqref{eq:true_post} yields a posterior model $r_i$ \emph{implied} by forward model $q(\h; x_i, \theta_q)$ and weak prior beliefs on each instance $p_i(\h)$.}
\label{tab:em_vae_comparison}
\centering
\begin{tabular}{lccc}
        & VAE/WS & EM & IP  \\ 
    \midrule
     generative $p$& $p(x|\ell; \theta_p)$ & $p(x|\ell; \theta_p)$ & $a_{i,\ell}$ \\
     posterior $q$& $q(\ell|x; \theta_q)$& $a_{i,\ell}$  & $q(\ell|x; \theta_q)$ \\
\end{tabular}
\end{table}

As discussed in \S\ref{sec:implicit_generative_model}, the \textbf{QR} loss guarantees continual improvements in the free energy (\ref{eq:free_eng}).  On the other hand, option \textbf{RQ} is equivalent to performing a gradient step on the cross-entropy of $q_i$ and $r_i$ and a gradient step on the \emph{negative} entropy of $r_i$. In the case that the priors $p_i(\ell)$ are hard (supported only on one ground truth label), the same is true of $r_i$, and the \textbf{RQ} loss is equivalent to cross-entropy.
This option reverses the KL distance in a manner reminiscent of the training procedure in the wake-sleep algorithm \citep{hinton1995wake}, where parameter updates for the forward and reverse models are iterated, but the KL distance optimized always places the probabilities under the model being optimized in the second position in the KL distance (inside the logarithm), so that the generative and the inference models each optimize log-likelihoods of their predictions. %
The wake-sleep algorithm, however, also trains a generative model rather than treating it as an auxiliary distribution as we do, and that requires sampling. As opposed to VAEs, the wake-sleep algorithm samples the generative model, not the posterior.

It is interesting to contrast our approach to the expectation-maximization (EM) formulation. In standard EM, the $q$ distributions are considered auxiliary, rather than parametrized as direct functions of the inputs $x$. The $q_i(\h) = a_{i,\h}$ is simply a matrix of numbers normalized across $\h$. Its dependence on the data $x$ arises through the iterative re-estimation of the minimum of the free energy, where the link $x-\h$ is modeled directly in the parametrized forward distribution $p(x|\h)$ (see Table~\ref{tab:em_vae_comparison}). We instead model forward probabilities $p(x_i|\h)$ as auxiliary parameters, a matrix of numbers $a_{i,\h}$ normalized across $i$ that we fit to minimize the free energy at each data point, and optimize only the parameters of the $q$ model which explicitly models the link $x-\h$. This allows us to capture nonlinear (and `deep') structure and benefit from inductive biases inherent to training deep models with SGD, but without the cost of training an actual parametrized generative model and other problems associated with deep generative model fitting. The resulting $q$ network approximates the posterior in a generative model -- which (locally) maximizes the log likelihood of the data -- and it is usually highly confident (as seen in Fig.~\ref{fig:examples}).

The implicit modeling of the posterior in EM does not lead to overfitting of the generative model. But, given that degenerate solutions to optimization with implicit posterior models are possible when the prior is constant across all data points (\S\ref{sec:implicit_generative_model}), we can imagine that our approach of implicit posterior modeling might lead to degenerate solutions. As demonstrated in Fig.~\ref{fig:examples} and in our experiments, avoiding degenerate solutions is not too hard. We address this point further in \S\ref{sec:practical_considerations}.

\section{Experiment details}

\subsection{Land cover mapping}
\subsubsection{Datasets}
\label{app:landcover_data_sources}
\paragraph{Imagery Data}
Our land cover mapping experiments use imagery from the National Agriculture Imagery Program (NAIP), which is 4-channel aerial imagery at a $\leq 1$m/px resolution taken in the United States (US).

\paragraph{Chesapeake Conservancy land cover dataset}
The Chesapeake Conservancy land cover dataset consists of several raster layers of both imagery and labels covering parts of 6 states in the Northeastern United States: Maryland, Delaware, Virginia, West Virginia, Pennsylvania, and New York~\citep{robinson2019large}\footnote{Dataset can be downloaded from: \url{https://lila.science/datasets/chesapeakelandcover}.}. The raster layers include: high resolution (1m/px) NAIP imagery, high resolution (1m/px) land cover labels created semi-autonomously by the Chesapeake Conservancy, low resolution (30m/px) Landsat-8 mosaics imagery, low resolution (30m/px) land cover labels from the National Land Cover Database (NLCD), and building footprint masks from the Microsoft Building Footprint dataset. The dataset is partitioned into train, validation, and test splits per-state, where each split is a set of $\approx 7\text{km} \times 6\text{km}$ \textit{tiles} containing the aligned raster layers.

\paragraph{EPA EnviroAtlas data}

The EnviroAtlas land cover data consists of high resolution (1m/px) land cover maps over 30 cities in the US, and is collected and hosted by the US Environmental Protection Agency (EPA)~\citep{pickard2015enviroatlas}. A detailed description of the dataset and its land cover definitions is provided by \cite{enviroatlas_definitions}. 
As with most high-resolution land cover datasets (including the Chesapeake Conservancy land cover labels), the EnviroAtlas land cover labels are themselves derived by remote sensing and learning procedures, and thus are not themselves a perfect ``ground truth'' representation of land cover.
For example, the estimated accuracy of the provided labels is 86.5\% in Pittsburgh, PA, 83.0\% in Durham, NC, 86.5\% in Austin, TX, and 69.2\% in Phoenix, AZ \citep{enviroatlas_definitions}. 

The high-resolution label files were aligned to match the extent of the NAIP tiles from the closest available years to the years that the EnviroAtlas labels were collected: for Pittsburgh, PA and Phoenix, AZ, we used data from 2010 and for Durham, NC and Austin, TX, we used data from 2012.
We chose these four cities to get a wide coverage across the United States (US), and due to a mostly consistent set of classes being used between the four cities.

\paragraph{National Land Cover Database (NLCD)}
The National Land Cover Database is produced by the United States Geological Survey (USGS) and uses 16 land cover classes. Maps are generated every 2-3 years, with spatial resolution of 30m/px. Data and more information can be found at: \url{https://www.usgs.gov/centers/eros/science/national-land-cover-database}.

\paragraph{Microsoft Building Footprint dataset}
The Microsoft Building Footprint dataset consists of predicted building polygons over the continental US from Bing Maps imagery. As of the time of writing, the most updated Microsoft Building Footprints dataset in the US can be accessed at: \url{https://github.com/Microsoft/USBuildingFootprints}.

\paragraph{Open Street Map (OSM) data}
Open Street Map (\url{https://www.openstreetmap.org/})
is an ongoing effort to make publicly available and editable map of the world, generated largely from volunteer efforts. The  data is available under the Open Database License. From the many different sources of information provided by OSM~\citep{OpenStreetMap}, we download raster data for road networks, waterways, and water bodies, using the OSMnx python package \citep{boeing2017osmnx}.

\paragraph{Data splits and data processing}
For experiments using the Chesapeake Conservancy dataset (\Cref{tab:landcover_results_chesapeake}), we used established train, test, and validation splits. In particular, we used the 20 test tiles in New York (NY) and the 20 test tiles in Pennsylvania (PA) on which to conduct our experiments. Here a \emph{tile} matches the extent of a NAIP tile, roughly 7km $\times$ 6km.
To facilitate comparison of  our results with previous published results on this dataset, we condensed the labels into four classes: (1) water, (2) impervious surfaces (roads, buildings, barren land), (3) grass/field, and (4) tree canopy.

For experiments with the EnviroAtlas dataset (\Cref{tab:landcover_results_enviroatlas}), we aligned the high resolution land cover data, NLCD, OSM, and Microsoft Building Footprints data with NAIP imagery tiles, matching years as closely as possible to the EnviroAtlas data collection year for NLCD and NAIP.
We instantiated a split of 10 train, 8 validation, and 10 test tiles in Pittsburgh, and 10 test tiles in Durham, NC, Austin, TX, and Phoenix, AZ. For Pittsburgh we assigned tiles to splits randomly from the set of 28 tiles that had no missing labels. There were not enough such tiles in Durham to follow the same procedure, so we chose the ten evaluation tiles at random from a set with no number of missing labels per tile. For Austin and Phoenix, we chose the 10 evaluation tiles at random from the tiles in each city that had no agriculture class (as it is not present in Pittsburgh or Durham) and no missing labels. We set aside 5 separate tiles in each city %
for use in ``learning the prior'' (in Pittsburgh these 5 tiles are a subset of the 8 validation tiles). As above, each tile corresponds to one NAIP tile.
The tiles in these constructed sets for Pittsburgh, Durham, and Austin contain 
five unique labels: (1) water, (2) impervious surfaces (roads, buildings), (2) barren land, (4) grass/field, and (5) trees. Phoenix additionally has a ``shrub'' class; when forming the prior we merge this class with trees, and we ignore the shrub class when evaluating in Phoenix.
We cropped all data tiles to ensure no spatial overlap in any tiles between or within the train/val/test splits.

\subsubsection{Forming the priors}
\label{app:forming_priors_landcover}

To form the priors for the land cover classification tasks, we first spatially smooth the NLCD labels by applying a 2D Gaussian filter (with a standard deviation of 31 pixels) across every channel in a one-hot representation of the NLCD classes. The main reason for applying this smoothing is to reduce artifacts due to the 30m$^2$ boundaries of the NLCD data, to undo the blocking procedure induced by the aggregation to 30m $\times$ 30m extents, to incorporate the spatial correlations between nearby NLCD blocks, and to remove erroneous sharp differentials between inputs that can cause artifacts during later training stages.

\begin{figure}
    \centering
    \includegraphics[width=.8\columnwidth]{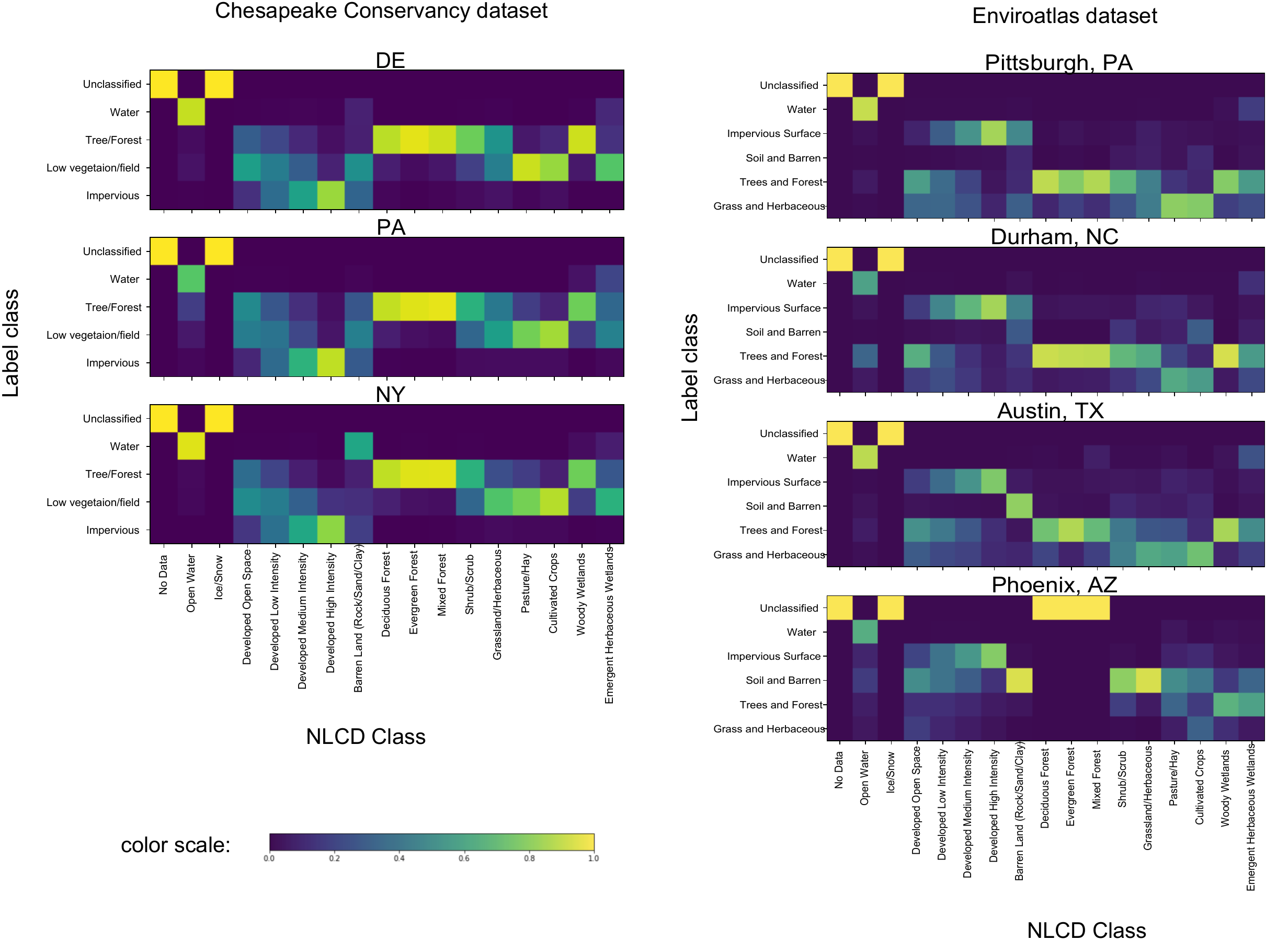}
    \caption{Cooccurrence matrices between NLCD classes and high resolution land cover labels for each region we study.}
    \label{fig:cooccurrence_matrices}
\end{figure}

We then remap the blurred NLCD layers to the classes of interest by multiplying by a matrix of cooccurrence counts between the (unblurred) NLCD data and the high resolution labels in each region. For the Chesapeake region, we use the train tiles provided with the Chesapeake Conservancy land cover dataset to define cooccurrence matrices in NY and PA. For EnviroAtlas, we compute cooccurrences using the entire city (excluding tiles with agriculture in Phoenix AZ, and Austin, TX). The cooccurrence matrices for each region we study are shown in \Cref{fig:cooccurrence_matrices}.

The priors for the Chesapeake Conservancy dataset are then generated by normalizing the blurred and remapped NLCD data so that summing over all five classes gives probability 1 for each pixel.

For the EnviroAtlas data, we augment this prior with publicly available data on buildings, road networks, water bodies, and waterways. We obtain building maps from the Microsoft Buildings Footprint database and road, water bodies, and waterways data from Open Street Map, using the OSMnx tool \citep{boeing2017osmnx} to download the data (see \Cref{app:landcover_data_sources}). We apply a small spatial blur to each of these input sources to account for (a) vector representation of roads and waterways being unrealistically thin, and (b) possible data-image misalignment on the order of pixels. Where this results in probability mass on impervious surfaces or water, we add these probability masses to the blurred NLCD prior, and then renormalize to obtain a valid set of probabilities for each pixel.

In \S\ref{sec:enviroatlas_experiments}, we describe a method for ``learning the prior,'' which uses a more sophisticated process to aggregate the individually weak and coarse inputs that we use in the handmade prior. 
In this method, we train a neural net to take as input the blurred, remapped NLCD representation (5 classes) concatenated with the 4 classes of additional data: buildings, roads, waterways, water bodies, and to predict high-resolution labels in each city. 
We train these networks using 5 tiles of imagery and high-resolution labels from the EnviroAtlas Dataset in each city which are distinct from the 10 test tiles in each city. 
The training procedure for these prior generation networks is described in
in \S\ref{app:experiment_details_landcover}. 
To create the priors that we then train our method on (`learned prior' rows in \Cref{tab:landcover_results_enviroatlas})
we ran these learned models forward on (blurred and remapped NLCD, buildings, roads, waterways, and waterbodies) input for each of the 10 evaluation tiles in each city.

\subsubsection{Experimental procedure}

We use priors generated as described in \Cref{app:forming_priors_landcover}, with Gaussian spatial smoothing with standard deviation of 31 pixels, and cooccurrence matrix determined via the training splits in each city/state. We apply a pixel-wise additive smoothing constant of 1e-4 to the probability vectors output by the neural network as well as to the prior probability vectors used as the model supervision data. This additive smoothing constant ensures that there are no extremely low probability classes in either the prior or the predicted outputs during training.

Experiments summarized in \Cref{tab:landcover_results_chesapeake} and \Cref{tab:landcover_results_enviroatlas} use a 5-layer fully connected network with kernel sizes of 3 at each layer, 128 filters per layer, and leaky ReLUs between layers. Note that the receptive field of this model is only $11\times11$ pixels. We use batch sizes of 128 instances during training, where each image instance is a cropped 128 $\times$ 128 pixels from a larger tile.  Training and model evaluation is done within the torchgeo framework for geo-spatial machine learning \citep{torchgeo}. All models use the AdamW optimizer~\citep{loshchilov2017decoupled} during training and torchgeo defaults unless otherwise noted.

\label{app:experiment_details_landcover}
\paragraph{Comparison to previous label super-resolution for LC mapping}
To obtain the parameter setting used for the runs in New York (NY) and Pennsylvania (PA) in \Cref{tab:landcover_results_chesapeake}, we first perform a hyperparameter search with the 20 tiles test set in Delaware (DE) from the same overall dataset. 
We use a learning rate schedule that decreases learning rate when the validation loss plateaus, as well as early stopping to prevent over training of models.
Of the grid of learning rates in \{1e-3, 1e-4,1e-5\}, we describe below, we pick learning rate as 1e-4 for both \textbf{QR} and \textbf{RQ} variants of our method, as this is the setting that minimizes the IoU of the $q$ output on the 20 DE tiles for both variants.
 
When training on NY and PA jointly (``Chesapeake" in \Cref{tab:landcover_results_chesapeake}), we use the per-state cooccurrence matrices. This ensure that the cooccurrence matrices used are consistent between our method and the self-epitomic LSR benchmark across all columns in \Cref{tab:landcover_results_chesapeake}.

\paragraph{Generalization across cities.}

For the high-resolution model with NAIP imagery from Pittsburgh as input, we consider learning rates in $\{10^{-2},10^{-3},10^{-4},10^{-5}\}$ and pick based on the best validation performance on the validation set in Pittsburgh. The chosen learning rate is 1e-3. We search over the same set of learning rates for the model with NAIP imagery and the prior concatenated as input; the chosen learning rate is also 1e-3. For this model with concatenated image and prior as input, only the number of input channels changes in the fully connected network model architecture. When training on the high-resolution land cover labels, we use a very small additive constant (1e-8) for the last layer of the model.

When training our methods, we initialize model weights using the best NAIP image input model from the Pittsburgh validation set runs, and then train using the priors and the training procedure described in the main text. We pick the learning rate for this training step using again the validation set in Pittsburgh; we search learning rates in $\{10^{-3},10^{-4},10^{-5}\}$, and pick 1e-5 as the learning rate for \textbf{QR} and 1e-3 as the learning rate for \textbf{RQ}, since these resulted in the best performance for the Pittsburgh validation set with the randomly initialized model. We discuss the results of a similar procedure using randomly initialized model weights in \Cref{app:additional_results}.

For the learned prior, we use a 3 layer fully connected network is kernel sizes of 11,7, 5 respectively, 128 filters per layer and leaky ReLUs between layers. For each city, we train this model on the prior inputs (blurred and remapped NLCD, roads, buildings, waterways, and water bodies) using a validation set of 5 tiles separate from from the 10 evaluation tiles in each city. We considered learning rates in $\{10^{-3},10^{-4},10^{-5}\}$ for learning the prior in each city, and chose 1e-4 as it gave most often resulted in the highest accuracies of each validation set. For learning \emph{on} this learned prior, we again initialize model weights using the best NAIP image input model from the Pittsburgh validation set runs, and set the learning rate to 1e-5 for \textbf{QR} evaluation runs and 1e-3 for \textbf{RQ} evaluation runs to match the other variants of the experiment.

\begin{table*}[htb]
    \centering
    \caption{Supplementary results to accompany \Cref{tab:landcover_results_enviroatlas}.}
    \resizebox{\textwidth}{!}{
\begin{tabular}{llllllllll}
\toprule
& & \multicolumn{2}{c}{Pittsburgh, PA} & \multicolumn{2}{c}{Durham, NC} & \multicolumn{2}{c}{Austin, TX} & \multicolumn{2}{c}{Phoenix, AZ} \\
\cmidrule(lr){3-4}\cmidrule(lr){5-6}\cmidrule(lr){7-8}\cmidrule(lr){9-10}
Train region & Model & acc \% & IoU \% & acc \% & IoU \% & acc \% & IoU \% & acc \% & IoU \% \\\midrule
Pittsburgh & HR & 89.3 & 69.3 & 74.2 & 35.9 & 71.9 & 36.8 & 6.7 & 13.4 
 \\
(supervised) & HR + aux & 89.5 & 70.5 & 78.9 & 47.9 & 77.2 & 50.5 & 62.8 & 24.2 
 \\\midrule
Same as test& \textbf{QR} ($q$)& 80.5 & 56.8 & 78.3 & 44.4 & 79.2 & 50.5 & 75.2 & 29.5   \\
(random  & \textbf{QR} ($r$) & 80.7 & 57.5 & 78.5 & 46.4 & 79.7 & 52.0 & 75.9 & 33.8
 \\
initialization) & \textbf{RQ} ($q$) & 77.6 & 53.3 & 65.8 & 23.3 & 73.8 & 43.0 & 61.8 & 18.6
  \\
& \textbf{RQ} ($r$) & 77.6 & 53.3 & 65.8 & 23.3 & 73.8 & 43.1 & 61.8 & 18.6 
 \\\midrule
Same as test
& \textbf{QR} ($q$)& 80.6 & 58.5 & 78.9 & 47.7 & 76.6 & 49.1 & 75.8 & 45.4 
  \\
(pretrained & \textbf{QR} ($r$)& 80.6 & 58.7 & 79.0 & 48.4 & 76.6 & 49.5 & 76.2 & 46.0 

  \\
in Pittsburgh)  & \textbf{RQ} ($q$) & 84.3 & 59.6 & 75.6 & 28.6 & 76.5 & 47.5 & 63.7 & 19.5 
  \\
& \textbf{RQ} ($r$) & 84.3 & 59.6 & 75.4 & 31.5 & 76.5 & 47.5 & 63.7 & 19.5 
  \\\midrule
Same as test
& \textbf{QR} ($q$) & 82.4 & 63.7 & 79.0 & 48.7 & 79.4 & 51.3 & 73.4 & 42.8 

 \\
(learned prior ) & \textbf{QR} ($r$) & 82.4 & 64.0 & 79.2 & 49.5 & 79.1 & 51.9 & 73.6 & 43.1 
 \\ \midrule
Full US* \cite{robinson2019large} & U-Net Lrg. & 79.0 & 61.5 & 77.0 & 49.6 & 76.5 & 51.8 & 24.7 & 23.6 
 \\
\bottomrule
\end{tabular}
}
    \label{tab:landcover_results_enviroatlas_full}
\end{table*}

\subsubsection{Additional Results}
\label{app:additional_results}

\begin{table*}[htb]
    \centering
    \caption{Comparison of the Full US* U-Net Large \citep{robinson2019large} map predictions when evaluated on the full 5 classes considered in \Cref{tab:landcover_results_enviroatlas} (water, grass/field, trees/shrub, impervious surfaces, and barren land) and evaluated on the four prediction classes predicted by the model (where barren land and impervious surfaces are merged as a single class), and when barren is post-facto assigned whenever the predicted class is ``impervious surfaces" and the label class is ``barren land". }
    \begin{tabular}{lllllllll}
\toprule
&  \multicolumn{2}{c}{Pittsburgh, PA} & \multicolumn{2}{c}{Durham, NC} & \multicolumn{2}{c}{Austin, TX} & \multicolumn{2}{c}{Phoenix, AZ} \\
\cmidrule(lr){2-4}\cmidrule(lr){4-5}\cmidrule(lr){6-7}\cmidrule(lr){8-9}
Classication Scheme & acc \% & IoU \% & acc \% & IoU \% & acc \% & IoU \% & acc \% & IoU \% \\\midrule
      5 Classes & 78.8 & 55.1 & 76.6 & 43.4 & 76.2 & 49.1 & 18.2 & 18.8 
\\
     4 Classes & 79.0 & 68.7 & 77.0 & 54.1 & 76.5 & 60.4 & 24.7 & 16.8 
\\
      Barren reassigned  & 79.0 & 61.5 & 77.0 & 49.6 & 76.5 & 51.8 & 24.7 & 23.6 
\\
\bottomrule
\end{tabular}
    \label{tab:cvpr_2019_enviroatlas_4_and_5_class}
\end{table*}

\paragraph{Extended results for generalizing across EnviroAtlas cities.}

\begin{figure*}
    \centering
    \includegraphics[width=\columnwidth]{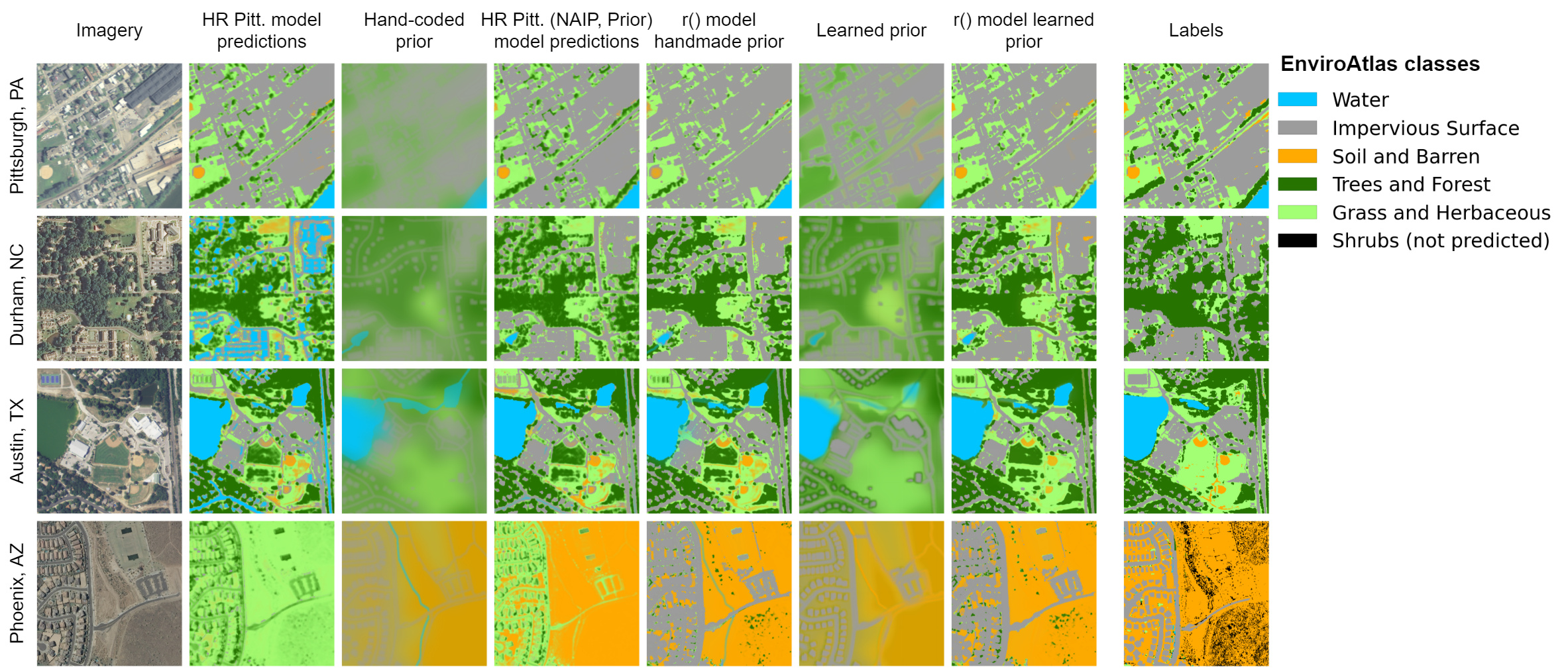}
    \caption{Example predictions on the hand-coded and learned prior in each EnviroAtlas city we study.}
    \label{fig:enviroatlas_preds_with_learned_prior}
\end{figure*}

The extended results for generalizing across cities with the EnviroAtlas datasets in \Cref{tab:landcover_results_enviroatlas_full} contain the results of the \textbf{RQ} runs trained on the handmade prior in each city. Evaluation results in Pittsburgh, PA give further context for comparison of generalization across cities by each method.

\Cref{tab:landcover_results_enviroatlas_full} also details the result of initializing the model weights randomly for the \textbf{QR} method.  \Cref{tab:landcover_results_enviroatlas_full} shows that the choice of model initialization can be important for our method -- this is most apparent in Pittsburgh, PA (unsurprisingly since the high-resolution model was trained in Pittsburgh) and Phoenix, AZ. In Phoenix, much of the handmade prior is consistent across geographies and the randomly initialized model has trouble distinguishing between infrequent classes that most often occur together in the handmade prior. The results in \Cref{tab:landcover_results_enviroatlas_full} suggest that using pre-trained models as a starting point for our method can help to break some of these symmetry issues in resolving the information in the prior. Results in \Cref{tab:landcover_results_enviroatlas} suggest that using a more detailed prior map may help with this as well.

\paragraph{Evaluating the Full US map from \citet{robinson2019large}.}
Recall that the row for the full US Map \citep{robinson2019large} in \Cref{tab:landcover_results_enviroatlas} reflects the performance of the model evaluated on all 5 classes we consider in our experiments, where we give the map predictions the ``benefit of the doubt" in that any prediction of ``impervious surfaces" where the true label is ``barren land" gets assigned a correct classification of ``barren land." The results reported in \Cref{tab:landcover_results_enviroatlas} are thus a sort of upper bound on the predictive performance of the method that generated the predictive maps. It was important for us to keep the barren class while evaluating across cities, as it is the dominant class in Phoenix, AZ. In the remaining three cities, the barren class is challenging to predict as it is  infrequent.
In \Cref{tab:cvpr_2019_enviroatlas_4_and_5_class}, we compare this classification scheme with two alternatives: a 5 class scheme that will penalizes the map predictions for never predicts the barren class, and a 4 class scheme that merges the barren land and impervious surfaces classes in evaluation.  \Cref{tab:cvpr_2019_enviroatlas_4_and_5_class} shows that while the choice of evaluation scheme does not greatly effect accuracy (outside of Phoenix, AZ, where the accuracy of the Full US Map is low for both classification schemes), the average IoU drops significantly for all cities apart from Phoenix.

\paragraph{Comparing loss functions: qualitative results with land cover mapping.}

\begin{figure}
    \centering
    \includegraphics[width=0.8\textwidth]{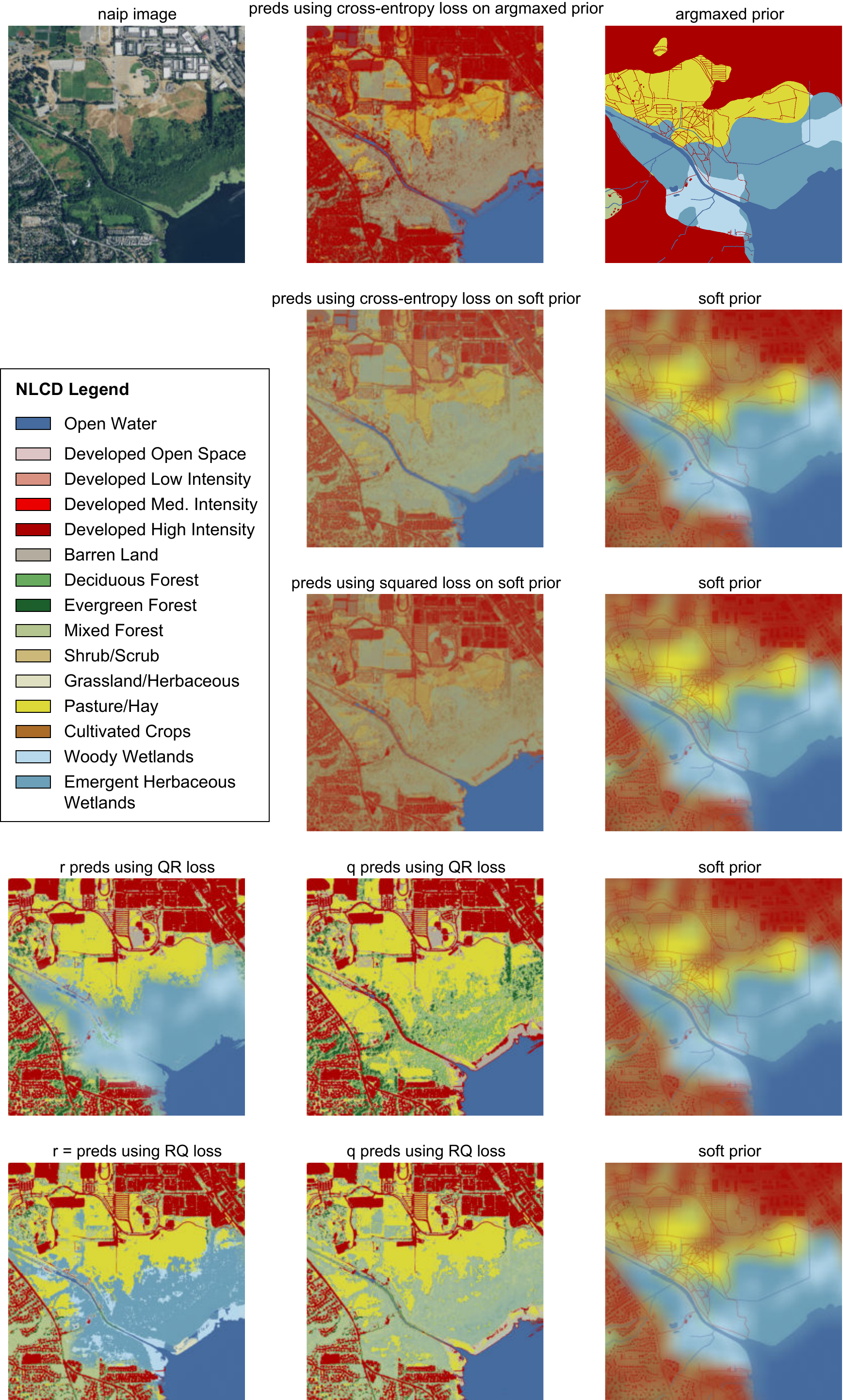}
    \caption{Comparison of different loss functions on hard and soft prior.}
    \label{fig:sammamish_loss_comparison}
\end{figure}

\Cref{fig:sammamish_loss_comparison} compares predictions under different loss functions with an illustrative example. Here the prior is similar to the ``hand-coded" prior described in \Cref{app:forming_priors_landcover}, but with the prior defined over all NLCD classes. We train each model (a slight variant on the network used in experimental results) on the single NAIP tile region encompassing the zoom-in in the figure for 2000 iterations with the Adam algorithm \citep{kingma2014adam}, a batch size of 64, and a learning rate fixed at 1e-4 during training. Qualitative comparisons show that predictions made by the \textbf{QR} and \textbf{RQ} loss functions are more certain (sharper colors in plots) than training with cross entropy or squared-error loss on the soft priors, and, in in most places, arrive at better solutions than training with a standard cross entropy loss on the argmax of the prior.

\begin{figure*}
    \centering
    \includegraphics[width=\textwidth]{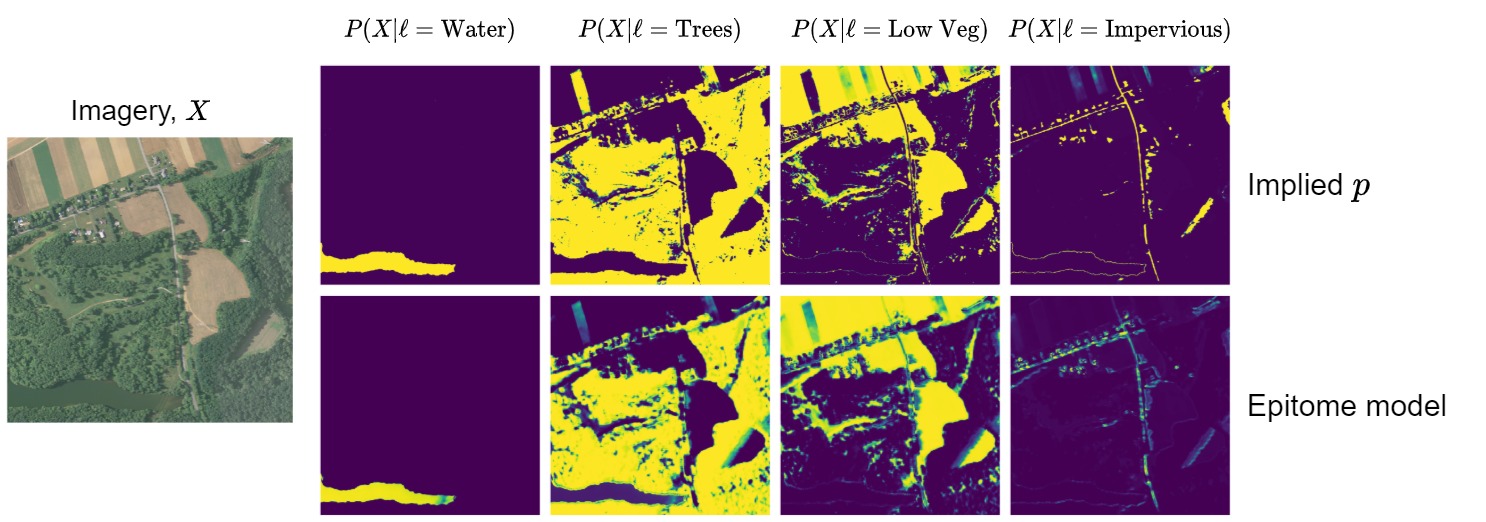}
    \caption{Comparison of forward model likelihoods under the generative model trained with \textbf{QR} loss (above) and the likelihood under an epitome model \citep{malkin2020mining} for part of a test tile from \S\ref{sec:chesapeake_experiments}.}
    \label{fig:p_x_given_c_figure}
\end{figure*}

\section{Additional experiments}
\label{sec:additional_exp}

\subsection{Self-supervision for unsupervised image clustering}
\label{sec:aaai_regret}

Neural networks are usually trained on large amounts of hard-labeled data $\{x_i,\h_i\}$, yet, due to the biases induced by the typical architectures and learning algorithms, much of the modeling power of these networks seem to focus on correlations in the input space \citep{shwartz2017opening}. This means that a network trained for one application, i.e., for one label space $\h \in L_1$, can be adopted to another application, i.e., a different labels space $\h \in L_2$, as long as the input features are in a similar domain. The canonical example of this is the use of lower levels of the networks pre-trained on ImageNet as part of the networks solving a completely different set of image classification problems. Pretrained networks require smaller training sets in fine tuning, as long as they have learned to represent the variation in the input space well. Self-supervised models attempt to go a step further and learn these representations without \emph{any} labels.  In our framework, self-supervision can simply be seen as the appropriate choice of subset priors $p(\h_T)$ over appropriately chosen tuples of labels.

To discuss the pitfalls and opportunities, consider again the \textbf{QR} loss (\ref{eq:QR})
\begin{equation}
    F=-\sum_{i,\h} q_i(\h)\log p_i(\h) +\sum_{i,\h} q_i(\h) \log \left( \sum_j q_j(\h) \right) ~.
    \label{eq:QR_again}
\end{equation}
If we were to simply set $p_i(\h)$ to a constant (e.g., uniform) distribution $p(\h)$ for all data points $i$, then the optimal solution would be any function $q_i(\h)=q(\h|x_i)$ such that $\frac{1}{N}\sum_i q(\h|x_i) = p(\h)$. Thus simply using the uniform prior may not lead to appropriate unsupervised clustering (or self-supervised learning of the network $q$). The inductive biases in the network architecture and training may not help, because one solution is $q(\h|x)=p(\h)$, which can be achieved by zeroing out all weights except for biases in a final softmax layer that outputs probabilities for labels $\h$. As the softmax bias vector is the closest to the top in back-propogation with gradient descent, it will quickly be learned to match $\log p(\h)$. This will not only slow down the propagation of gradients into the network, but could eventually stop it completely, as this solution is a global optimum. Another optimal solution would be a function satisfying $\frac{1}{N}\sum_i q(\h|x_i) = p(\h)$, but where individual entropies for each data point are small:  $-\sum_\h q(\h|x_i) \log q(\h|x_i)<\epsilon$, which motivates an alternative cost criterion:
\begin{equation}
    F=-\sum_{i,\h} q_i(\h)\log q_i(\h) + \sum_{i,\h} q_i(\h) \log \left(\sum_j q_j(\h)\right)~.
    \label{eq:diverse_clustering}
\end{equation}
where the first term promotes certainty in predictions $q(\h|x_i)$ for each point $i$ and the second is promoting the diversity of the predictions across the different inputs, i.e., a high entropy of the average $\frac{1}{N}\sum_h q_i(h)$. This prevents learning a network with a constant output $q(h)=p(h)$ and forces the model to find some statistics in the input data that break it into clusters indexed by labels $\h$. The result will be highly dependent on the inductive biases associated with the network architecture and SGD method used, as we can imagine degenerate solutions here as well. For example, we can ignore completely some subset of features and still train a network that is certain in its modeling of the remaining ones, and achieves a high diversity of predicted classes across the dataset. This may be dangerous if the features omitted end up being the most important ones for the downstream task. However, due to the stochastic gradient descent training as well as their architecture, it has been difficult to prevent neural networks from learning statistics involving all the input features. For example, training a neural network using a weak generative model as a teacher  corresponds to using a simpler mixture model, whose posterior is used as a target $p_i(\h)$ and then learning a neural network that can approximate it. The inductive bias then leads to networks that do not match $p_i(\h)$ exactly but learn more complex statistics instead. 

Equation (\ref{eq:diverse_clustering}) can be seen as a degenerate example of using a tuple prior where the tuple has the same data point repeated and the prior simply expects the two predictions to be the same. In many applications, there are natural constraints involving multiple data points that are easily modeled with priors over tuples or over the entire collection of labels. Consider unsupervised image segmentation, for an example. It is usually expected that nearby pixels should belong to the same class (or a small subset of classes), and that faraway pixels are more likely to belong to a different subset of classes. This belief is typically modeled in terms of Markov random field models of joint probabilities of labels in the image,
\begin{equation}
    p(\{\h_i\}) \propto \exp\sum_i \phi(\h_i, \{\h_j\}_{j\in N_j}).
\end{equation}
We experimented with potentials of the form 
\begin{equation}
    \phi(\h_i=\h, \{\h_j\}_{j\in N_j})=\gamma_\ell + \alpha_\h \frac{1}{|S_i|}\sum_{j\in S_i} \mathbbm{1}[\h=\h_j] +\beta_\h \frac{1}{|L_i|}\sum_{j\in L_i} \mathbbm{1}[\h=\h_j], 
\end{equation}
where for pixel $i$, $S_i$ is a small ($5\times 5$) neighborhood around it and $L_i$ is a larger ($50 \times 50$) neighborhood. If we set $\alpha_\h=1$, $\beta_\h=-1$ for all $\h$, then we consider this a contrastive prior, as it favors labels $\h_i$ to match the labels found more concentrated in its immediate neighborhood than in the larger scope. On the other hand $\alpha_\h$, and $\beta_\h$ can be estimated based on the current statistics in the label distribution using logistic regression. We refer to this as a self-similarity prior $p(\{\h_i\}; \alpha_\h, \beta_\h, \gamma_\h)$ with parameters which are periodically fit to the current statistics in the predictions $\sum_{j\in S_i} q(\h|x_j)$, and $\sum_{j\in L_i} q(\h|x_j)$ to promote similar label patterns across the image. The criterion (\ref{eq:diverse_clustering}) can also be seen as a degenerate version of this setting with $S$ being $1 \times 1$ and $L$ being infinite (or the whole image).

The contrastive version of this prior relies on the insight previously pursued in image self-supervision, e.g., \cite{jean2019tile2vec}. In our formulation, contrasting is accomplished without sampling triplets, but considering all the data jointly, by expressing the goal of contrasting with far away regions within the prior in our framework.

\begin{figure*}[th]
    \centering
    \includegraphics[width=0.9\textwidth]{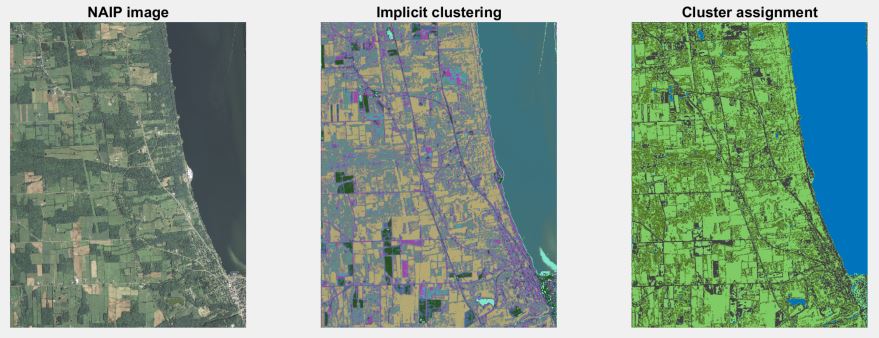}
    \caption{Unsupervised clustering using implicit \textbf{QR} loss (middle) of a NAIP tile (left). On the right, we show the assignment of the 12 clusters to 4 land cover labels: water (blue), tall vegetation (darker green), low vegetation (lighter green) and impervious/barren (gray).}
    \label{fig:selfsup}
\end{figure*}

As an example of self-supervised pretraining in our framework, in Fig. \ref{fig:selfsup} we show an example of clustering a large tile of aerial imagery into 12 classes using 5 layer FCN as network $q$ of the architecture used in \S\ref{sec:enviroatlas_experiments}. The clustering is achieved by updating the prior every 50 steps of gradient descent on batches of $200$ $256\times 256$px patches. The prior is initialized to a contrasting prior, and then updated through gradient descent. After 7 iterations, the result is sharpened by continuing training using (\ref{eq:diverse_clustering}). 

This tile was recently used in testing the fine-tuning of a pretrained model with minimal amount of new labels in a new region \citep{robinson2020human}. Both the pre-training region, the state of Maryland, and the testing region, the tiles in New York State, come from the 4-class Chesapeake Land Cover dataset (\S\ref{sec:chesapeake_experiments}). Yet, the slight shift in geography results in reduction of accuracy from around $90\%$ in Maryland down to around $72.5\%$ in New York. In \cite{robinson2020human}, various techniques for quick model adaptation are studied, on labels acquirable in up to 15 minutes of human labeling effort per tile. In Table \ref{tbl:aaai_regret} we compare the tunability of our self-supervised models on the four 85km$^2$ regions tested in \cite{robinson2020human} with active learning approaches to tuning a pre-trained Maryland model with 400 labeled points. We show in the table the accuracy and mean intersection over union from \cite{robinson2020human} for tuning the pretrained model's last $64 \times 4$ layer with different active learning strategies for selecting points to be labeled. For example, random selection of 400 points for which the labels are provided yields an average accuracy improvement from $72.5\%$ to $80.6\%$. 

On the other hand, recall that we have created an unsupervised segmentation into 12 clusters, with posteriors over the clusters $q_i(\h)$. To investigate how well these clusters align with ground truth land cover labels, we compute a simple assignment of clusters to land cover labels. Given a set of labeled points $\{(i,c_i)\}_{i\in I}$, we infer a mapping from clusters to four target labels, \[p(c|\h)\propto\sum_{i\in I:c_i=c}q_i(\h).\] The label of any point $j$ can now be inferred as $\hat\h_j=\argmax_c\sum_\h q_i(\h)p(c|\h)$.
This procedure, using 400 randomly selected labeled points, yields an average accuracy of $81.1\%$ (averaged over 50 random collections of labeled points), which is above the performance of the pretrained model tuned on as many randomly selected points, and on par with the more sophisticated methods for point selection and the use of  the pretrained model (Table~\ref{tbl:aaai_regret}). (Note that the large model pretrained was trained on a large similar dataset in a nearby state).

\begin{table*}[thb]
\centering
\caption{Finetuning a pre-trained model by gradient descent \citep{robinson2020human} versus implicit \textbf{QR} clustering + label assignment in low-label regimes.}
\begin{tabular}{lccccc}
\toprule
& \multicolumn{4}{c}{{pretrained model in \cite{robinson2020human}}} & \multicolumn{1}{c}{{Implicit QR}} \\
\cmidrule(lr){2-5}\cmidrule(lr){6-6}
Query method & \multicolumn{1}{c}{No tuning} & \multicolumn{1}{c}{Random} & \multicolumn{1}{c}{Entropy} & \multicolumn{1}{c}{Min-margin} & \multicolumn{1}{c}{Random}  \\ \midrule
Tuned parameters & \multicolumn{1}{c}{0} & \multicolumn{1}{c}{64$\times$4} & \multicolumn{1}{c}{64$\times$4} & \multicolumn{1}{c}{64$\times$4} & \multicolumn{1}{c}{12$\times$4}  \\ \midrule
Accuracy \% & 72.5 & 80.6 & 73.6 & 81.1 & 81.1   \\
IoU \% & 51.0 & 60.8 & 50.1 & 60.8 & 59.8 \\
\bottomrule
\end{tabular}
\label{tbl:aaai_regret}
\end{table*}

\subsection{Tumor-infiltrating lymphocyte segmentation}
\label{sec:lymphocytes}

The setup of this experiment mimics that of the land cover label super-resolution experiment in \S\ref{sec:chesapeake_experiments}. The training data consists of 50,000 $240\times240$px crops of H\&E-stained histological imagery at 0.5$\mu$m/px resolution, paired with coarse estimates of the density of tumor-infiltrating lymphocytes (TILs) created by a simple classifier, at the resolution of $100\times100$ blocks. The goal is to produce models for high-resolution TIL segmentation. Models are evaluated on a held-out set of 1786 images with high-resolution point labels for the center pixel.

The coarse density estimates $c$ belong to one of 10 classes, from 0 (no TILs) to 9 (highest estimated TIL density). We use an estimated conditional likelihood $p(\h|c)$ of the likelihood of the positive TIL label at pixels with each low-resolution class $c$ to construct a prior $p_i(\ell)$ over the TIL label probability. Notice that this prior is the same for all pixels in any given low-resolution, coarsely labeled block.\footnote{We experimented with setting $p_i(\h|c)$ to conditional likelihoods estimated from a held-out set and with simply setting $p_i(\h=1|c=0)=0.05$, $p_i(\h=1|c=1)=0.15$, \dots, $p_i(\h=1|c=9)=0.95$. The latter gave better results, perhaps due to the bias of the evaluation set, in which every image is known to be centered on a cell of some kind.}

We train a small CNN with receptive field $11\times11$ (five ReLU-activated convolutional layers with 64 filters) under the \textbf{RQ} loss against this prior for 200 epochs with learning rate $10^{-5}$, then evaluate on the held-out testing set. Inspired by \citet{malkin2020mining}, we apply a spatial blur of 11 pixels to the predicted log-likelihoods (again correcting for the model's small receptive field and the dataset bias). 

The AUC scores of this model and of the baselines are shown in Table~\ref{tab:lymphocyte_results}. Interestingly, the best-performing models -- \textbf{RQ} and epitomic super-resolution (a generative model) -- both have receptive fields of $11\times11$, much smaller than those of the U-Net and fully supervised CNNs. This means that prediction of TIL likelihood is possible using only \emph{local} image data, but the challenge is learning to resolve highly uncertain label information. Unlike U-Nets and deep CNN autoencoders, small models are not able to learn and overfit to \emph{distant} spurious clues to the classes of nearby pixels.

\begin{table*}[t!]
    \centering
    \caption{Area under ROC curve for various predictors on the TIL segmentation task.}
    \begin{tabular}{lcccccc}
\toprule
& \multicolumn{3}{c}{fully supervised} & \multicolumn{3}{c}{weakly supervised} \\ \cmidrule(lr){2-4}\cmidrule(lr){5-7}
Model & SVM$^{a,b}$& CNN$^b$ & CSP-CNN \cite{hou2018sparse} & U-Net$^c$ & Epitome$^d$ & \textbf{RQ} \\ \midrule
AUC & 0.713 & 0.494 & 0.786 & 0.783 & 0.801 & 0.802 \\
\bottomrule
\end{tabular}

\footnotesize
$^a$\cite{zhou2017evaluation}
$^b$\cite{hou2018sparse}
$^c$\cite{malkin2019label}
$^d$\cite{malkin2020mining}
    \label{tab:lymphocyte_results}
\end{table*}

\subsection{Video segmentation with a structured prior}
\label{sec:tracking}

To demonstrate the use of priors with latent structure, we set up the problem of video segmentation as follows. Given a frame $t$, we tune networks $q_t(\ell_{i,t}|x_{i,t})$ predicting one of $L$ pixel classes for a pixel at coordinate $i$ in frame $t$. The prior in each frame comes from a Mask R-CNN model \citep{he2017maskrcnn} pre-trained on still images in the COCO dataset \citep{coco}. The Mask R-CNN model finds several possible instances of objects of different categories and outputs the soft object masks in form of confidence scores for each pixel. We convert this into a probability distribution over the index $f$ (foreground/background) of the form $p(f_{i,t}|m_t)$, where $m_t$ are different detected instances by the model, and the distributions $p(f_{i,t}|m_t)$ are the soft masks for these instances converted to probability distributions, i.e. value of the probability of foreground differs for each pixel and each instance based on the Mask R-CNN confidence scores. Although the COCO dataset may not have had instances of object of interest in our frame $x_t$, we assume that some admixture (i.e., mixture with sample-dependent weights) of detected instances (likely involving unrelated types of objects) does model reasonably well the foreground segmentation in the frame. Mathematically, $p(f_{i,t})=\sum_{m_t} p(f_{i,t}|m_t)p(m_t)$, where $p(m_t)$ expresses the probabilistic selection of the foreground masks for different instances from which the foreground is constructed. (One can think of instances $m_t$ as akin to topics in topic models, which are also admixture models). To complete the prior, we fix the distribution $p(\h|f)$ as fixed binary $L \times 2$ matrix assigning a subset of $L$ pixel classes to foreground and the rest to the background. (For example, we assign first 3 classes to foreground and the remaining 5 to the background for a total of L=8 pixel classes). Therefore,
\begin{equation}
    p(\h_{i,t}=\h)=\sum_f p(\h|f) \sum_{m_t} p(f_{i,t}=f|m_t)p(m_t) ~.
\end{equation}
We can now select the instances $m_t$ in each frame by optimizing the free energy with this prior over $p(m_t)$. The procedure involves standard variational inference of the posterior distribution over possible instances $m_t$ for each pixel $i$ in frame $t$ which involves the posterior $q_t(\h_{i,t}|x_{i,t})$. In practice we found that it is enough to do this inference once, using the network $q_{t-1}$ estimated in the previous frame.

\begin{figure*}[t]
    \centering
    \begin{tabular}{@{}c@{\hspace{8pt}}c@{\hspace{8pt}}c@{}@{\hspace{8pt}}c@{}}
    \includegraphics[width=0.22\textwidth]{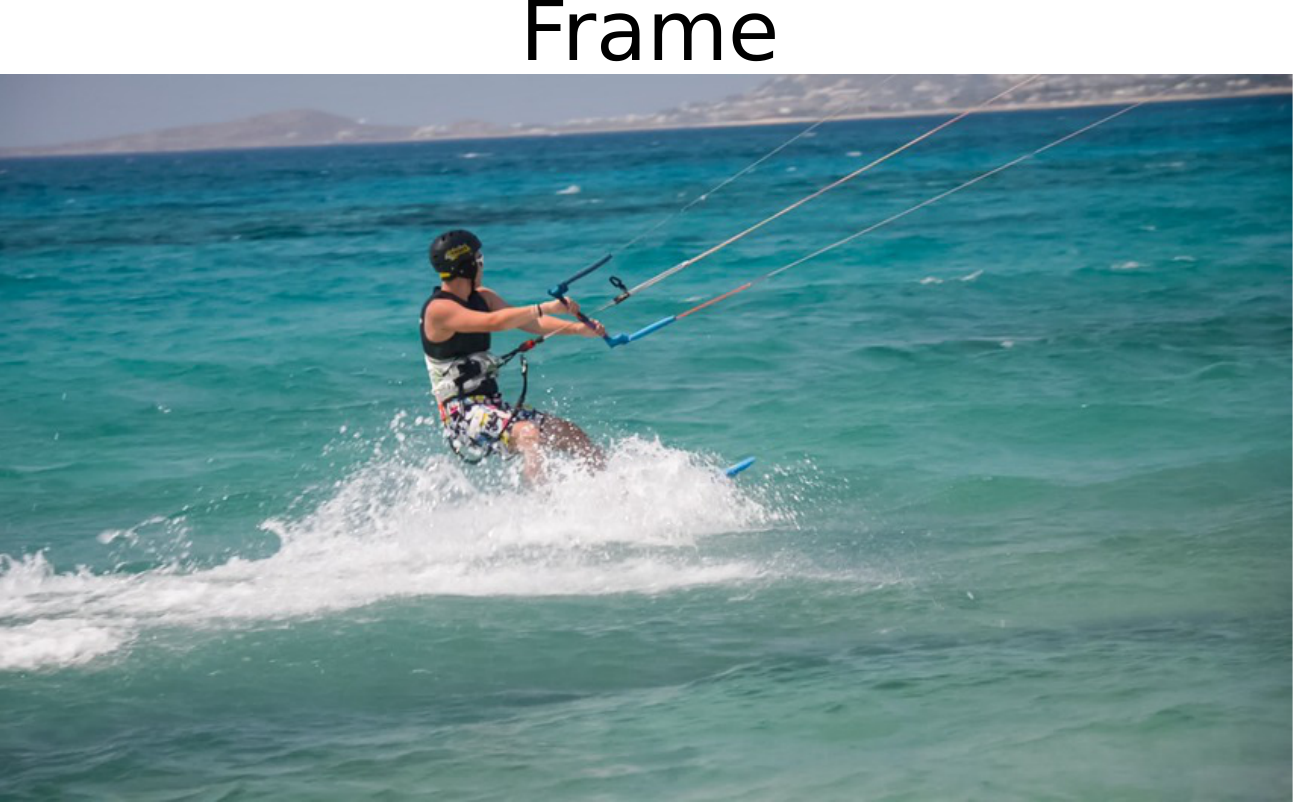} 
    & \includegraphics[width=0.22\textwidth]{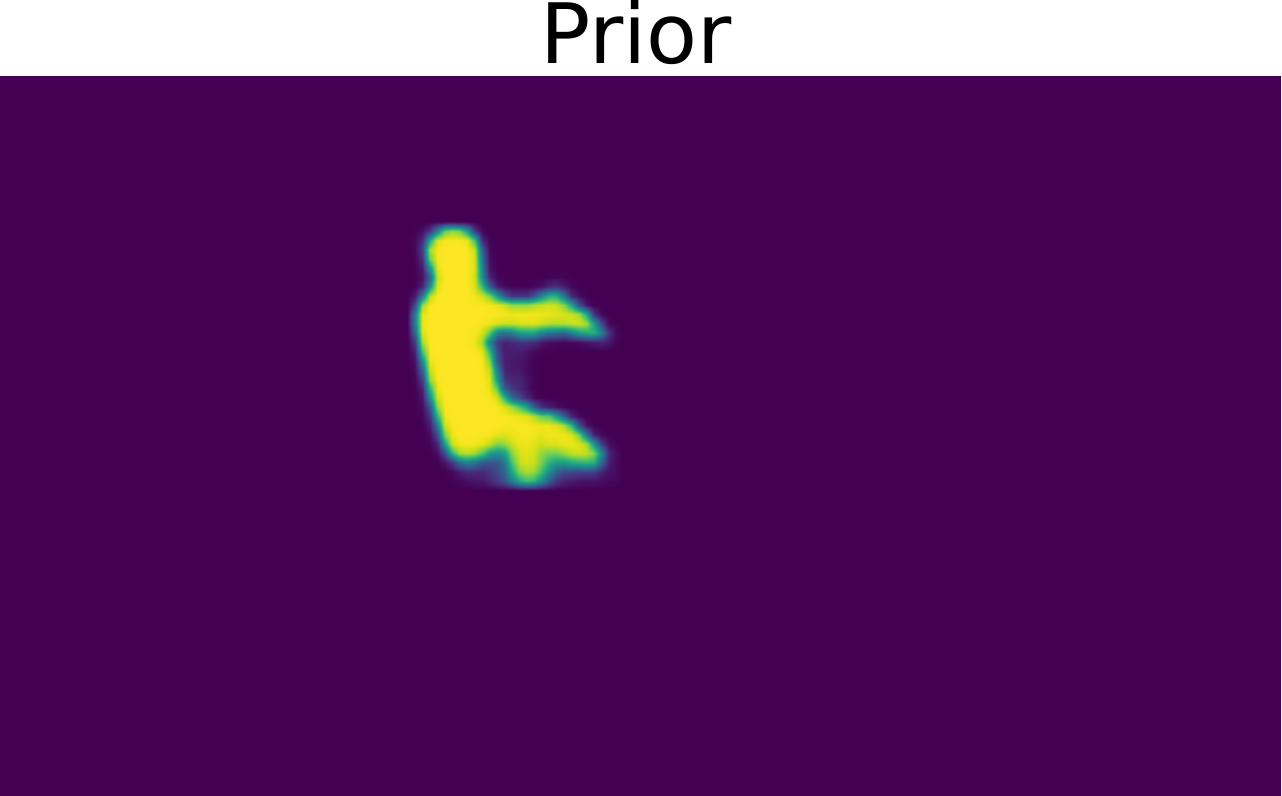}
    & \includegraphics[width=0.22\textwidth]{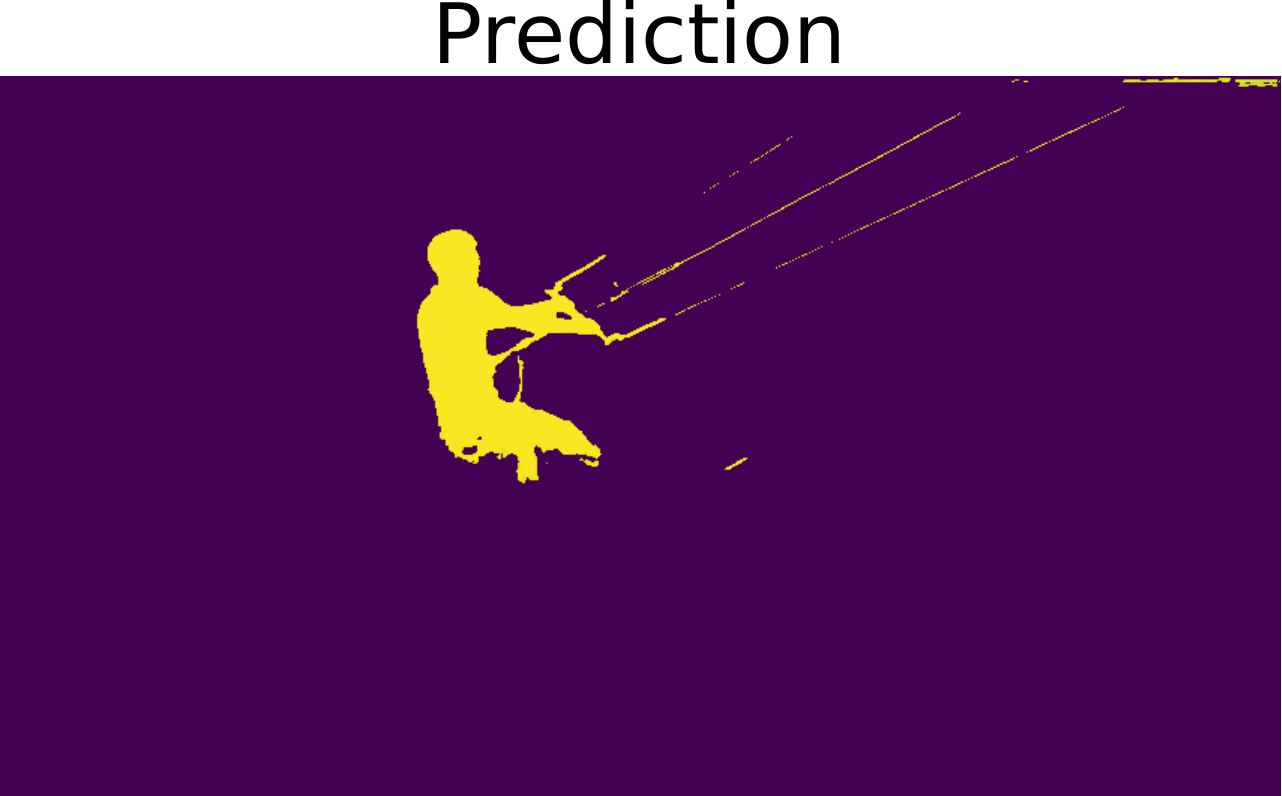}
    & \includegraphics[width=0.22\textwidth]{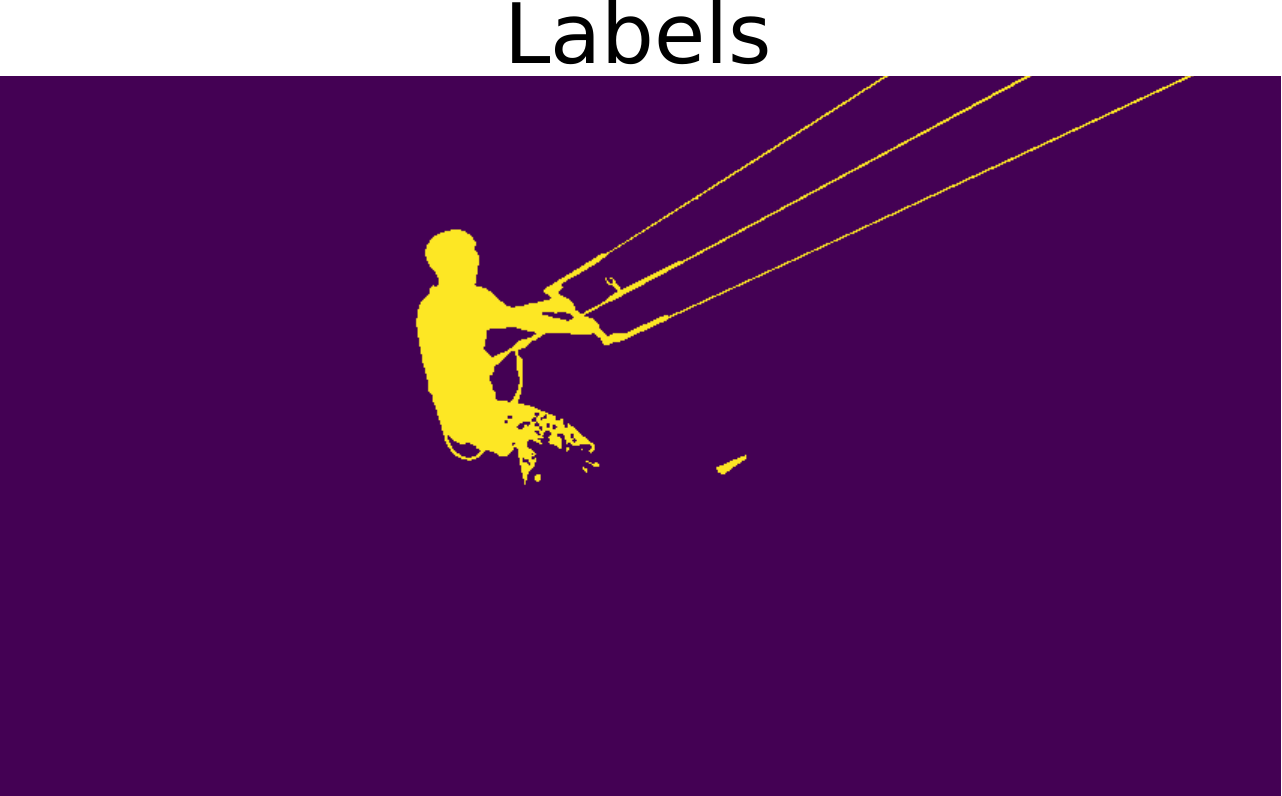}
    \end{tabular}
    \caption{Example of inferring the foreground mask for a single frame.}
    \label{fig:video_example}
\end{figure*}

This requires the inference of $m_t$ for each pixel $i$:
\begin{equation}
    s_i(m_t) \propto \exp\left(\sum_i \sum_{\h,f} p(\h|f) q_t(\h_{i,t}=\h|x_{i,t}) \log p(f_{i,t}=f|m_t)p(m_t)\right),
\end{equation}
and then optimizing $p_{m_t}$ as the count of times each instance is used,
\begin{equation}
    p(m_t) \propto \sum_i s_i(m_t) ~.
\end{equation}
Selection of instances $m_t$ in frame $t$ therefore involves comparing the predictions from the network $q_t(\h_{i,t}=\h|x_{i,t})$ grouped into foreground/background segmentation with the foreground/background segmentation for different instances from Mask R-CNN, and making a selection of a subset (probabilistically in $p(m_t)$) based on which instances most overlap with the predictions from network $q_t$. While the above two equations should in principle be iterated, and iterated with updates to network $q_t(\h_{i,t}=\h|x_{i,t})$, we found that in practice it is sufficient to just select the instances $m_t$ based on their intersection with the network predictions once, at the very beginning, to make a soft fixed prior, and leave it to optimizing the prediction network with the \textbf{RQ} loss to find confident segmentation (Fig.~\ref{fig:video_example}). 

We tested the approach on the DAVIS 2016 dataset \citep{DAVIS2016}. The dataset is comprised of 50 unique scenes, accompanied by per-pixel foreground/background segmentation masks. The objective is to produce foreground segmentation masks for all frames in a scene, given only the ground truth annotations of the first frame (Semi-Supervised). We evaluated our method on the 20-scene validation set at 480p resolution. 

The network $q$ used in this experiment combines both the pixel intensities and spatial position information for its predictions. At each pixel location $i,j$, we augment the intensity information with learned Fourier features $[\sin(W[i, j]^T),\ \cos(W[i, j]^T)]^T$ \citep{fourierFeats}. The image and spatial position are first processed separately; A 4-layer, 64-channel, fully-convolutional network with $3 \times 3$ kernels, ReLU activations and Batch Normalization produces the image features. A 3-layer, 16-channel, pixel-wise MLP with ReLU activations and Batch Normalization processes the learned Fourier features. These two are concatenated and passed through a single $3 \times 3$ convolution-ReLU-Batch Normalization layer before being mapped to output predictions. We also experimented with adding optical flow as another auxiliary input to the network.

\begin{figure*}
    \centering
    \begin{tabular}{@{}c@{\hspace{8pt}}c@{\hspace{8pt}}c@{}}
    \includegraphics[width=0.31\textwidth]{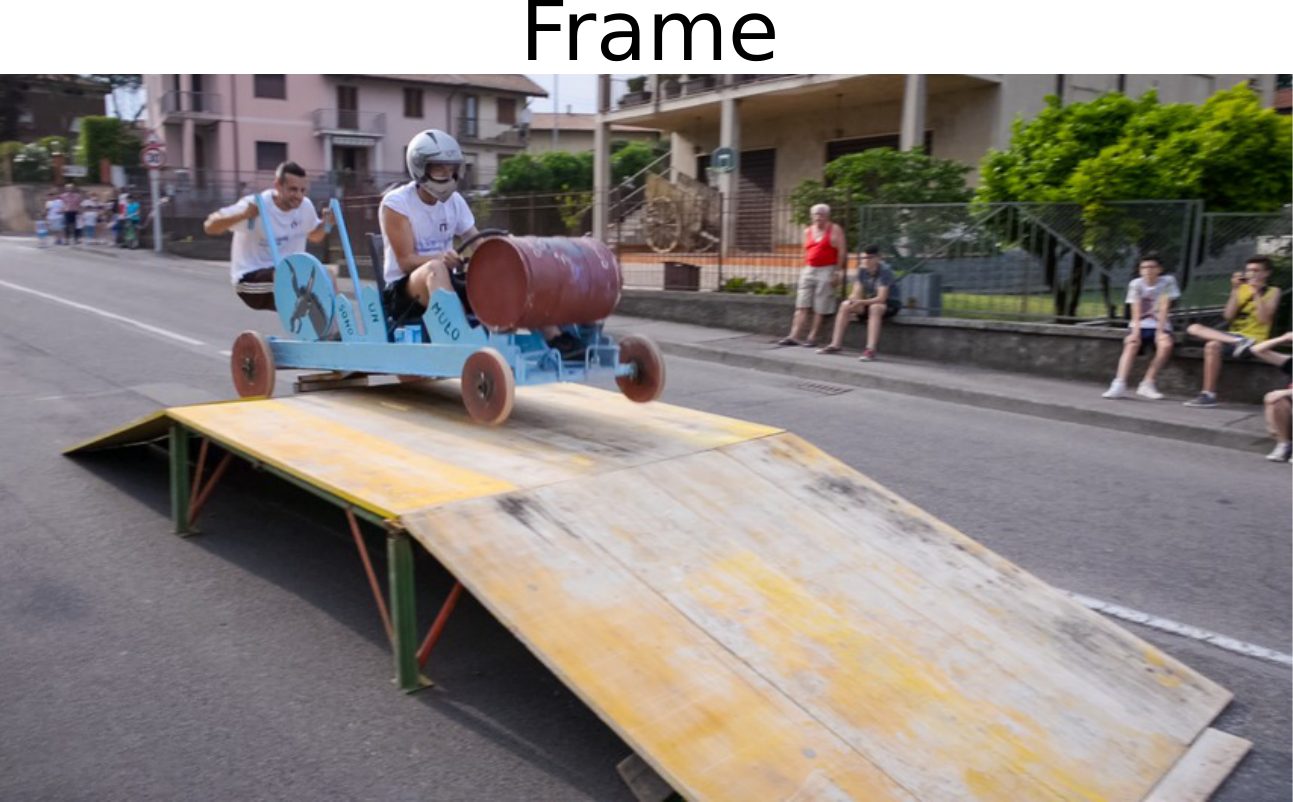}
    & \includegraphics[width=0.31\textwidth]{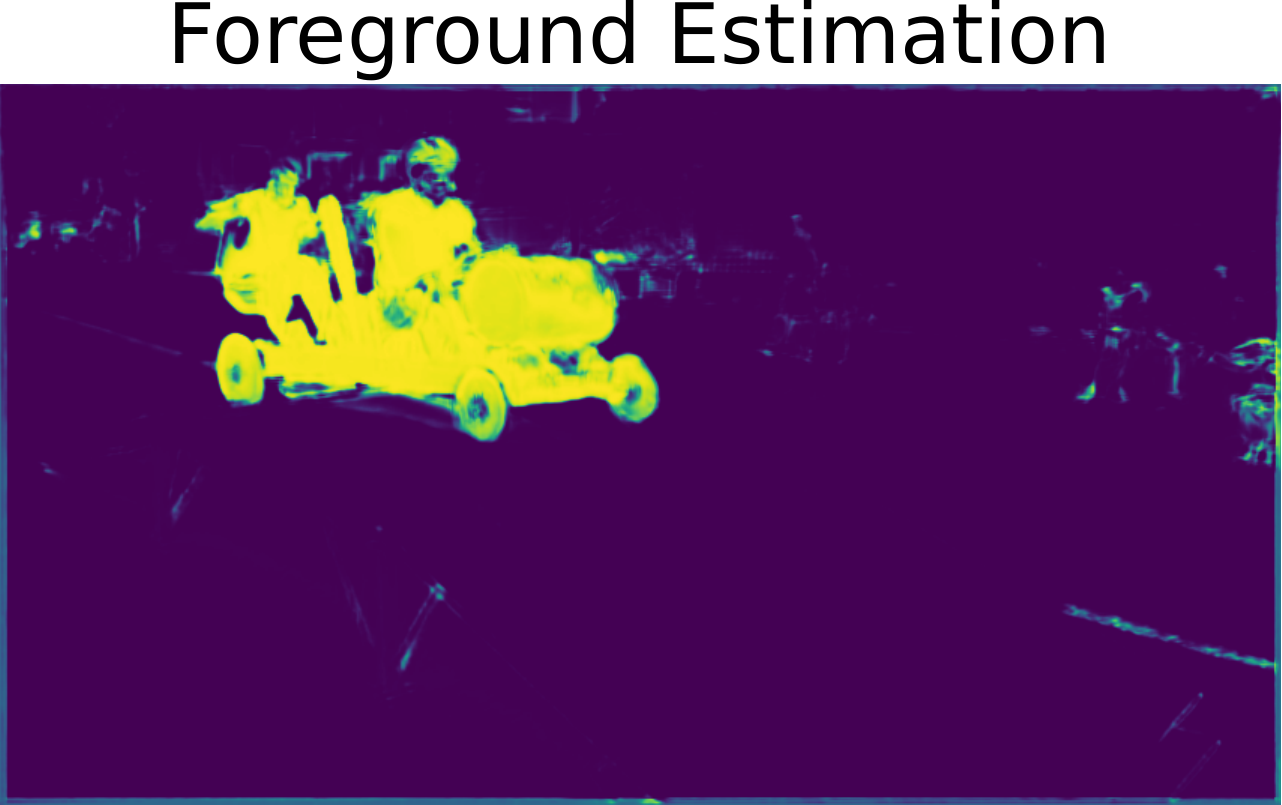} \\
    \end{tabular}
    \begin{tabular}{@{}c@{\hspace{8pt}}c@{\hspace{8pt}}c@{}@{\hspace{8pt}}c@{}}
    \includegraphics[width=0.31\textwidth]{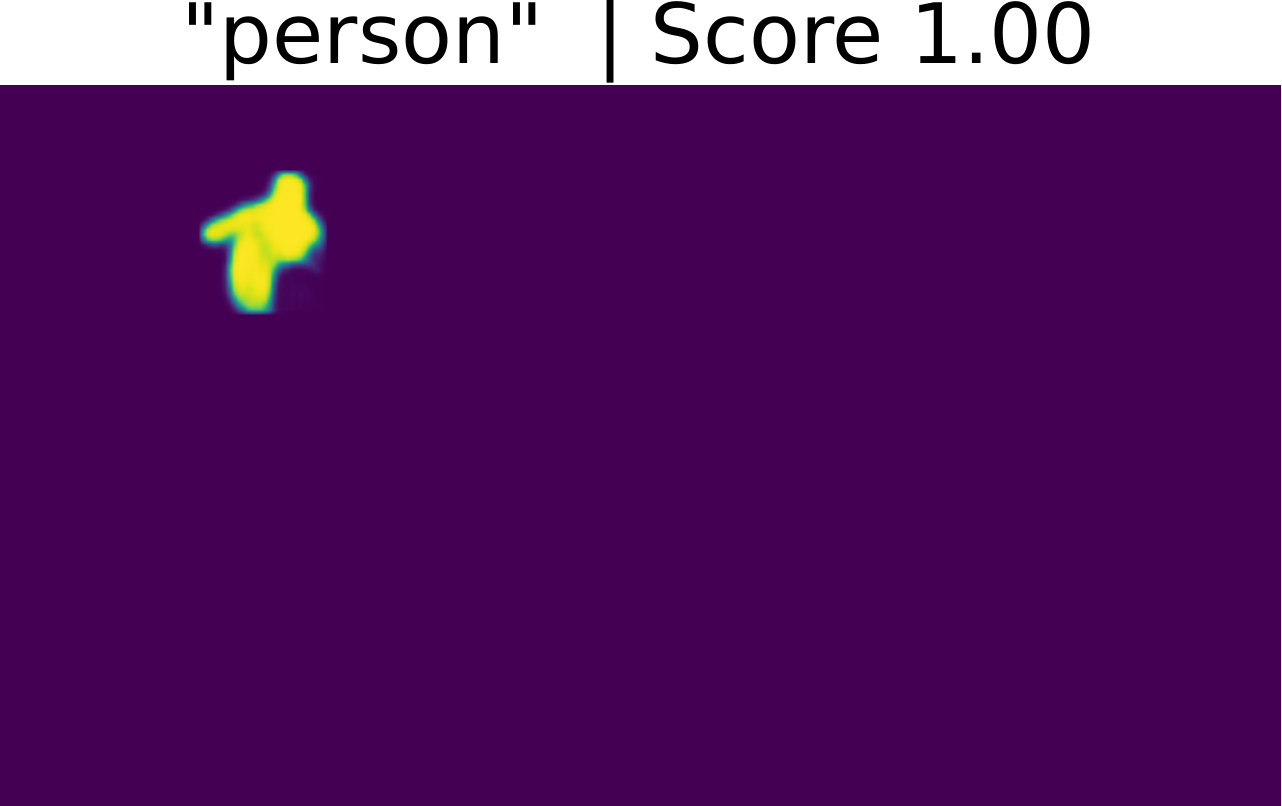}
    & \includegraphics[width=0.31\textwidth]{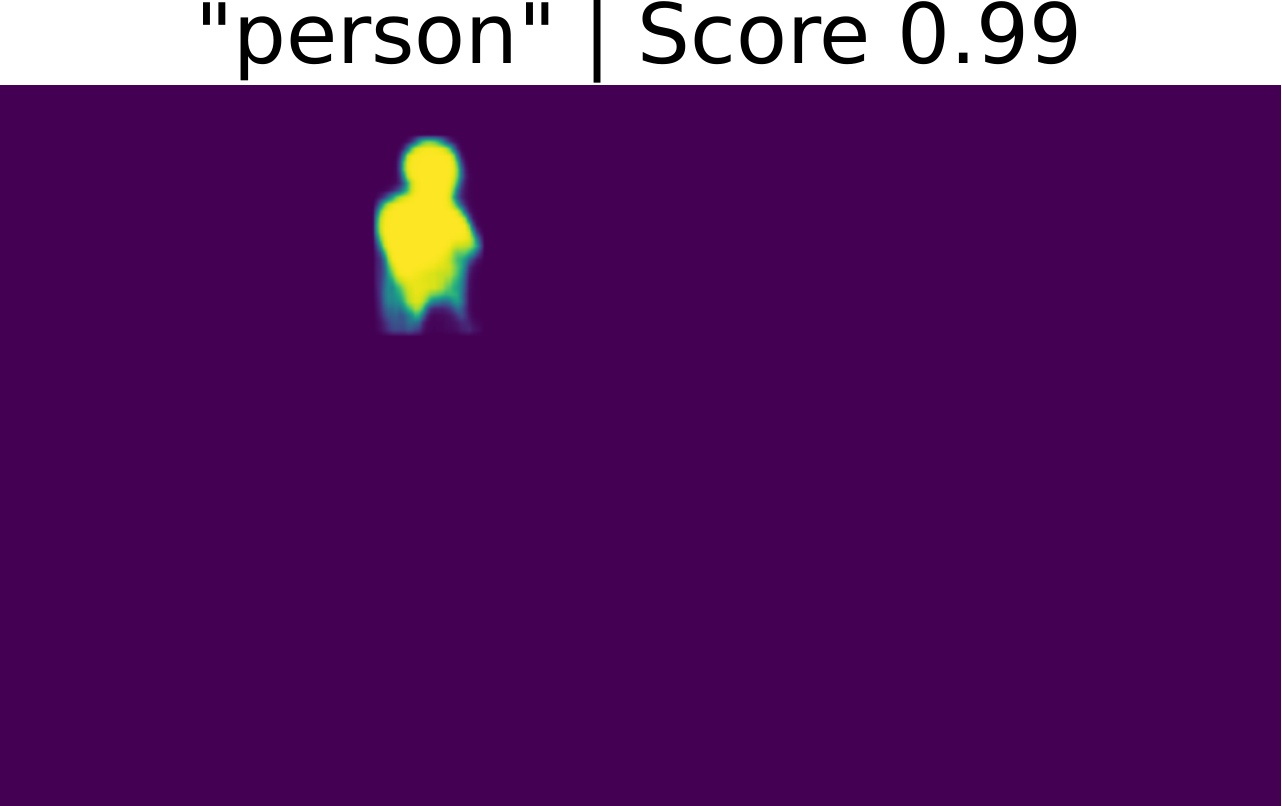}
    & \includegraphics[width=0.31\textwidth]{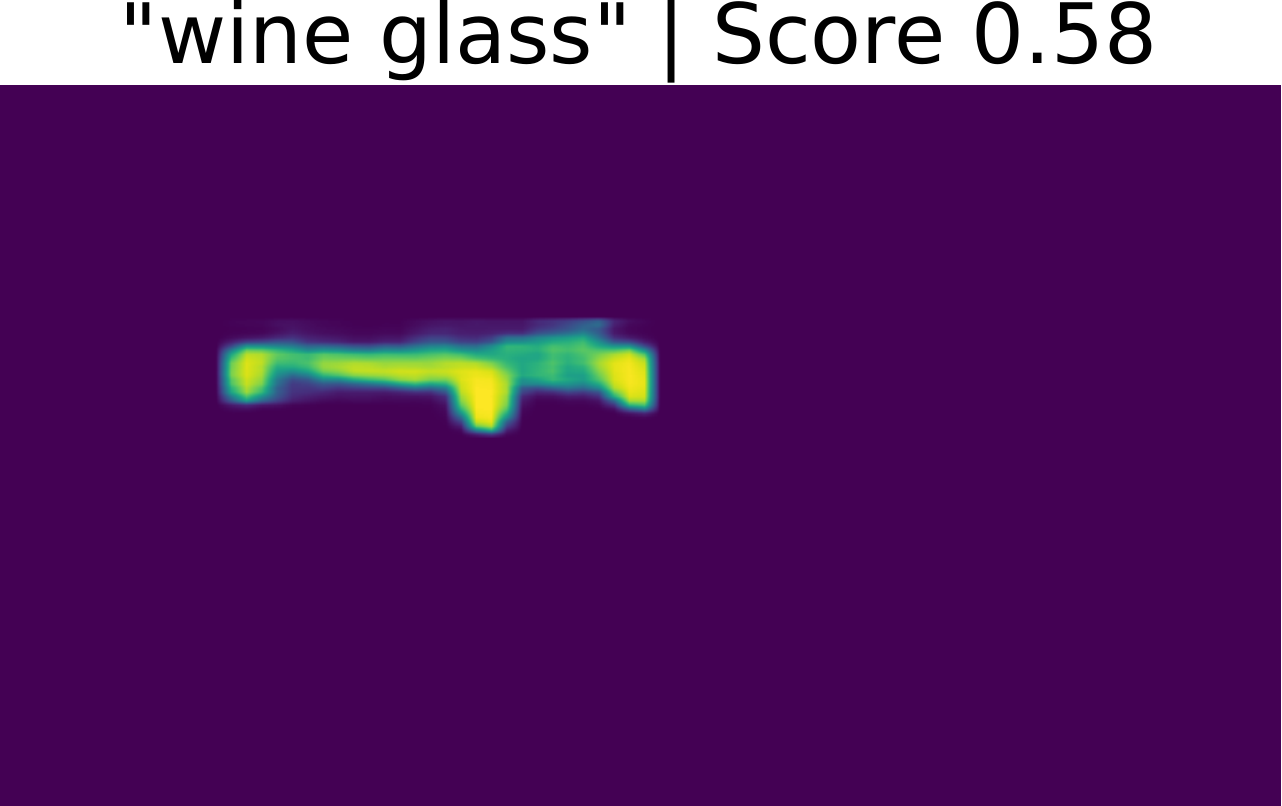} \\
    \includegraphics[width=0.31\textwidth]{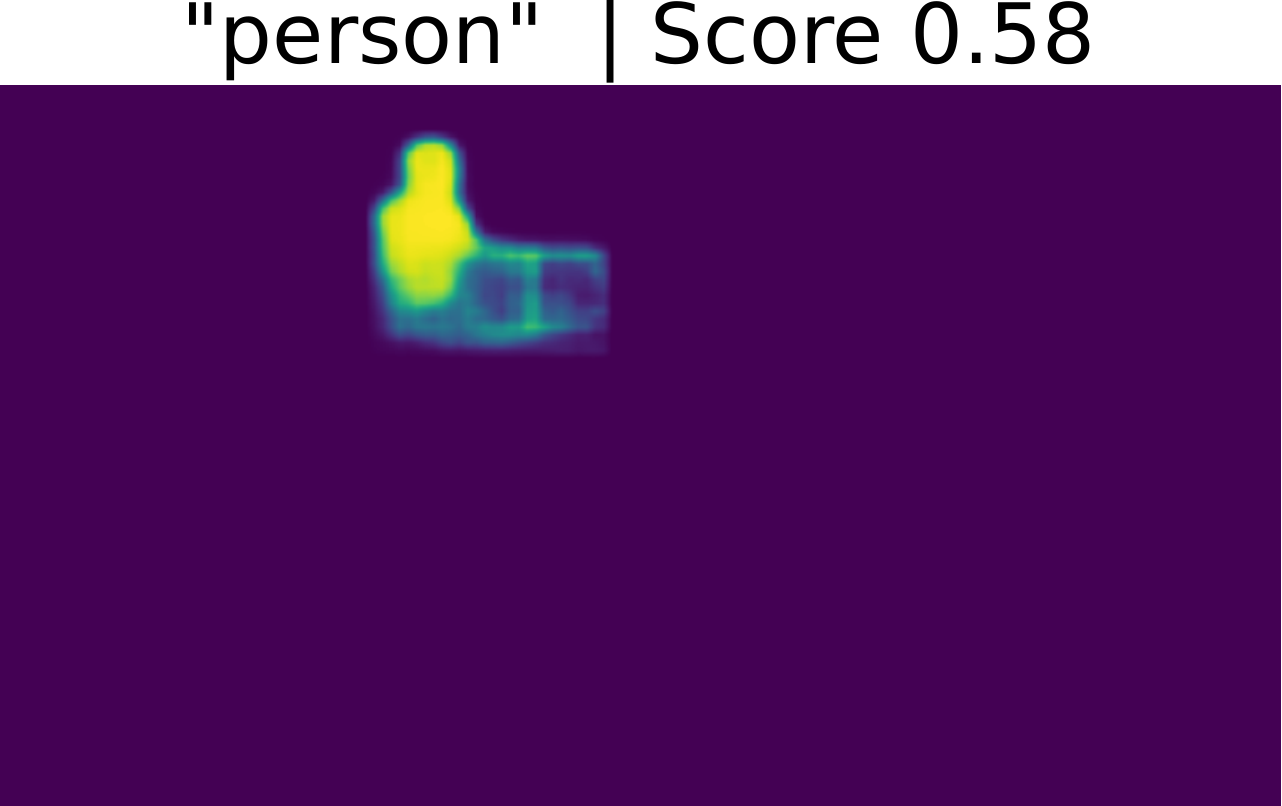}
    & \includegraphics[width=0.31\textwidth]{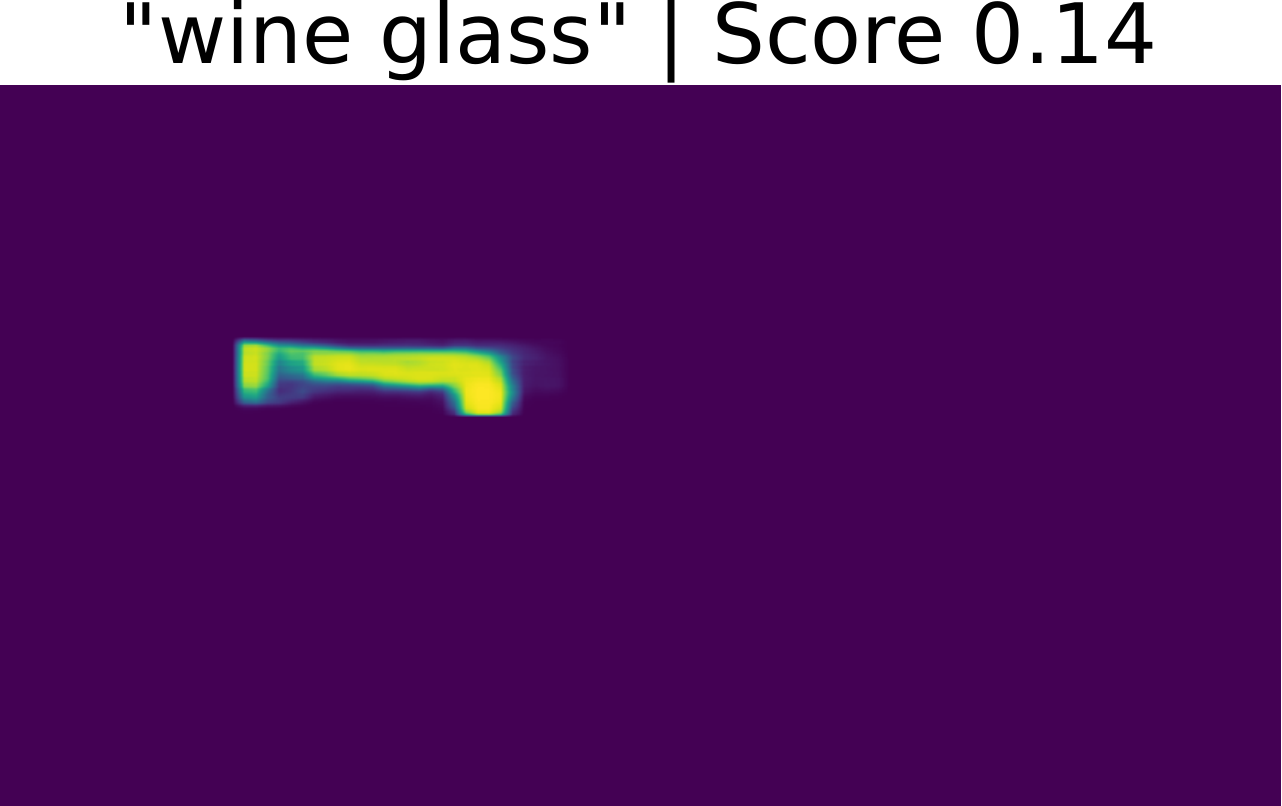}
    & \includegraphics[width=0.31\textwidth]{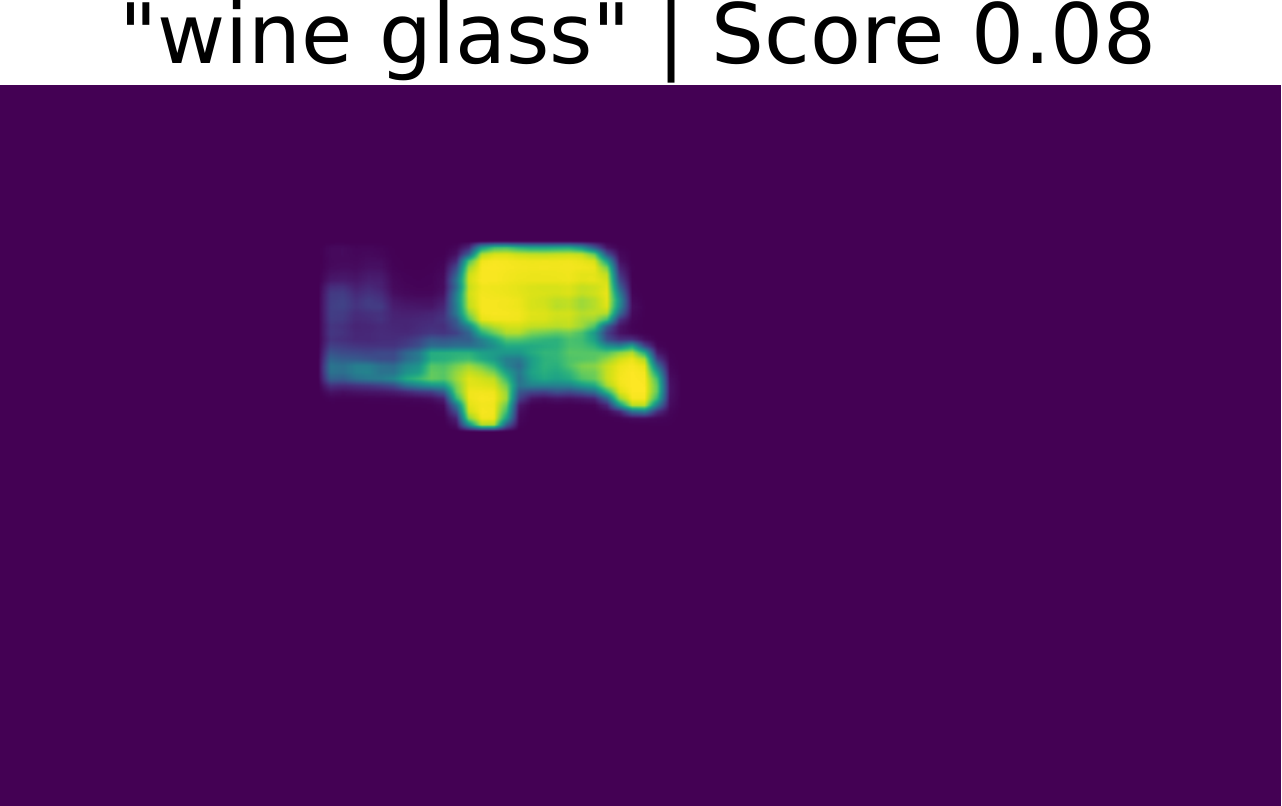} \\
    \end{tabular}
    \begin{tabular}{@{}c@{\hspace{8pt}}c@{\hspace{8pt}}c@{}@{\hspace{8pt}}c@{}}
    \includegraphics[width=0.31\textwidth]{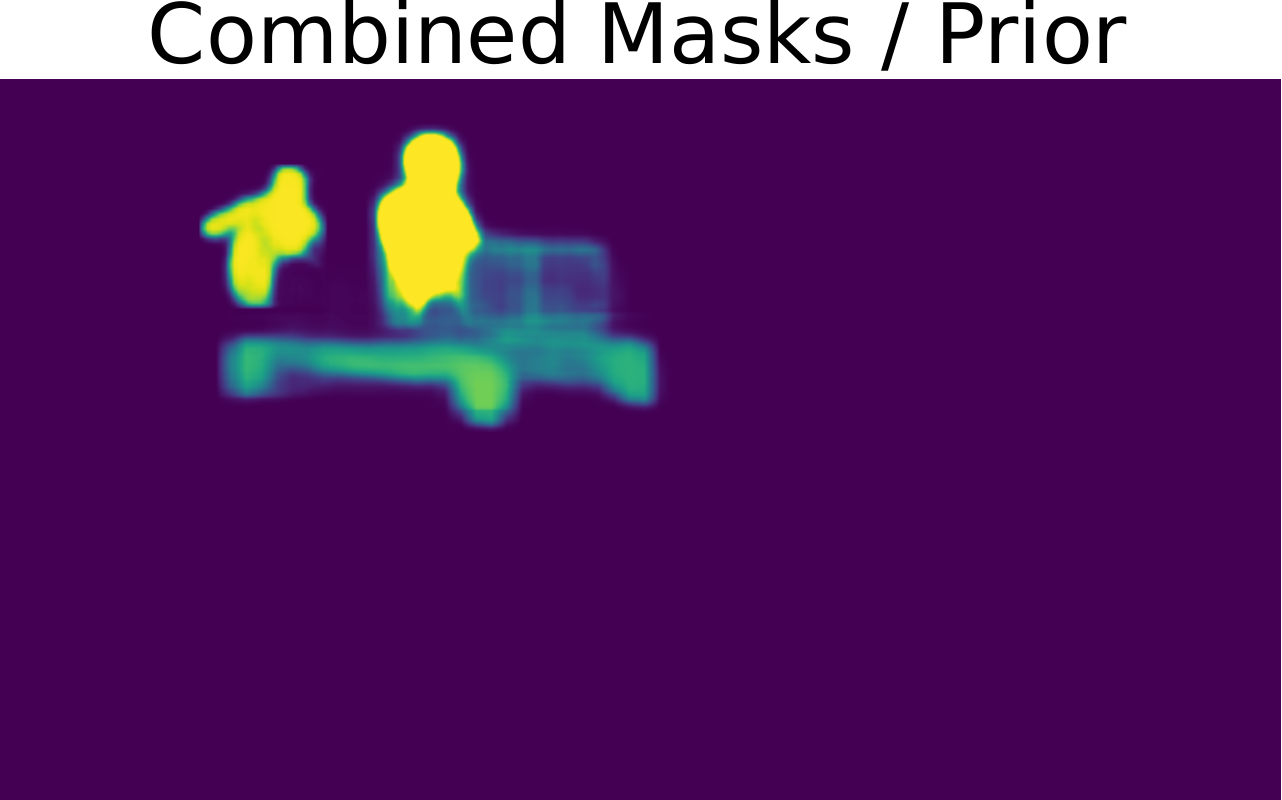}
    & \includegraphics[width=0.31\textwidth]{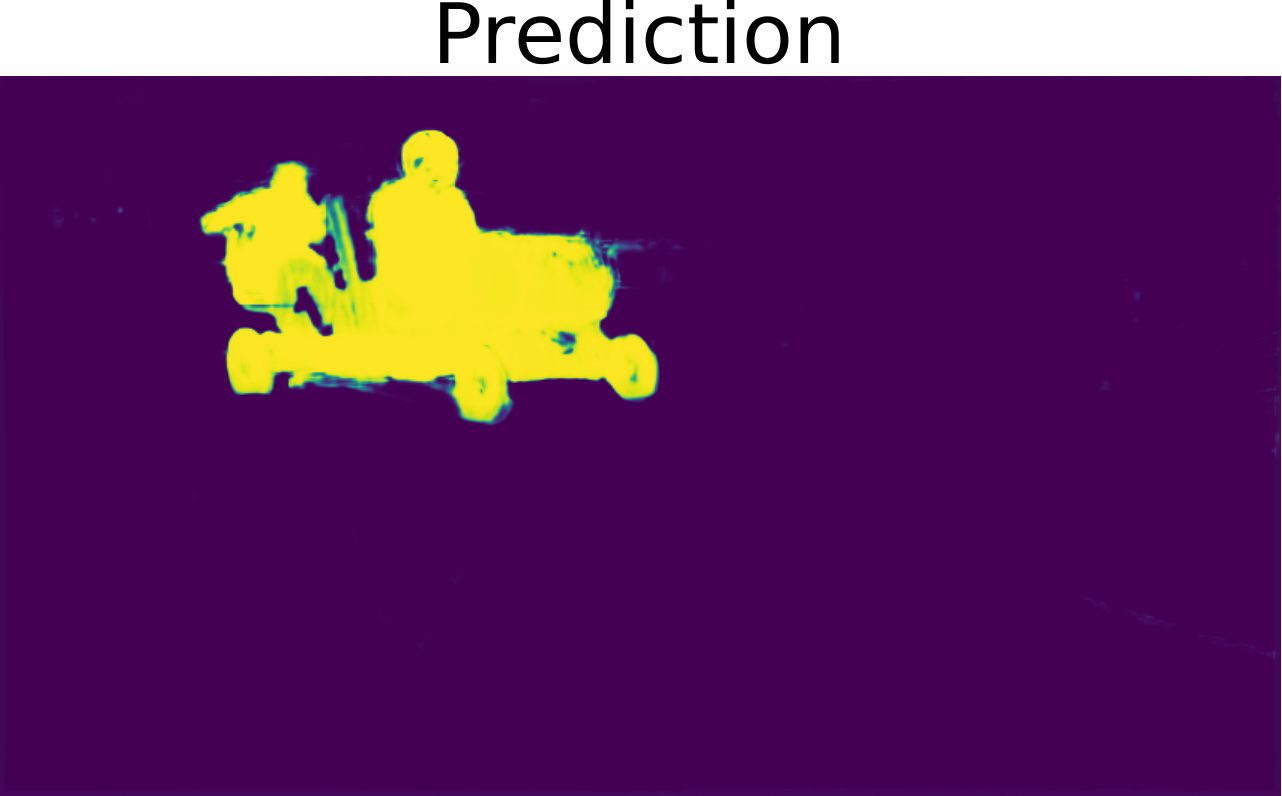}
    & \includegraphics[width=0.31\textwidth]{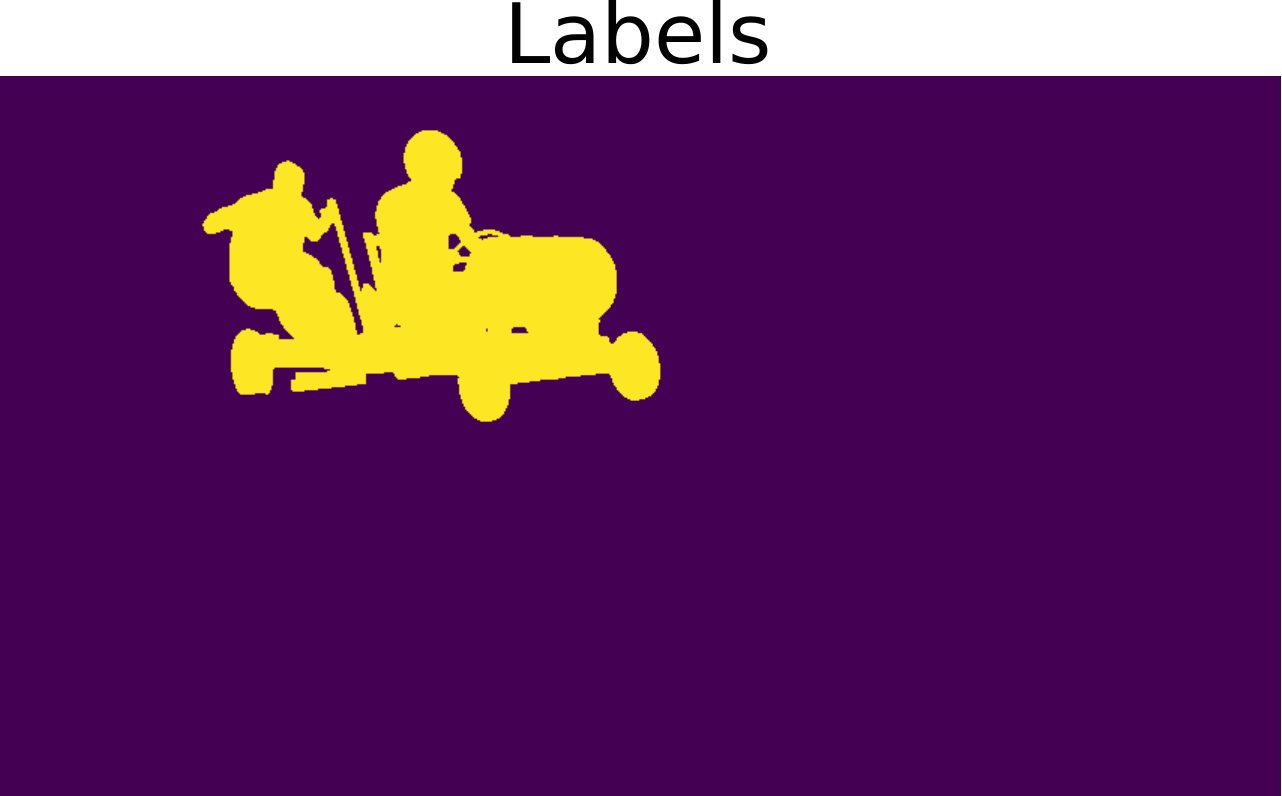}
    \end{tabular}
    \caption{Video frame segmentation procedure. Starting with a network $q_{t-1}$ trained on frame $t-1$, we apply $q_{t-1}$ on frame $t$ to get a rough foreground estimation (top). By running the pre-trained Mask R-CNN model on frame $t$ and selecting only the masks that overlap with the $q_{t-1}$ prediction we get the candidate object masks (middle). The prior is constructed as the sum of the candidate masks, weighted by their corresponding Mask R-CNN scores (bottom), and $q_{t-1}$ is finetuned on frame $t$ with this prior to produce the predictions (bottom).}
    \label{fig:video_procedure_figure}
\end{figure*}

For each scene, the network $q_0$ is trained on the first frame, using the given ground truth annotations split uniformly between 3 foreground and 5 background classes as prior, for 300 iterations. This network is then used to predict the foreground pixels in the next frame and after computing the intersection over union between the predicted foreground pixels and the Mask R-CNN output masks, we select masks that overlap more than a pre-specified threshold. The chosen masks are then summated, weighted by their Mask R-CNN confidence scores (0-1), to form the prior for the next frame. The process of selecting masks from the Mask-RCNN predictions and forming the prior for a frame is showcased in Figure \ref{fig:video_procedure_figure}. The network $q_0$ is then fine-tuned for 10 iterations to obtain $q_1$ and this process repeats for all subsequent frames. We used the Adam optimizer, with a starting learning rate of $10^{-3}$ for the first frame, reduced to $10^{-5}$ for fine-tuning, and trained with batches of 128 64$\times$64 patches. 

To infer the foreground pixels we start with a Mask R-CNN pre-trained on the COCO dataset. Then, for each scene we only require $\sim$1min of training time on the ground truth-annotated first frame and $\sim$3s per every following frame for the entire process of forming the prior and inferring the foreground pixels. We do not train on any video data, in contrast to most video object segmentation methodologies that rely on both a pre-trained network on static image datasets (such as COCO) and additionally on offline training on video sequences. In Table \ref{tab:tracking_results} we compare our results on the DAVIS 2016 validation set to other video object segmentation algorithms from 2017 - present.

\begin{table*}[bh]
    \centering
    \caption{Jaccard and F1 measures for various algorithms on the video instance segmentation task.}
    \resizebox{\linewidth}{!}{
\begin{tabular}{lcccccccc}
    \toprule
     &  & \multicolumn{3}{c}{J} & \multicolumn{3}{c}{F} &  \\ \cmidrule(lr){3-5}\cmidrule(lr){6-8}
    Model & J\&F $\uparrow$ & Mean $\uparrow$ & Recall $\uparrow$ & Decay $\downarrow$ & Mean $\uparrow$ & Recall $\uparrow$ & Decay $\downarrow$ & Year \\\midrule
    OSVOS \cite{Cae17_osvos} & 80.2 & 79.8 & 93.6 & 14.9 & 80.6 & 92.6 & 15 & 2017 \\
    MSK \cite{Perazzi_2017_CVPR} & 77.55 & 79.7 & 93.1 & 8.9 & 75.4 & 87.1 & 9 & 2017 \\
    OnAVOS \cite{Voigtlaender_onavos} & 85.5 & 86.1 & 96.1 & 5.2 & 84.9 & 89.7 & 5.8 & 2017 \\
    Lucid \cite{LucidDataDreaming_CVPR17_workshops} & 82.95 & 83.9 & 95 & 9.1 & 82 & 88.1 & 9.7 & 2017 \\
    OSVOS-S \cite{Maninis_osvoss} & 86.55 & 85.6 & 96.8 & 5.5 & 87.5 & 95.9 & 8.2 & 2018 \\
    FAVOS \cite{Cheng_favos_2018} & 80.95 & 82.4 & 96.5 & 4.5 & 79.5 & 89.4 & 5.5 & 2018 \\
    PReMVOS \cite{luiten2018premvos} & 86.75 & 84.9 & 96.1 & 8.8 & 88.6 & 94.7 & 9.8 & 2018 \\
    OSMN \cite{Yang2018osmn} & 73.45 & 74 & 87.6 & 9 & 72.9 & 84 & 10.6 &  2018 \\
    AGAME \cite{Johnander_AGAME} & 81.85 & 81.5 & 93.6 & 9.4 & 82.2 & 90.3 & 9.8 &  2019 \\
    STM \cite{seoung_stm} & 89.4 & 88.7 & 97.4 & 5 & 90.1 & 95.2 & 4.2 &  2019 \\
    FEELVOS \cite{Voigtlaender_2019_CVPR} & 81.65 & 81.1 & 90.5 & 13.7 & 82.2 & 86.6 & 14.1 &  2019 \\
    CFBI \cite{yang2020CFBI} & 89.4 & 88.3 & - & - & 90.5 & - & - & 2020 \\
    e-OSVOS \cite{e_osvos_2020_NeurIPS} & 86.8 & 86.6 & - & - & 87 & - & - & 2020 \\
    STCN \cite{cheng2021stcn} & 91.7 & 90.4 & 98.1 & 4.1 & 93 & 97.1 & 4.3 &  2021 \\ \midrule
    Ours & 83.8 & 84 & 96.2 & 8.4 & 83.6 & 94.2 & 10.2 & \\
    Ours (+flow) & 83.9 & 83.2 & 95.5 & 9.5 & 84.6 & 93.3 & 9.1 & \\
    \bottomrule
\end{tabular}
}
    \label{tab:tracking_results}
\end{table*}

\newpage
\subsection{In-collection inference for multi-domain learning: return to Le S\'educteur}
\label{sec:seducer}
\begin{figure*}[ht]
    \centering
    \begin{tabular}{@{}c@{\hspace{8pt}}c@{\hspace{8pt}}c@{}}
    \includegraphics[width=0.31\textwidth]{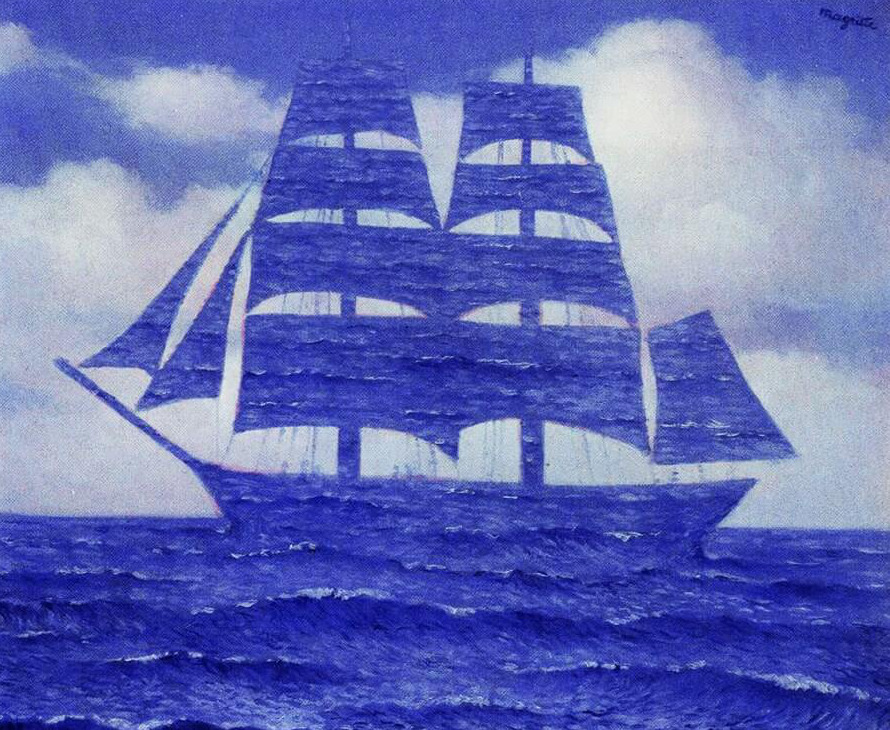} 
    & \includegraphics[width=0.31\textwidth]{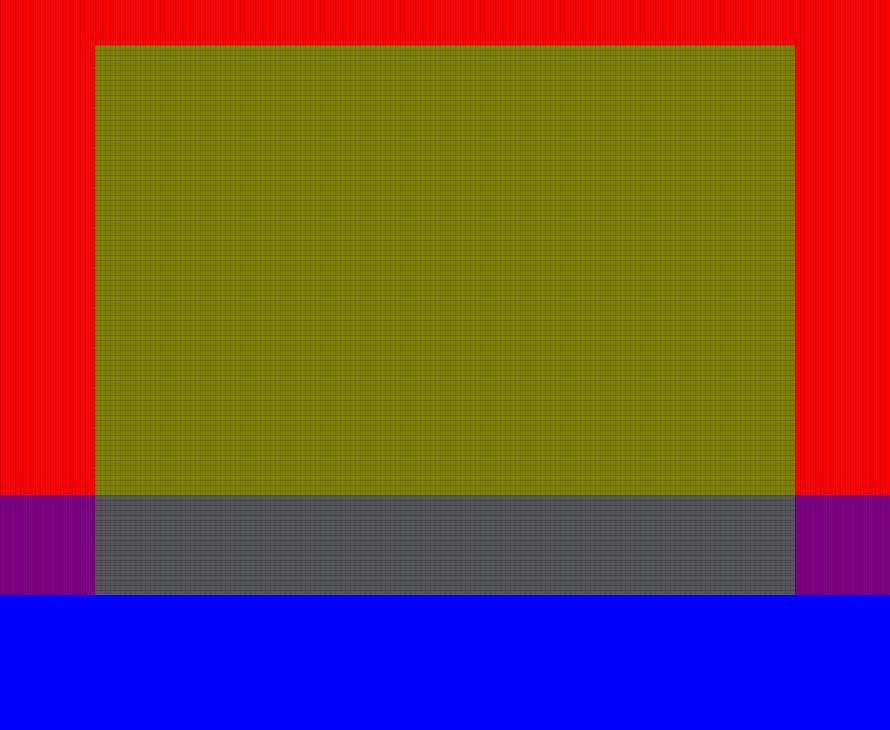}
    & \includegraphics[width=0.31\textwidth]{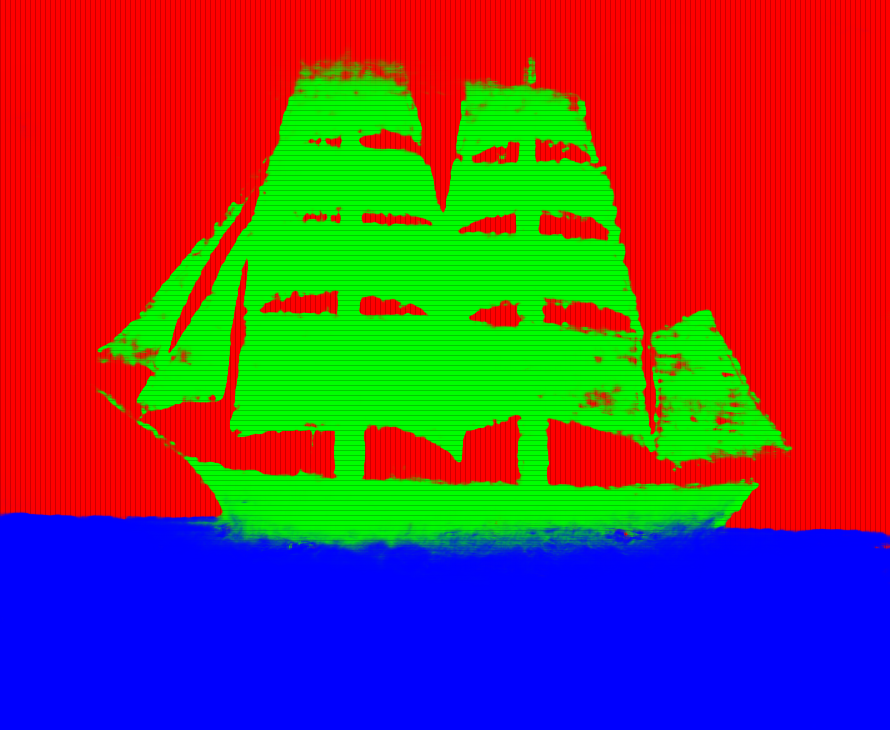} \\
    \includegraphics[width=0.31\textwidth]{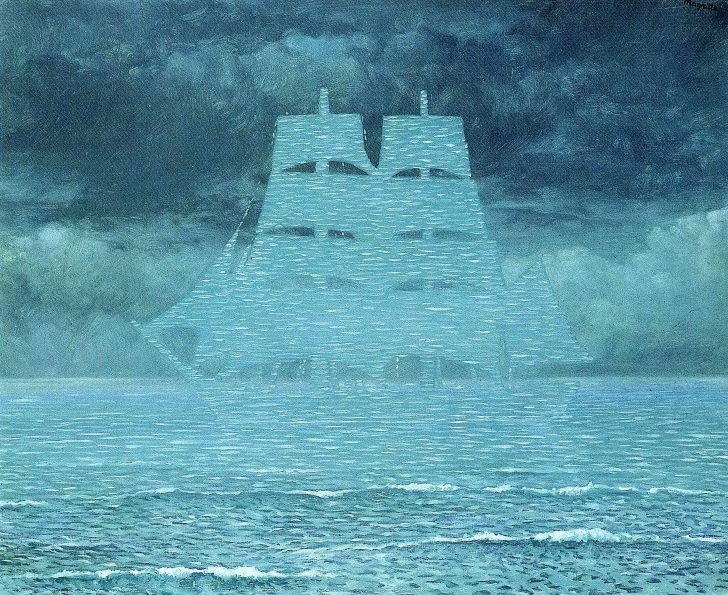} 
    & \includegraphics[width=0.31\textwidth]{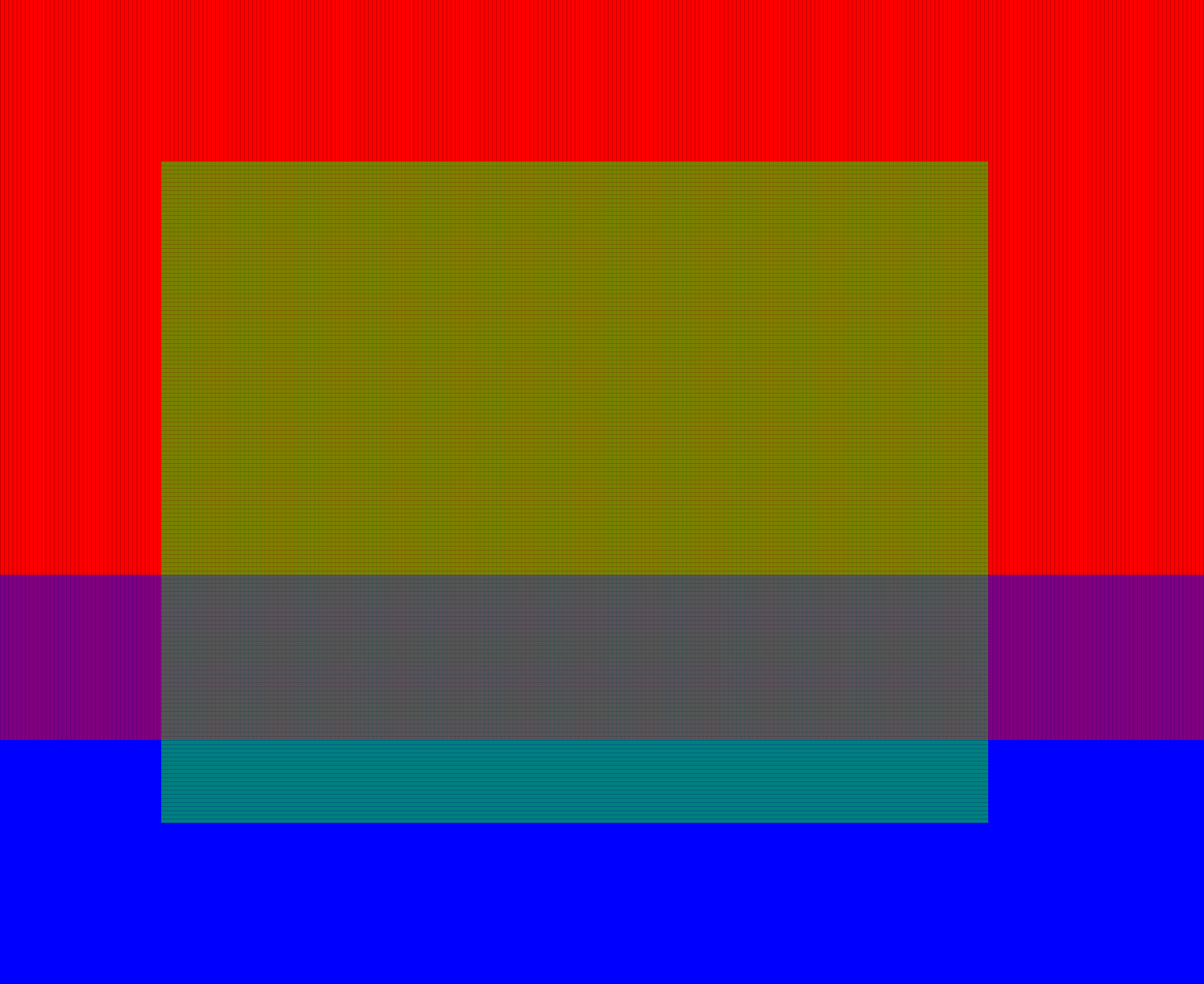}
    & \includegraphics[width=0.31\textwidth]{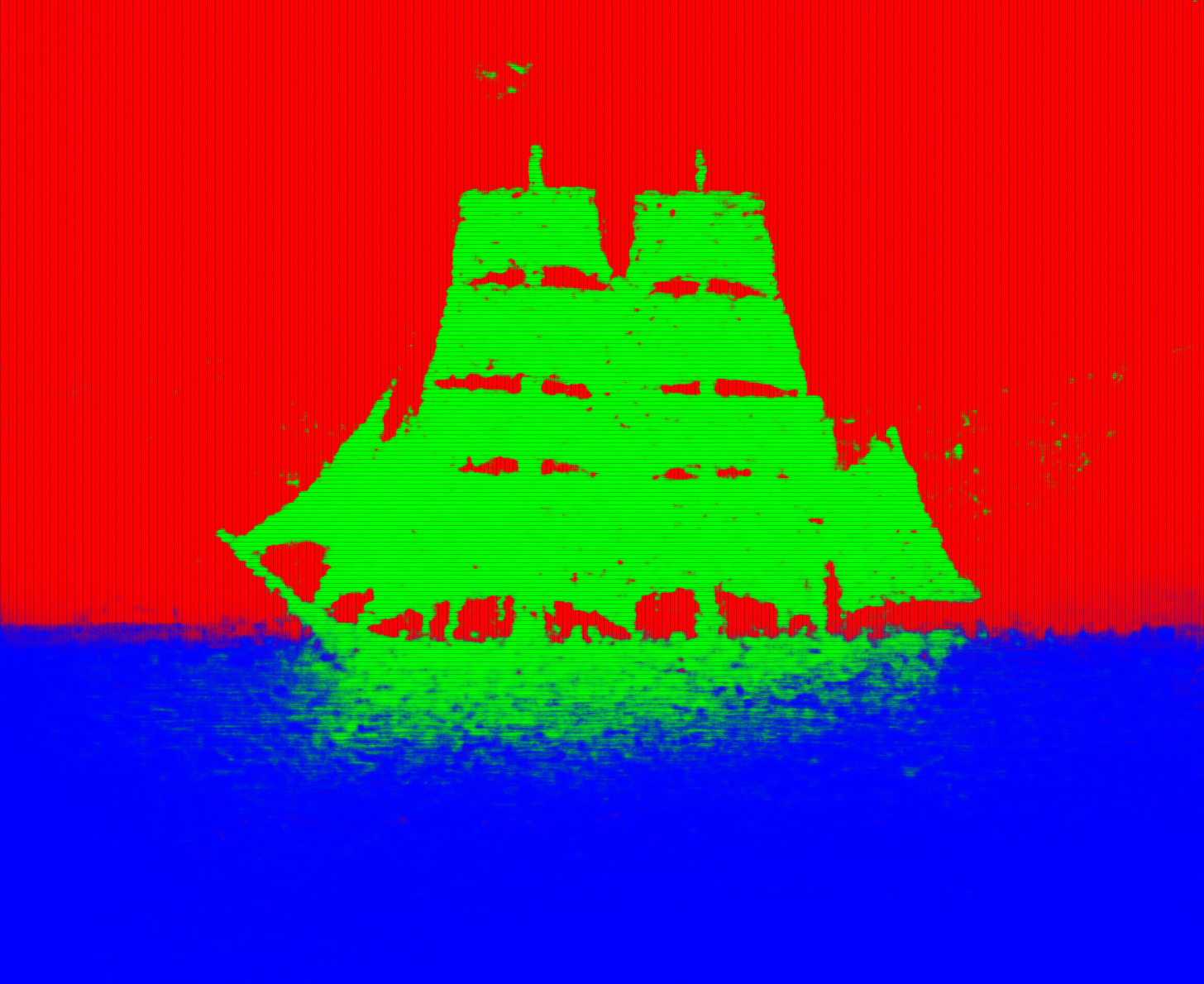}\\
    \end{tabular}
    \caption{Two additional versions of \textit{Le s\'educteur} (left), hand-made priors (middle) and inferred segmentations (right).}
    \label{fig:seducer_v2}
\end{figure*}

\begin{figure*}[h]
    \centering
    \begin{tabular}{@{}c@{\hspace{8pt}}c@{\hspace{8pt}}c@{}}
    \includegraphics[width=0.31\textwidth]{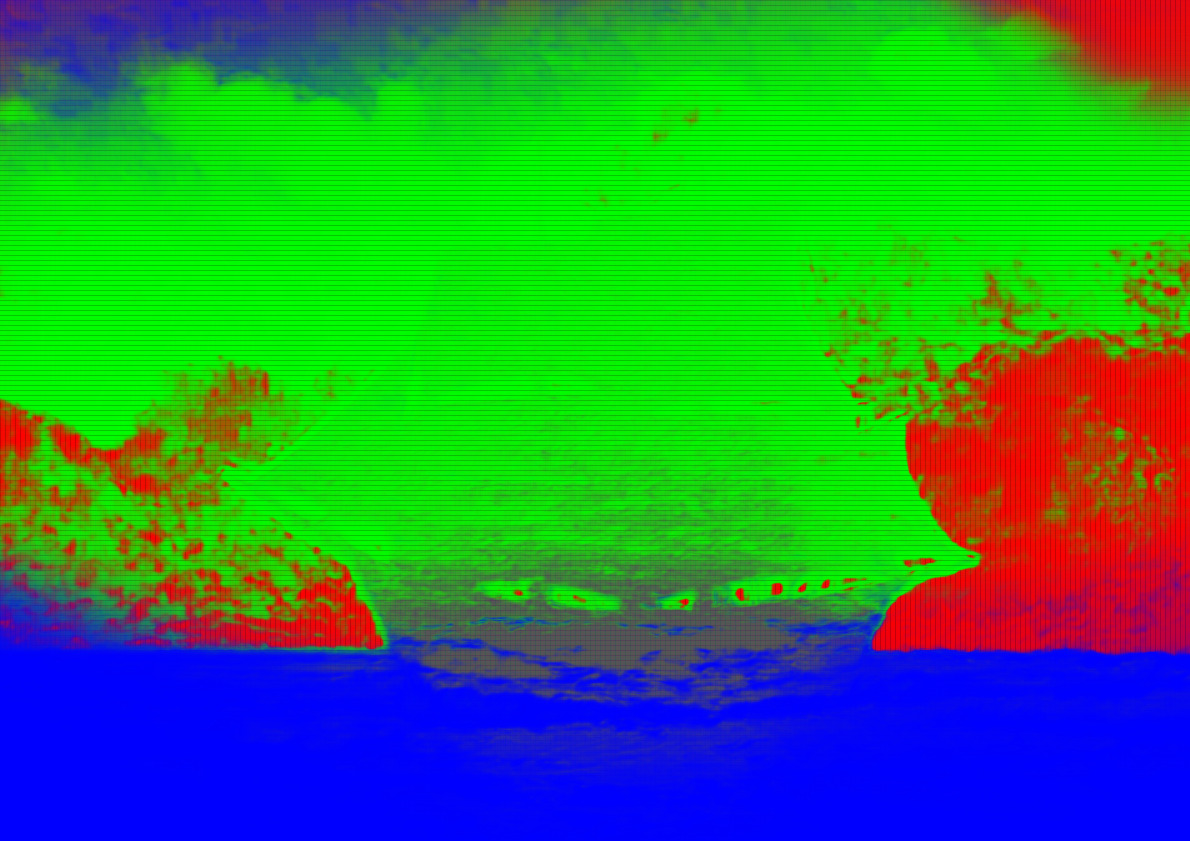} 
    & \includegraphics[width=0.31\textwidth]{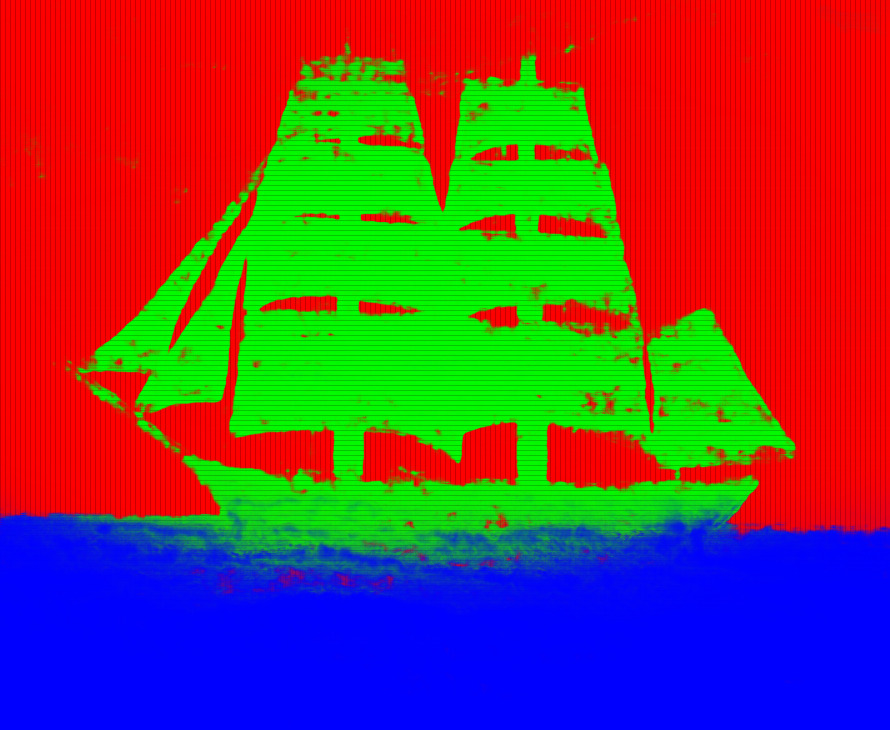}
    & \includegraphics[width=0.31\textwidth]{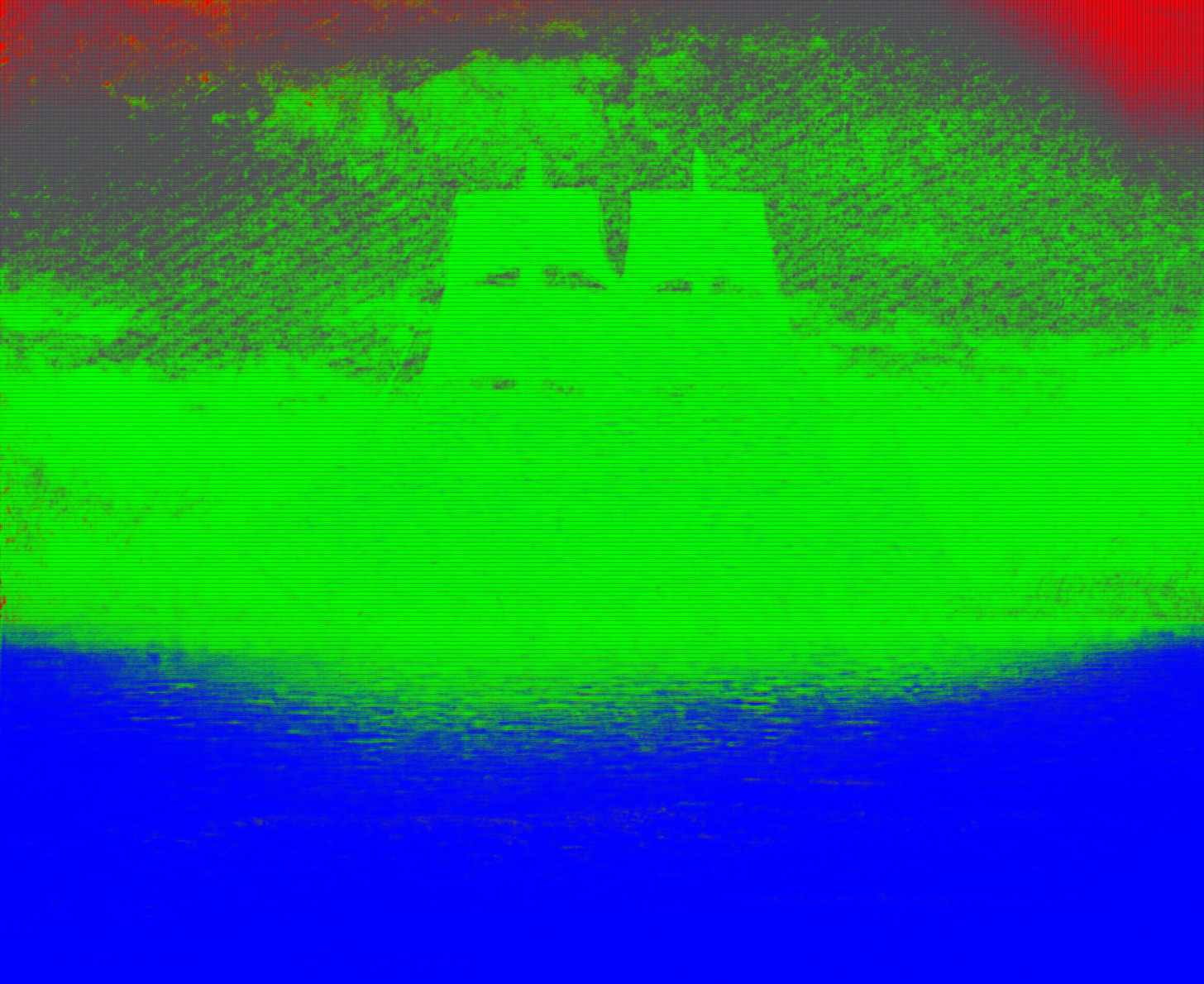}\\
    (a)
    & (b)
    & (c)
    \end{tabular}
    \caption{Result of applying a network $q$ trained to infer (b), on all three \textit{Le s\'educteur} versions.}
    \label{fig:seducer_seducing}
\end{figure*}

One of the conclusions from our experiments on the EnviroAtlas landcover mapping task (\S\ref{sec:enviroatlas_experiments}) is that training a network with the goal of generalizing to new input data is often inferior to simply performing in-collection inference for each domain . In other words, given the collection of pairs $x_i, p_i(\h)$, learning the posterior $q$ under the implicit posterior  model is optimized for resolving ambiguities in that collection, and possibly that collection alone. As pointed out in \cite{malkin2020mining}, which performs collection inference using large generative models to mine self-similarity among the examples in the collection, this is appropriate when we can expect our data $x_i$ to always come paired with prior beliefs $p(\h_i)$. It is interesting to reconsider the Seducer example from Fig.~\ref{fig:examples}. The artist created several versions of that painting in differing styles. Fig. \ref{fig:seducer_v2} shows that collection inference applied separately to each of these paintings works equally well. However, using a learned $q$ network from one image onto others yields inferior segmentations (Fig. \ref{fig:seducer_seducing}), as the learned network specialized for inference in the data it saw. (A fully generative model would be expected to similarly overtrain on the input data features $x_i$, as would a supervised  neural network trained on hard-labeled pairs $(x_i,\h_i)$ due to the  domain shift.) Yet, if we know we will always be given collections with beliefs in the form of priors $p_i(\h)$, local (collection) inference may be all we need.

\end{document}